%% file: main.tex
\documentclass[11pt, a4paper]{article}

\usepackage[margin=1in]{geometry}
\usepackage[utf8]{inputenc}
\usepackage[T1]{fontenc}
\usepackage{lmodern}
\usepackage[dvipsnames, table]{xcolor}     %
\usepackage{pifont}   %
\usepackage{multirow}
\usepackage{booktabs} %
\usepackage{siunitx}  %
\usepackage{fancyvrb}
\usepackage{microtype}
\usepackage{placeins}

\definecolor{tiiPurple}{RGB}{122, 0, 255}
\usepackage[x11names]{xcolor} 
\usepackage{hyperref}
\usepackage{subcaption} %
\hypersetup{
    colorlinks=true,
    linkcolor=tiiPurple,
    urlcolor=tiiPurple,
    citecolor=tiiPurple
}

\usepackage[most]{tcolorbox}
\usepackage{graphicx}
\usepackage{tabularx}
\usepackage[useregional]{datetime2}
\DTMsetdatestyle{iso}  %
\usepackage[normalem]{ulem}
\usepackage{wrapfig}
\usepackage{titlesec}
\usepackage{adjustbox}
\usepackage{enumitem}
\usepackage{booktabs}
\usepackage{multirow}
\usepackage{array}
\usepackage{float}
\usepackage{caption}
\usepackage{amsfonts}       %
\usepackage{amsmath, amssymb}       %
\usepackage{nicefrac}       %
\usepackage{hyperref}       %
\usepackage{url}            %
\usepackage[utf8]{inputenc} %
\usepackage[T1,T5]{fontenc}    %
\usepackage{natbib}
\usepackage{graphicx}
\usepackage{enumitem}
\usepackage{makecell}

\definecolor{bestcolor}{RGB}{220,255,220}
\usepackage{CJKutf8}
\usepackage[para]{threeparttable}

\usepackage{xspace}

\makeatletter
\DeclareRobustCommand\onedot{\futurelet\@let@token\@onedot}
\def\@onedot{\ifx\@let@token.\else.\null\fi\xspace}

\def\eg{\emph{e.g}\onedot} \def\Eg{\emph{E.g}\onedot}
\def\ie{\emph{i.e}\onedot} 
 
\def\etc{\emph{etc}\onedot} \def\vs{\emph{vs}\onedot}

\makeatother

\titleformat{\section}
  {\normalfont\Large\bfseries}{\thesection.}{1em}{}
\usepackage{fancyhdr}
\begin{document}
\noindent
\begin{minipage}[t]{0.49\textwidth}
    \vspace{-4.2em}
    \includegraphics[height=1.3cm,width=2.5cm]{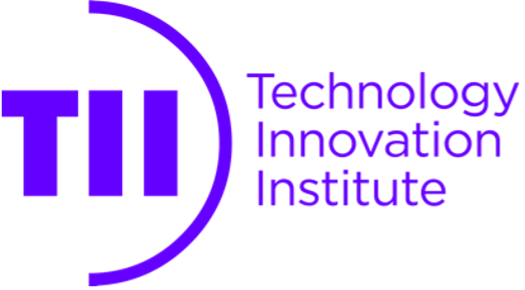}  %
\end{minipage}%
\hfill
\begin{minipage}[t]{0.49\textwidth}
    \vspace{-1.5em}
\end{minipage}
\vspace{-0.5em}
\hrule
\vspace{3em}

{\noindent \huge \textbf{\textcolor{tiiPurple}{Falcon Perception}}}

\noindent

\vspace{1.2em}
\noindent\href{https://github.com/tiiuae/Falcon-Perception}{https://github.com/tiiuae/Falcon-Perception}

\vspace{0.5em}

{\noindent \small Falcon Vision Team, TII}

\vspace{1.5em}

    \noindent Perception-centric systems are typically implemented with a modular encoder-decoder pipeline: a vision backbone for feature extraction and a separate decoder (or late-fusion module) for task prediction. This raises a central question: \emph{is this architectural separation essential or can a single early-fusion stack do both perception and task modeling at scale?} We introduce \textbf{Falcon Perception}, a unified dense Transformer that processes image patches and text tokens in a shared parameter space from the first layer, using a hybrid attention pattern (bidirectional among image tokens, causal for prediction tokens) to combine global visual context with autoregressive, variable-length instance generation. To keep dense outputs practical, Falcon Perception retains a lightweight token interface and decodes continuous spatial outputs with specialized heads, enabling parallel high-resolution mask prediction.

    Our design promotes simplicity: we keep a single scalable backbone and shift complexity toward data and training signals, adding only small heads where outputs are continuous and dense. On SA-Co, Falcon Perception improves mask quality to 68.0 Macro-F$_1$ compared to 62.3 of SAM3. We also introduce \textbf{PBench}, a benchmark targeting compositional prompts (OCR, spatial constraints, relations) and dense long-context regimes, where the model shows better gains. Finally, we extend the same early-fusion recipe to \textbf{Falcon OCR}: a compact 300M-parameter model which attains 80.3\% on olmOCR and 88.64 on OmniDocBench.

\tableofcontents
\clearpage
\begin{figure}
    \centering
    \includegraphics[width=\linewidth, trim=0 2cm 0 2cm, clip]{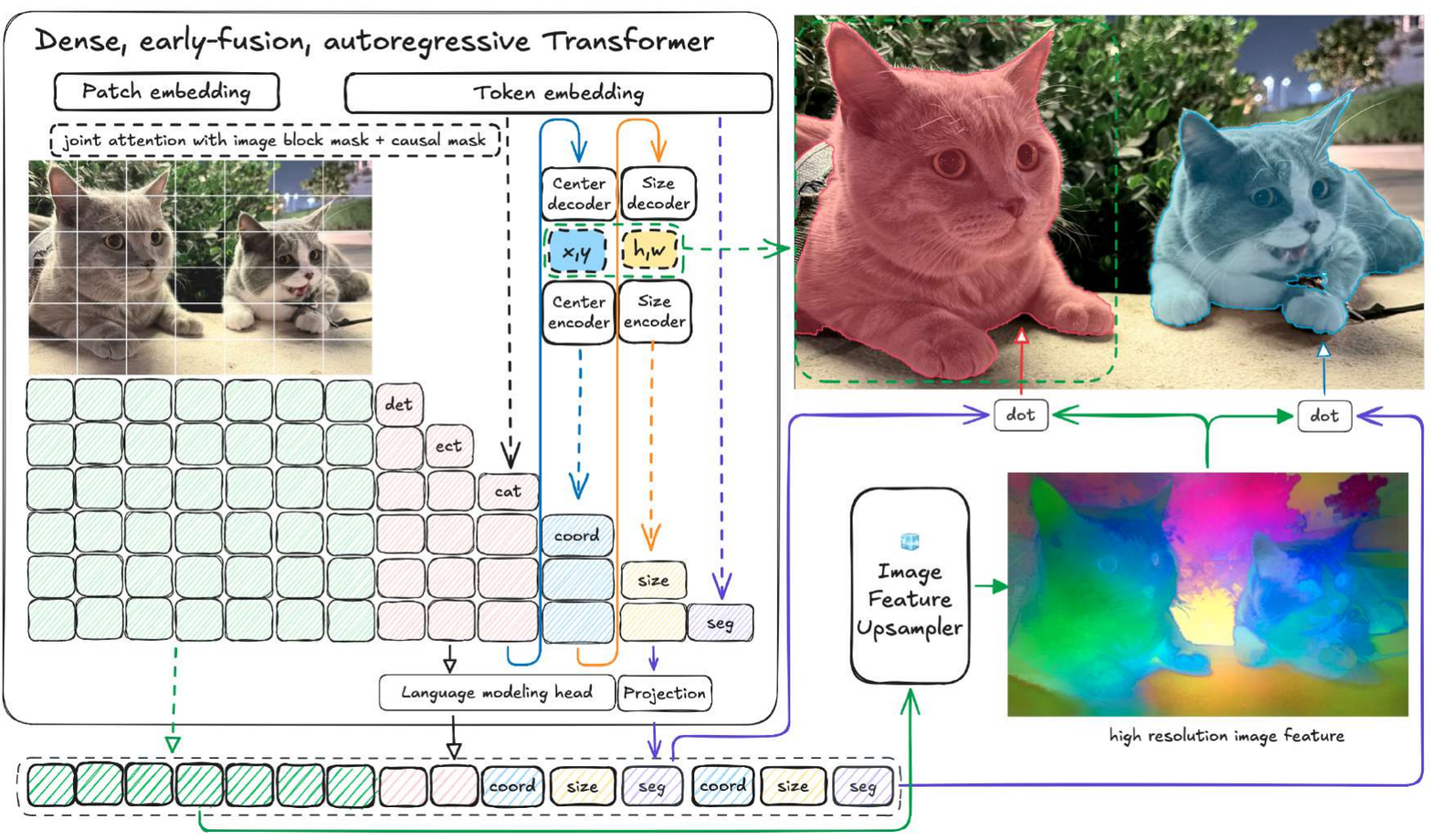}
    \caption{\textbf{Falcon Perception Architecture:} A single autoregressive Transformer processes a unified sequence of image patches, text, and task tokens. The model predicts object properties in a fixed order: \texttt{<coord>} $\to$ \texttt{<size>} $\to$ \texttt{<segm>}. Bounding boxes' coordinate and size tokens are decoded via specialized heads and re-injected as Fourier features to condition subsequent steps. High resolution segmentation masks are generated by a dot product between the \texttt{<segm>} token of each instance and the upsampled image features, leveraging the early-fusion backbone for instance-aware localization. Visual data flow is shown in \textcolor{Green}{green}, coordinates in \textcolor{blue}{blue}, size in \textcolor{orange}{orange}, and segmentation in \textcolor{BlueViolet}{purple.}}
    \label{fig:arch}

\end{figure}

\input{sections/introduction}

\input{sections/arch}

\input{sections/pbench}

\input{sections/training}

\input{sections/results}

\input{sections/ocr}

\input{sections/discussion}

\newpage

\section*{Contributions}

\noindent\textbf{Architecture design:}
Yasser Dahou, Phúc H. Lê Kh\'\abreve{}c, Sanath Narayan, Sofian Chaybouti

\noindent\textbf{Perception training and ablations:}
Sanath Narayan, Yasser Dahou

\noindent\textbf{Distillation:}
Sofian Chaybouti

\noindent\textbf{Data:}
Yasser Dahou, Sanath Narayan, Ankit Singh, Wamiq Reyaz Para, Ngoc Dung Huynh

\noindent\textbf{Evaluation and benchmarks:}
Phúc H. Lê Kh\'\abreve{}c, Yasser Dahou, Sanath Narayan, Ankit Singh, Wamiq Reyaz Para, Ngoc Dung Huynh, Sofian Chaybouti

\noindent\textbf{OCR training and ablations:}
 Ankit Singh, Wamiq Reyaz Para, Ngoc Dung Huynh

\noindent\textbf{Codebase and infrastructure:}
Phúc H. Lê Kh\'\abreve{}c, Sanath Narayan, Ankit Singh, Aviraj Bevli

\noindent\textbf{Integrations:}
Aviraj Bevli, Phúc H. Lê Kh\'\abreve{}c, Sanath Narayan, Ankit Singh, Yasser Dahou

\noindent\textbf{Leadership:}
Yasser Dahou, Hakim Hacid

\section*{Acknowledgments}
We would like to thank Mouad Yagoubi, Ahmed AlHammadi, and Qiyang Zhao for reviewing the technical report. We also thank Mikhail Lubinets for setting up the demonstration infrastructure, and Younes Belkada for his valuable insights regarding llama.cpp and MLX integrations.

\vspace{4em}
\bibliographystyle{iclr2025_conference}
\bibliography{references}

\clearpage
\appendix
\input{sections/appendix}

\end{document}

%% file: sections/introduction.tex
\section{Introduction}

Perception systems for open-vocabulary segmentation and OCR are still largely built around a modular encoder--decoder recipe: a vision backbone produces features, and a separate decoder or late-fusion module turns them into task outputs \citep{carion2020detr,chen2022pix2seq,kamath2021mdetr,kirillov2023segment,sam2}. This design has been effective, but it also encourages adding task-specific mechanisms (modality fusion, query matching, post-processing) that complicate scaling and limit effective feature fusion and learning between vision and text modality \citep{carion2025sam3,minderer2024scalingopenvocabularyobjectdetection,Liu2023GroundingDM,li2022grounded}.

This raises a fundamental question: \emph{do dense grounding systems actually need an encoder--decoder split to ``see'' and to ``predict''?} If not, what should replace the standard late-fusion interface so that language and visual features can interact from the first layer, especially for compositional prompts involving spatial constraints, and relations. What is the right output interface for dense perception, where the number of instances can range from zero to hundreds, without making decoding prohibitively expensive \citep{chen2022pix2seq,chen2022unified,kolesnikov2022uvim}? And finally, how should we benchmark progress once standard referring datasets saturate, in a way that isolates these compositional capabilities and stresses long-context crowded scenes?

In this report, we present Falcon Perception, a unified dense transformer model for dense vision-language perception. Given an image and a prompt, the model must decide \emph{whether the concept is present} and, if so, identify the referred instance(s) and predict pixel masks~\citep{carion2025sam3,kamath2021mdetr}. This setting covers open-vocabulary / promptable segmentation and language-grounded perception~\citep{kirillov2023segment,sam2,li2022grounded,Liu2023GroundingDM,minderer2024scalingopenvocabularyobjectdetection}. While standard approaches typically decouple visual feature extraction (via a vision encoder) from task-specific decoding (via a separate transformer decoder). Falcon Perception unifies these capabilities into a single dense stack. By processing image patches and text tokens in a shared parameter space from the first layer, we avoid late fusion bottlenecks and keep the system simpler for dense grounding.

\noindent Our architecture is built on the hypothesis that a single transformer, equipped with the right inductive biases, can simultaneously learn good visual representations and support autoregressive task generation. To this end, we introduce several key architectural and training innovations:

\begin{enumerate}
    \item \textbf{Unified Dense Transformer with Hybrid Attention Mask:} We replace the encoder--decoder split with a single Transformer using a hybrid attention mask: image tokens attend bidirectionally to build global context, while text and task tokens attend causally to the full visual prefix. This induces encoder-like behavior for vision and decoder-like behavior for language within the same weights.

    \item We introduce \textbf{PBench}, a benchmark for compositional prompts and dense long-context evaluation, enabling capability-level analysis beyond saturated referring benchmarks.

    \item \textbf{Chain-of-Perception:} To balance model expressivity with fast inference, we decompose each instance segmentation into a sequence: \texttt{<coord>} $\rightarrow$ \texttt{<size>} $\rightarrow$ \texttt{<seg>}. This ordering forces the model to resolve an instance's position and size as conditioning signal before producing its segmentation mask.
    
    \item \textbf{Specialized heads:} The backbone is shared, but decoding is not: each Chain-of-Perception token has its own lightweight head. This makes coordinate/size prediction structured (via Fourier features) and keeps segmentation fast by generating high-resolution masks in parallel (Section~\ref{sec:heads}).
    
\end{enumerate}

\noindent Finally, we evaluate Falcon Perception on both referring expression segmentation and open-vocabulary segmentation benchmarks, where it compares favorably to state-of-the-art systems. We also extend the same early-fusion recipe to text-heavy vision tasks with Falcon-OCR and show that a compact 300M model can reach strong OCR performance relative to much larger systems. Falcon Perception connects to these lines of work:

\noindent \textbf{Open-vocabulary segmentation and grounding:} The Segment Anything family introduced promptable segmentation with interactive refinement~\citep{kirillov2023segment,sam2}. SAM~3~\citep{carion2025sam3} formalizes \emph{promptable concept segmentation} and introduces SA-Co together with a calibration-aware evaluation protocol (pmF$_1$, IL\_MCC, and cgF$_1$). A key takeaway for open vocabulary is that recognition and localization should be disentangled: the model must learn to say ``absent'' reliably, not only draw masks when confident. In parallel, open-vocabulary detection and phrase grounding methods typically rely on large vision--language pretraining and retain specialized decoding for boxes/masks, \eg, OWLv2~\citep{minderer2024scalingopenvocabularyobjectdetection}, GroundingDINO~\citep{Liu2023GroundingDM}, GLIP~\citep{li2022grounded}, and MDETR~\citep{kamath2021mdetr}. Falcon Perception targets a simpler interface: early fusion of image and text tokens using a shared backbone over a unified sequence modeling objective.

\noindent \textbf{Autoregressive token interfaces for perception:} Pix2Seq~\citep{chen2022pix2seq} shows that detection can be cast as language modeling by discretizing structured outputs and training with maximum likelihood. This provides a clean ``token interface'' to perception, but also highlights a main limitation: autoregressive decoding becomes expensive as output sequences grow, and dense prediction must avoid duplication and drift. Falcon Perception inherits the token-interface idea but is designed around dense instance segmentation solving the efficiency issue by using specialized heads. UViM~\citep{kolesnikov2022uvim} targets unifying multiple dense vision tasks and explicitly addresses the cost of modeling high-dimensional outputs by using a learned guiding code. In parallel, recent VLMs can output boxes/masks or interleave language with masks (\eg, LISA~\citep{lai2024lisa}, GLaMM~\citep{rasheed2024glamm}) and generalist systems (\eg, Gemini~\citep{comanici2025gemini}, Molmo~\citep{molmo}, Qwen3-VL series~\citep{qwen3}) demonstrate broad capabilities. Compact VLM releases such as Moondream2~\citep{moondream2} and Moondream3~\citep{moondream3} focus on: small, deployment friendly vision-language models. Falcon Perception is positioned as a perception first early-fusion design: a single dense stack optimized for dense grounding, with competitive results on segmentation benchmarks and strong OCR performance at small scale.

\vspace{0.5em}
\noindent \textbf{Scope:} Our goal is not to build a general-purpose VLM for open-ended reasoning (\eg, VQA with multi-step logic or long-form captioning). We focus on dense grounding regimes where the primary difficulty is \emph{localization} under open vocabulary: selecting the correct instance given a phrase (attributes, OCR text, etc) and producing an accurate mask. In this setting, early fusion is particularly natural: the model benefits from allowing prompt to condition on the visual tokens and task context throughout the stack, instead of restricting language-vision interaction to a late cross-attention module.

%% file: sections/arch.tex
\section{Architecture}
\label{sec:architecture}

Our architecture unifies visual perception and language understanding into a single transformer backbone $f_\theta$, deviating from the standard encoder-decoder pipeline. We process raw pixel patches and text tokens using the same set of weights, allowing for deep, early fusion where the same parameters train jointly across modalities. Clearly however, applying pure language modeling for dense perception tasks (\eg, generating polygon coordinates token-by-token) is computationally expensive. To resolve this, we adopt a hybrid approach: the backbone autoregressively predicts special object tokens (\eg, \texttt{<seg>}), which then act as queries for lightweight, specialized head. This design preserves the simple interface of a language model while enabling the efficient, parallel generation of dense binary masks.

\subsection{Overview}

\begin{wrapfigure}{r}{0.5\textwidth}
  \centering
  \includegraphics[width=0.5\textwidth]{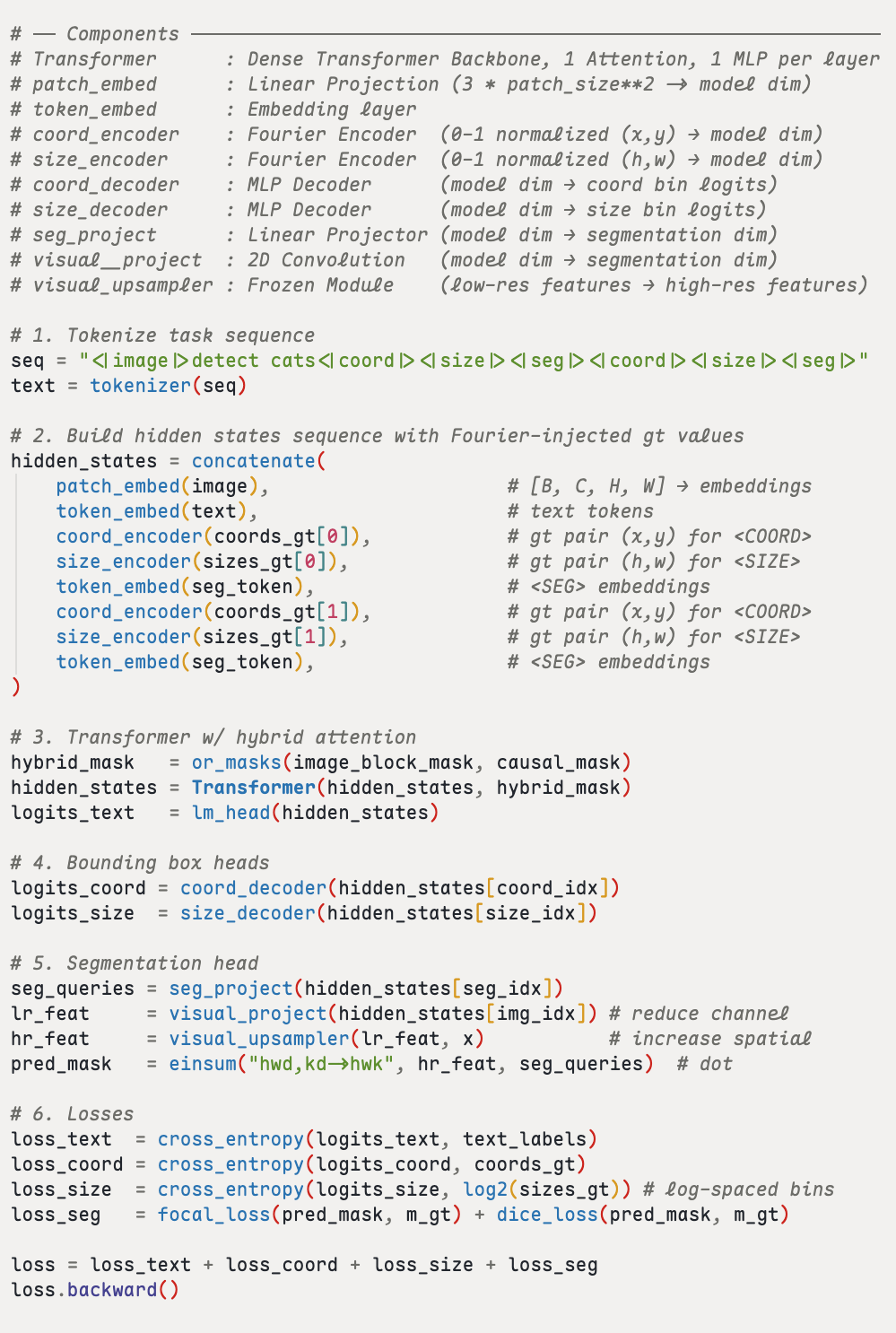}
  \caption{Training forward pass.}
  \label{fig:pseudocode}
\end{wrapfigure}

\noindent We formulate dense perception as a conditional generation task. Given an image $I \in \mathbb{R}^{H \times W \times 3}$ and a text prompt $\mathcal{P} = \{t_1, \dots, t_L\}$, the model must predict a set of $K$ objects $\{O_k\}_{k=1}^K$. Each object is defined by a tuple $O_k = (c_k, s_k, m_k)$ representing its center coordinates, size, and binary segmentation mask. To solve this, we serialize the output into a structured sequence, imposing a ``Chain-of-Perception'' order: $c_k \rightarrow s_k \rightarrow m_k$. This coarse-to-fine curriculum stabilizes training by forcing the model to resolve spatial ambiguity before committing to pixel-level details. The resulting sequence format is \texttt{[Image] [Text] <coord> <size> <seg> $\cdots$ <eos>}, as illustrated in Figure \ref{fig:arch} and \ref{fig:pseudocode}.

\noindent \textbf{Input Representation:} To process these heterogeneous inputs, we flatten the image into $N$ patches and project them into visual embeddings $V \in \mathbb{R}^{N \times d}$. Similarly, the text prompt is mapped to $L$ text embeddings $T \in \mathbb{R}^{L \times d}$. The unified sequence $X \in \mathbb{R}^{S \times d}$ is formed by concatenating these with the task tokens for the $K$ objects, where $S = N + L + 3K$ is the total sequence length and $d$ is the hidden dimension:
\begin{equation}
X = [\underbrace{v_1, \dots, v_N}_{\text{Visual Embeddings}}, \underbrace{t_1, \dots, t_L}_{\text{Text Embeddings}}, \underbrace{e_{\texttt{<coord>}}, e_{\texttt{<size>}}, e_{\texttt{<seg>}}, \dots}_{\text{Task Embeddings}}]
\end{equation}

\noindent The model processes this sequence autoregressively. The backbone $f_\theta$ is a standard dense Transformer with $L$ layers. Processing this sequence requires a specialized attention strategy. Standard causal masking is suboptimal for visual data, as image patches are inherently 2D thus benefit from bidirectional context. We therefore employ a hybrid masking strategy: image tokens attend to all other image tokens (bidirectional), while text and task tokens attend to all image tokens but only to preceding text/task tokens (causal). This allows a single set of weights to act simultaneously as a bidirectional vision encoder and an autoregressive language decoder. Unlike Prefix-LM masking \citep{Beyer2024PaliGemmaAV}, we do not allow prefix text tokens to attend to each other bidirectionally.

\subsection{Specialized Heads}\label{sec:heads}
While keeping a single Transformer backbone to maximize interaction between tokens of all modalities, we employ specialized encoders and decoders to inject extra inductive biases for spatial task at the input and output layers.

\noindent \textbf{Coordinate \& Size Encoders using Fourier Features:} Standard tokenization of coordinates (\eg, 1000 bins) suffers from limited precision and \textit{spectral bias}, the tendency of neural networks to prioritize learning low-frequency functions, resulting in imprecise spatial grounding.  To overcome this, we adopt a Fourier Feature mapping strategy \citep{FourierFeat}.  We map input coordinates $\mathbf{c} \in [0,1]^2$ to a higher-dimensional feature space using a random gaussian matrix $\mathbf{B} \in \mathbb{R}^{d/2 \times 2}$:
\begin{equation}
\gamma(\mathbf{c}) = [\cos(2\pi \mathbf{B}\mathbf{c}), \sin(2\pi \mathbf{B}\mathbf{c})]
\end{equation}
This projection transforms the coordinate learning problem from a low-dimensional regression (which MLPs struggle with) into a high-dimensional pattern matching task over a rich frequency spectrum. The scale $\sigma$ (set to 10.0) controls the bandwidth of the kernel, tuning the model's sensitivity to fine-grained spatial details. These continuous embeddings are then linearly projected to the transformer dimension and injected directly into the input stream, replacing the static \texttt{<coord>} token. For decoding, we use a simple MLP to project the hidden state into $2 \times N_{bins}$ logits. \textit{Motivation:} predicting coordinates and size \textit{before} segmentation acts as a strong conditioning signal. It forces the model to explicitly resolve instance ambiguity (\eg, "which cat?") and spatial extent. This offloads the localization from the segmentation head, allowing the subsequent \texttt{<seg>} token to focus purely on fine-grained pixel refinement rather than instance discrimination, eliminating the need for heavy mask encoders.

\noindent \textbf{Segmentation (Upsampling + Dot Product):} Generating a high-resolution mask $m \in \mathbb{R}^{H \times W}$ from a single token is challenging. Standard approaches, such as Mask2Former \citep{cheng2022masked}, require complex Hungarian matching to resolve instance ambiguity. We simplify this by leveraging our unified architecture. Since the \texttt{<seg>} token is generated after the coordinates and size, the model has already resolved the object identity. Furthermore, due to our early fusion, its hidden state $h_{seg}$ has direct access to the global visual context. Thus, we can generate the mask via a simple dot product, without separate mask queries.

\noindent To recover fine-grained spatial details, we employ a content-aware upsampler $\mathcal{U}: (V_{out}, I) \to V^*$ \citep{wimmer2025anyup}. This module formulates upsampling as a cross-attention operation where the query is derived from the high-resolution input image $I \in \mathbb{R}^{H \times W \times 3}$ (to guide boundaries), and the key/value are derived from the backbone's output visual features $V_{out} \in \mathbb{R}^{N \times d}$. This allows the model to "paint" semantic features onto the high-resolution pixel grid, producing an enhanced feature map $V^* \in \mathbb{R}^{H \times W \times d}$. The final mask $\hat{m}$ is generated by computing the dot product between these high-fidelity features and the projected hidden state of the segmentation token $h_{seg}$:
\begin{equation}
\hat{m} = \sigma(V^* \cdot \text{Proj}(h_{seg})^T)
\end{equation}

\noindent This design is simple and intuitive, our approach offloads the "localization" to the dense backbone. Because $h_{seg}$ is generated autoregressively and has access to the full early-fusion context $V_{out}$, it already contains the necessary instance-specific information. Consequently, we do not need a separate query learning stage, Hungarian matching, or point-based sampling strategies. The segmentation task is reduced to a straightforward projection, making the architecture easier to implement and scale. 

\subsection{Important Details}

\noindent Beyond the core backbone and specialized heads, several key architectural decisions are needed for handling dense spatial information effectively. We detail these choices below.

\noindent \textbf{3D Rotary Positional Embeddings (Sequence + Space):} To preserve the spatial structure of the input within the flattened sequence $X$, we inject positional information directly into the attention mechanism. Standard 1D RoPE treats the image as a linear sequence, destroying the 2D grid relationships critical for dense prediction. We instead decompose the head dimension $d_{head}$ into two distinct subspaces: a sequential component and a spatial component.

The first half of the head dimension ($d_{head}/2$) encodes the 1D sequence index $t$ using standard RoPE, capturing the causal order of the text tokens and the relative position of patches in the flattened sequence. The second half ($d_{head}/2$) encodes the 2D grid position $\mathbf{p} = (x, y)$ using \textbf{Golden Gate RoPE (GGRoPE)} \citep{xiong2025ndrope}. Unlike standard Axial RoPE, which restricts attention to orthogonal $x$ and $y$ axes, GGRoPE rotates each dimension pair $j$ based on the projection of the spatial position $\mathbf{p}$ onto a unique direction vector $\mathbf{u}_j$:
\begin{equation}
\theta_j = \omega_j (\mathbf{p} \cdot \mathbf{u}_j)
\end{equation}
where the direction vectors $\{\mathbf{u}_j\}$ are sampled uniformly from the unit circle using the Golden Ratio $\phi$. This mechanism allows the attention heads to attend to relative positions along arbitrary 2D angles, producing isotropic attention maps that are robust to rotation and aspect ratio variations. Furthermore, this spatial rotation is applied sparsely only to valid visual tokens, ensuring computational efficiency and preventing interference with text embeddings does not have spatial coordinates.

\noindent \textbf{Native Resolution Handling:} To avoid distorting aspect ratios and degrading performance on small objects. We process images at their native resolution and aspect ratio (up to a maximum token budget). This introduces high variance in sequence lengths: a $256 \times 256$ image yields $\sim 256$ patches, while a $768 \times 768$ image yields $\sim 2,304$. To handle the resulting variance in token counts efficiently, we employ a scatter-and-pack strategy: we patchify the image and discard all padding tokens \textit{before} the transformer, keeping only the valid visual patches. These variable-length sequences are then packed into a single sequence of fixed length $C_{max}$. We utilize FlexAttention to implement the appropriate masking, ensuring that self-attention is restricted within each image sample's boundaries. This maximizes GPU utilization by processing only valid data, but results in a variable number of images per batch across ranks. We address the resulting optimization instability via a global loss balancing strategy \citep{chaybouti2025amoe}.

%% file: sections/pbench.tex
\section{PBench: Perception Benchmark}
\label{sec:benchmark}
\begin{figure}[t]
  \centering
  \includegraphics[width=1.0\textwidth, trim=4.5cm 4.5cm 4.5cm 4.5cm, clip]{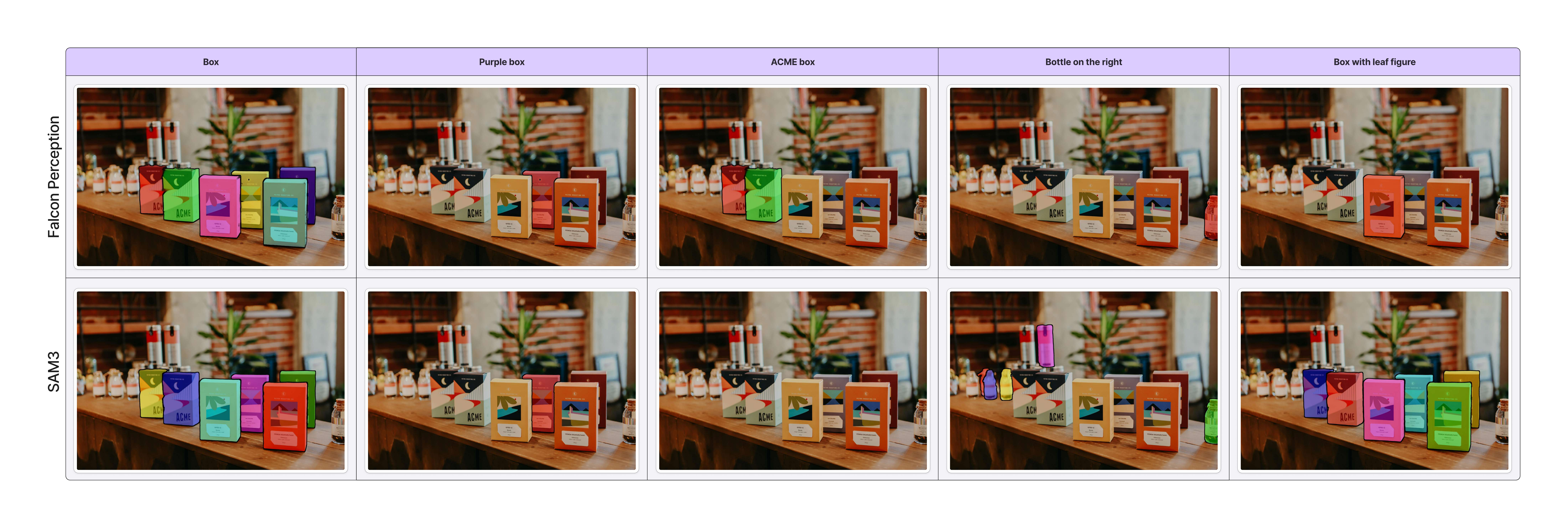}
\caption{\textbf{Prompt progression across complexity levels.} We keep the image fixed and vary the text prompt to isolate the capabilities targeted by Levels~0--4 (Table~\ref{tab:complexity_levels}). Each panel shows the predicted masks for the same scene under progressively more specific prompts: from generic object classes (Level~0, \eg ``box''), to attribute binding (Level~1, \eg ``purple box''), to OCR-based identification (Level~2, \eg ``ACME box''), and then spatial/layout constraints (Level~3, \eg ``bottle on the right'') and fine-grained relational descriptions (Level~4). Falcon Perception remains stable as prompts become more compositional and text-dependent, while SAM~3 starts to fail on OCR-driven queries and higher-level constraints, either returning no masks or segmenting visually plausible but semantically wrong instances.}
\label{fig:levels_progression}
\end{figure}

Before diving into the training dynamics, we work on measuring progress for referring expression segmentation. Current benchmarks, such as RefCOCO, RefCOCO+~\citep{yu2016modeling}, and G-Ref~\citep{kazemzadeh2014referitgame}, have been important driving progress in the field. However, they suffer from two critical limitations. First, performance saturation: state-of-the-art models now routinely achieve $>80 \%$ acuracy, making it difficult to rank models. Second, and more importantly, they lack \textit{semantic granularity}. A \textit{hard} sample in these datasets conflates various challenges: complex prompt, spatial ambiguity, or world knowledge into a single metric. A model might fail because it cannot detect small objects, or because it does not understand a complex expression, but the benchmark score provides no signal to differentiate these failure modes.

\noindent To rigorously evaluate Falcon Perception and understand its capabilities, we introduce a hierarchical evaluation benchmark. Designed to disentangle the capabilities required for open-vocabulary segmentation, organizing samples into five distinct levels of complexity. This allows us to track capabilities progress effectively. 

\subsection{Complexity Levels}
\label{sec:complexity-levels}

Each sample is assigned a \emph{single} level label by construction, and prompts are intentionally written to target one dominant capability while avoiding cross-level cues (\eg, OCR prompts avoid spatial qualifiers such as ``left/right'', and spatial prompts avoid in-image text disambiguators). This yields a capability profile rather than a single scalar score. Please refer to Table \ref{tab:complexity_levels} for the levels definition and examples.

\begin{table}[t]
\centering
\footnotesize
\caption{Complexity-level taxonomy for our internal referring expression segmentation benchmark. Each sample is assigned a single level label by construction, enabling per-level capability profiling rather than a saturated aggregate score.\vspace{-0.3cm}}
\label{tab:complexity_levels}
\setlength{\tabcolsep}{5pt}
\renewcommand{\arraystretch}{1.2}
\begin{tabularx}{\textwidth}{p{0.05\textwidth} p{0.20\textwidth} >{\raggedright\arraybackslash}X p{0.21\textwidth}}
\toprule
\textbf{Level} & \textbf{Primary capability} & \textbf{Prompt ingredients \newline(single-skill by construction)} & \textbf{Examples} \\
\midrule
0 & General object classes &
\textbf{Rules:} whole objects only; machine-detectable categories; bbox-drawable; clear boundaries; no abstract concepts. &
\texttt{car}\newline\texttt{person}\newline\texttt{tree} \\
\midrule
1 & Fine-grained \newline attributes \& subtypes &
\textbf{Attributes:} color; size; shape; material.\newline
\textbf{State:} old/new; clean/dirty; broken/cracked; worn/well-maintained.\newline
\textbf{Subtype:} sedan/SUV/pickup; oak/palm/rose; tabby/golden retriever.\newline
\textbf{Functional:} open/closed; on/off; active/inactive.\newline
\textbf{Components:} sunglasses; cracked windshield; neon sign; petals. &
\texttt{red car}\newline\newline\texttt{broken fence}\newline\newline\texttt{open door} \\
\midrule
2 & Text as identifier (OCR) &
\textbf{Brand/product:} Coca-Cola; Nike; iPhone.\newline
\textbf{Variant:} Diet vs Regular; gluten-free vs regular.\newline
\textbf{Store/location:} Starbucks; McDonald's; Barnes \& Noble.\newline
\textbf{Signage/tags:} sale tag; prescription label; emergency exit. &
\texttt{Diet Coke bottle}\newline\newline\texttt{Nike shoes}\newline\newline\texttt{Emergency exit door} \\
\midrule
3 & Spatial relationships \& layout &
\textbf{Relative:} left/right/behind; above/below.\newline
\textbf{Depth:} foreground/middleground/background.\newline
\textbf{Ordering:} first/third/last (\eg, third from left).\newline
\textbf{Grouping:} in a row/line.\newline
\textbf{Containment:} inside/outside; boundary interactions. &
\texttt{car on the left}\newline\newline\texttt{third window from left}\newline\newline\texttt{person in the foreground} \\
\midrule
4 & Relationships \& interactions &
\textbf{Actions:} holding; wearing; pulling.\newline
\textbf{Functional links:} key for door; charger for phone.\newline
\textbf{Comparisons:} tallest/smallest/most damaged.\newline
\textbf{Physical:} resting on; hanging on; attached to.\newline
\textbf{Groupings:} dining set; lock and key; tools for repair. &
\texttt{person holding umbrella}\newline\newline\texttt{car pulling trailer}\newline\newline\texttt{tallest building in the row} \\
\bottomrule
\vspace{-0.2cm}
\end{tabularx}
\end{table}

\noindent \textbf{How to interpret the levels:} The levels are not a subjective notion of ``difficulty''; rather, each level isolates a dominant capability. Level~0 and Level~1 primarily probe open-vocabulary recognition and attribute binding. Level~2 isolates OCR-aware disambiguation where in-image text is the deciding signal between near identical instances. Level~3 isolates spatial grounding and layout understanding, testing whether the model forms a coherent 2D scene map. Level~4 isolates relational grounding where the target is defined through interaction rather than appearance. We therefore report per-level performance in addition to an overall average, yielding a capability profile that reveals \emph{where} a model fails (\eg, collapsing at OCR or relations). As shown in Figure \ref{fig:levels_progression}, for the same image, the different prompts highlight different capabilities from simply asking about boxes, to attributing with a color, then using the text written to retrieve the correct ones, followed by using spatial instructions, and finally using a finegrained description.\\

\subsection{Crowdedness and Long-Context Stress Test}
\label{sec:dense_stress}

In addition to semantic separation (Levels 0-4), we evaluate on \emph{dense} image with many instance per prompt. Crowded scenes stress long-context generation because the model must emit a long sequence of object queries and masks, maintain a stable ``object not found'' prior, and avoid duplication and drift as the number of instances $K$ increases. This regime is poorly characterized by existing referring benchmarks, which typically contain relatively few instances per image. We therefore report performance as a function of $K$ (\eg, $K \le 150$ in-distribution, and long-context settings up to $K \approx 600$). This isolates whether a model fails due to semantic understanding (captured by the complexity levels) or due to long-context dynamics (crowdedness), and directly measures the stability of our autoregressive Chain-of-Perception interface in dense environments. It is worth mentioning that this category still register under Level~-1 with simple nouns, and the complexity comes from the instance count.

\subsection{Creation and Statistics}
\label{sec:benchmark_construction}

\noindent \textbf{Construction:} The benchmark was created internally by team members who contributed to training Falcon Perception. Prompts and target masks were curated with domain expertise to ensure both high quality and clean semantic separation across the complexity levels. In particular, each sample is assigned a \emph{single} level label by construction: prompts are written to isolate one dominant capability (\eg, OCR disambiguation without spatial cues), enabling diagnostic evaluation rather than conflating multiple failure modes. In addition, we curate a separate crowdedness subset to stress long-context generation and dense instance segmentation.

\noindent \textbf{Size:} The benchmark contains approximately $\sim$5k samples  across Levels~0--4, and an additional $\sim$400 samples dedicated to crowdedness stress testing.

%% file: sections/training.tex
\section{Training details}
\label{sec:training}
\noindent We adopt in two stages to effectively learn dense perception capabilities. In the first stage, \textbf{Multi-Teacher Distillation}, we initialize the model weights by distilling knowledge from strong vision teachers (DINOv3 and SigLIP2) into our unified dense backbone. This provides a robust initialization for visual features. In the second stage, \textbf{Perception training}, we train the model on approximately 685 GT of multimodal data using our specialized "Chain-of-Perception" format (\texttt{<coord><size><seg>}). This stage fine-tunes the backbone and heads to perform autoregressive dense prediction tasks. We detail each stage and our design choices below.

\subsection{Multi-Teacher Distillation for Weights Initialization\label{sec:mtd}}

To help train our early-fusion perception models, we use the multi-teacher distillation pipeline described by~\citep{chaybouti2025amoe} to initialize the weights of the models. The motivation of this strategy is to benefit from the distinct strengths of diverse vision backbones: we use the ViT-H variant of DINOv3~\citep{simeoni2025dinov3} to inherit its strong local features, critical for our segmentation head; and SigLIP2-So400m~\citep{tschannen2025siglip} to inject knowledge and language-aligned features for open-vocabulary expression understanding.

\noindent We use this initialization on the two model scales: the 22-layer (300M parameters) and the 28-layer (600M parameters) dense models. The distillation is performed over the OpenLVD-200m~\citep{chaybouti2025amoe} dataset, supplemented by 9 million high-resolution scrapped images, 11 million images from the SAM~\citep{kirillov2023segment} dataset, and 5 million documents. Training consists of a single multi-resolution stage up to $1024 \times 1024$ pixels, spanning approximately \num{200}k steps. We use the Muon~\citep{jordan2024muon} optimizer. The training is distributed across four 8-A100 GPU nodes, employing sequence packing with a local batch size of 6 and a maximum sequence length of \num{4096} tokens to maximize computational throughput.

\noindent \textbf{Evaluation.} To assess the quality of the learned representations, we evaluate the models on standard vision benchmarks. For linear probing semantic segmentation tasks, we conduct a learning rate search and report the best mIoU obtained. As shown in Table~\ref{tab:distillation_results}, the multi-teacher distillation provides a strong foundation for both general classification and dense spatial tasks.

\begin{table}[t]
\centering
\caption{Performance of the two dense models initialized via multi-teacher distillation. All evaluations are run at $512\times512$ maximum number of pixels.\vspace{-0.2cm}}
\setlength{\tabcolsep}{6pt}
\adjustbox{width=\textwidth}{
\begin{tabular}{l c c c c c}
\toprule
\textbf{Model}  & \textbf{ImageNet-1k (zero-shot)} & \textbf{kNN} & \textbf{Cityscapes} & \textbf{Pascal VOC} & \textbf{ADE20K} \\
\midrule
22-layer & 70.84\% & 82.58\% & 60.77\% & 83.81\% & 48.10\% \\
28-layer & 74.25\% & 84.00\% & 62.72\% & 85.11\% & 49.10\% \\
\bottomrule
\end{tabular}
}
\label{tab:distillation_results}
\end{table}

\subsection{Data Curation}

\noindent We construct a large-scale, high-quality dataset of 54M images with 195M positive expressions, and 488M negatives ones, with a total of 570M masks through a multi-stage pipeline: 

\noindent \textit{Clustering $\to$ VLM Listing $\to$ Negative Mining $\to$ Ensemble Consensus $\to$ Human Verification}. This process ensures diversity of expressions and images, semantic granularity, and high-precision annotations.

\noindent \textbf{Seed Data via Hierarchical Clustering:}
Starting from a large pool of web-scraped images, we employ the hierarchical clustering and sampling technique to mitigate long-tail biases~\citep{vo2024automatic}. Specifically, we embed images using a DINOv3 ViT-B encoder and perform a 4-level hierarchical clustering (20M $\to$ 500k $\to$ 50k $\to$ 20k centroids). We then sample a balanced subset of \textit{54M diverse seed images}, ensuring broad and uniform concept coverage.

\noindent \textbf{Listing via VLM:}
We prompt a powerful  VLM to generate dense object listings for each image. Crucially, we instruct the VLM to categorize outputs according to our PBench complexity levels (Section~\ref{sec:complexity-levels}). The resulting distribution is heavily weighted towards foundational capabilities: 60\% of samples target Levels 0-1 (Objects \& Attributes), with the remaining 40\% spread across Levels 2-4 (OCR, Spatial, Relations).

\noindent \textbf{Negative Mining:}
To robustify the model against hallucinations, we generate hard negative queries for each image. We feed the image and its positive concepts back into the VLM to generate three types of negatives: (i) \textit{Semantic Negatives:} Concepts in the same category but absent (\eg, "sushi" vs. "pizza"); (ii) \textit{Visual Confounders:} Objects that look similar to present instances; and (iii) \textit{Hard Semantic Negatives:} Fine-grained distinctions (\eg, specific cat breeds).

\noindent \textbf{Annotation via Ensemble Consensus:}
We employ a multi-model agreement mechanism to generate high-quality synthetic  labels.
\begin{itemize}
    \item \textbf{Levels 0-1:} We use an ensemble of SAM 3, Qwen3-VL-30B, and Qwen3-VL-8B. For each sample, we randomly select a pair of models. A sample is accepted only if all matched the pair aligns with an IoU $> 0.8$.
    \item \textbf{Levels 2-4:} We use Qwen3-VL-30B, Moondream3, and Qwen3-VL-8B. Similarly, we require agreement between a randomly sampled pair. Since these are detection models, we convert agreed bounding boxes to segmentation masks using SAM 2.
\end{itemize}

The agreement algorithm uses Hungarian matching to align predictions and strictly enforces both count matching and high spatial overlap.

\noindent \textbf{Human Verification Loop:}
Disagreements are not discarded. We route the subset of non-agreed samples (approx. 4M) to human annotators for manual verification (Accept/Reject). Accepted samples are incorporated into the training mix, while rejected samples are then sent for full manual re-annotation from scratch. This loop recovers hard samples that confuse the automated ensemble, preventing the dataset from being biased towards "easy" consensus examples, and helps mine hard samples where most models fail.

\noindent \textbf{Public Datasets:} We supplement our curated data with: OpenImages~\citep{kuznetsova2020open}, Objects365~\citep{shao2019objects365}, WCS camera traps~\citep{lila_bc_wcscameratraps}, marine datasets~\citep{jansen2022underwater,saleh2020realistic,lila_bc_river_herring}, iSAID~\citep{waqas2019isaid}, Fashionpedia~\citep{jia2020fashionpedia}, iNaturalist-2017~\citep{van2017inaturalist} and VisDrone-Det~\citep{zhu2021detection}. For datasets with bounding boxes, we convert the annotations to masks using SAM-2~\citep{sam2}.

\noindent \textbf{Text only data:}
To enhance the language modeling capabilities of our unified backbone, we mix in high-quality text-only data during training. We utilize publicly available pre-training corpora and SFT datasets, such as Fineweb \citep{penedo2024fineweb}, to ensure the model develops general language abilities alongside the dense perception task.

\subsection{Design choices}
The aimed capabilities  introduce design challenges. We detail the key sweeps conducted to finalize the design. Settings for each experiment were kept constant across variants. For efficiency, we perform ablations only on the detection task, except for feature regularization, where segmentation is used to better analyze retention of high-quality image features. 

\noindent \textbf{Sequence format:} We serialize the task into a unified sequence format: 

\texttt{<image>expr$_1$<present><coord><size><seg>$\cdots$<eoq>expr$_2$<absent><eoq> $\cdots$ <eos>}. The end of query block is marked by \texttt{<eoq>}. We introduce explicit \texttt{<present>} and \texttt{<absent>} tokens immediately following the text expression to impose a strong inductive bias: the model must explicitly commit to a binary decision about the object's existence \textit{before} attempting any localization. 

\noindent \textbf{Configuration:} We train a 600M parameter model for Falcon Perception. Images are processed at a maximum resolution of $1024 \times 1024$ with a patch size of 16, ensuring high-fidelity visual encoding. This base configuration remains consistent across all stages, with specific adaptations to learning rates and masking strategies as detailed below.

\noindent \textbf{(\textit{i}) Loss Balancing across Ranks:}
In our multimodal training with packed sequences, the number of valid tokens varies significantly across ranks due to varying sizes of the images, number of expressions and instances per packed sample. We identified that standard FSDP gradient averaging biases the loss towards ranks with fewer tokens. We implemented a global normalization strategy where the loss is first aggregated across the ranks and then scaled by the \textit{global} average number of tokens ($N_{total}/R$) rather than the local rank count. Details in \ref{sec:loss}.

\begin{figure}[t]
\centering
\includegraphics[width=0.33\linewidth]{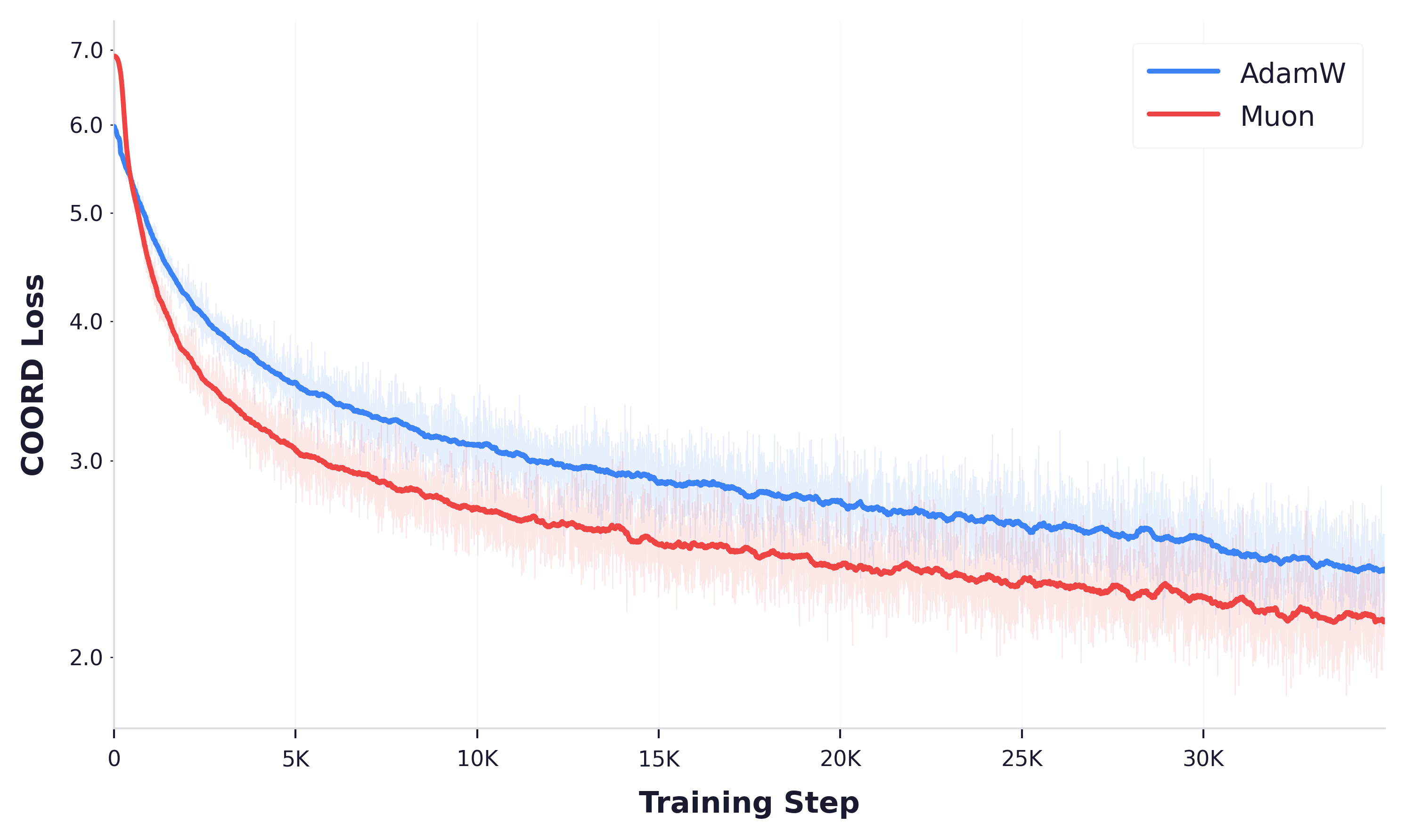}
\includegraphics[width=0.32\linewidth]{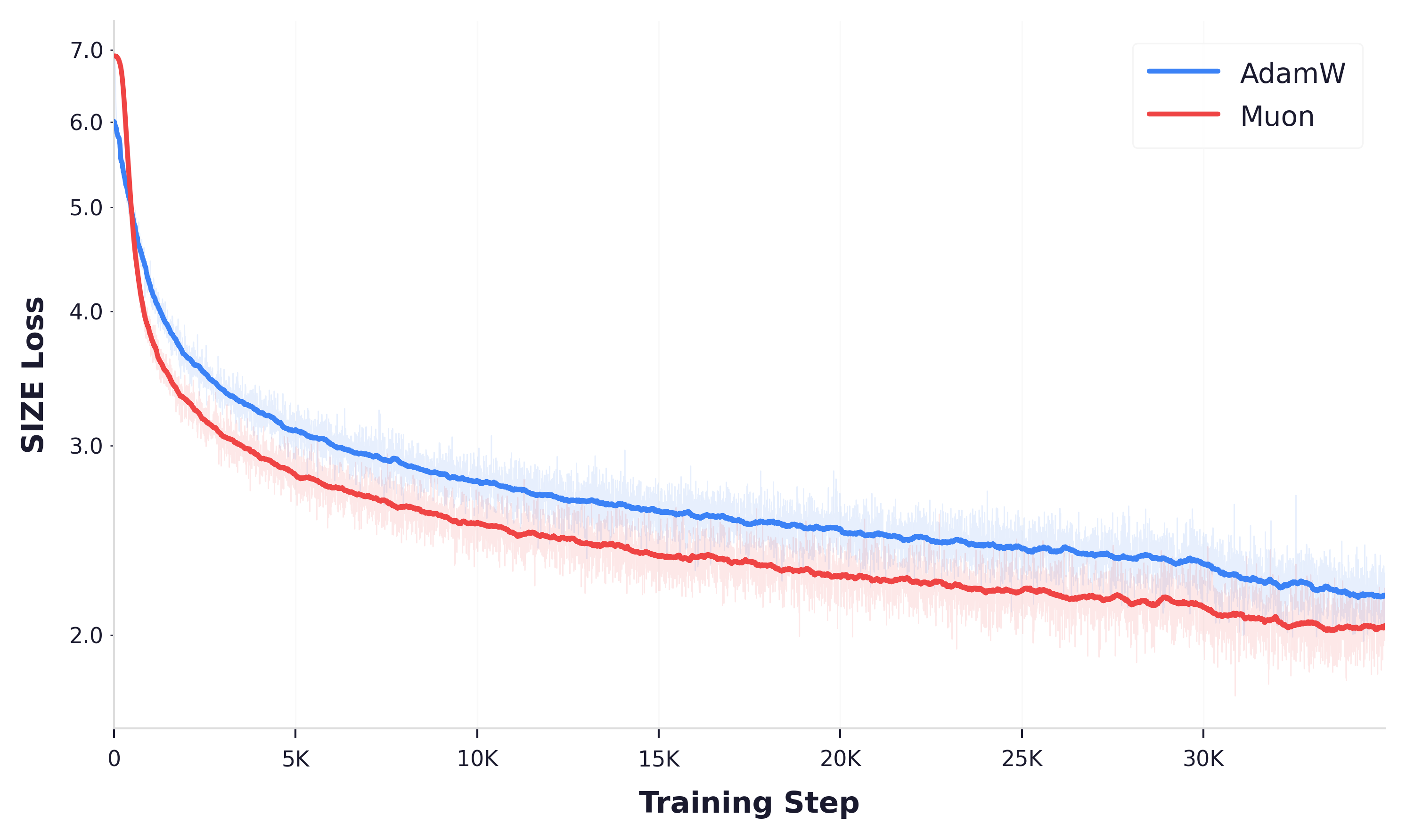}
\includegraphics[width=0.32\linewidth]{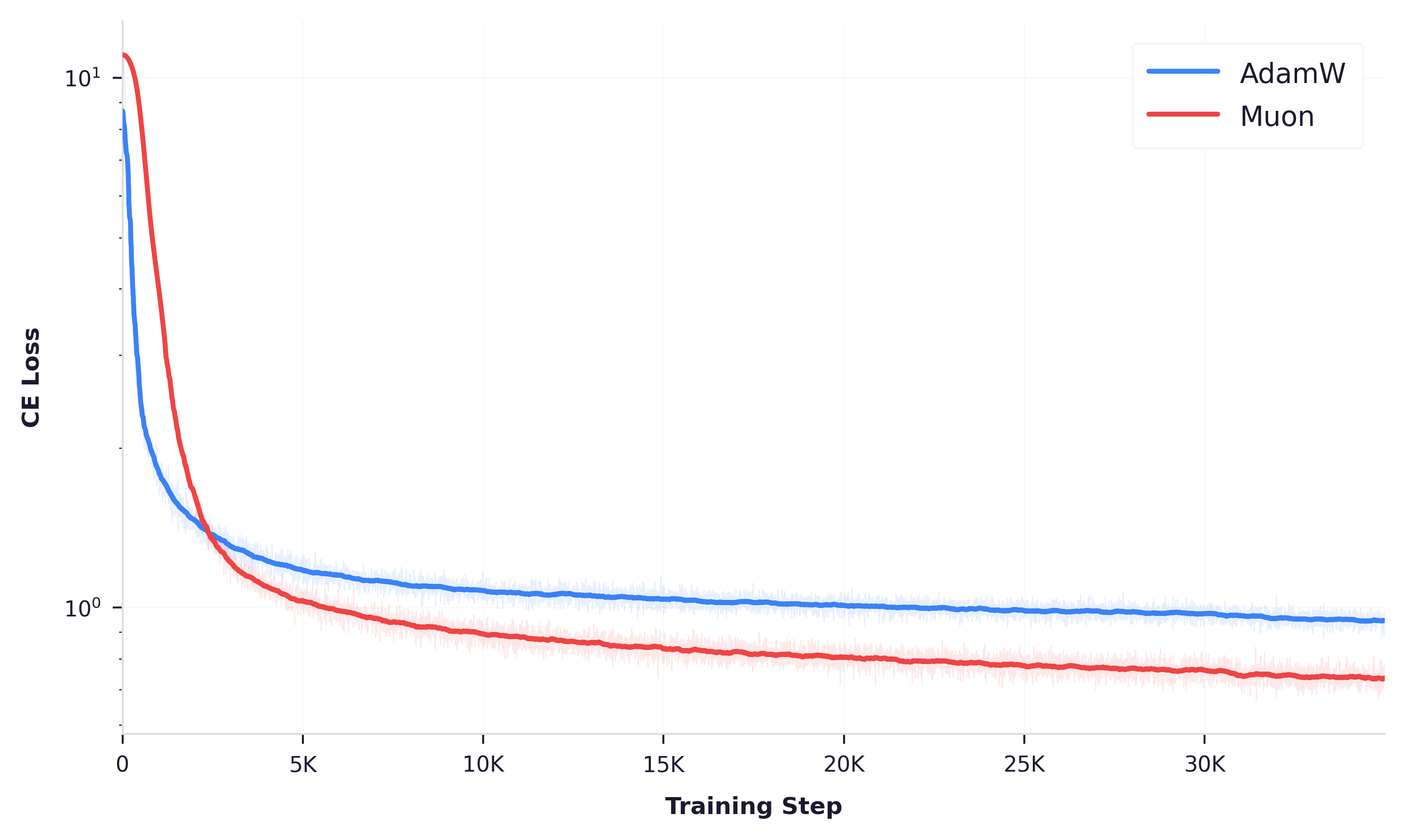}
\caption{Muon optimizer results in lower training losses for coord, size and cross-entropy, in comparison to AdamW, leading to improved performance (Tab.~\ref{tab:optim}).\label{fig:optim}}
\end{figure}

\noindent \textbf{(\textit{ii})  Optimizer Choice - Muon \vs AdamW:}
We observed that the specialized heads (Coordinates, Size, Segmentation) often lagged behind the language backbone during training. We compared standard \textit{AdamW} against \textit{Muon} (a momentum-based optimizer) for these heads to balance learning rates between the pre-trained backbone and initialized perception modules. Fig.~\ref{fig:optim} shows the training loss comparison between Muon and AdamW optimizer for the language, coord and size heads. We observe that employing the Muon optimizer leads to lower training losses across the different heads, yielding improved performance on both PBench and SaCo benchmarks in terms of macroF1 (Tab.~\ref{tab:optim}). 

\noindent \textbf{(\textit{iii}) Feature Regularization (Gram Loss):}
We introduce the Gram loss, similar to DINOv3~\citep{simeoni2025dinov3}, as a regularization to maintain the structural integrity of the visual feature space. This enforces that the correlation matrix (Gram matrix) of the student's patch features matches that of a frozen teacher model. From the ablation experiment in Tab.~\ref{tab:gram}, we observe that training with the Gram loss indeed results in better performance on both benchmarks since it enables better image feature retention that is required for the segmenting the instances, compared to learning the task without the Gram regularization.

\noindent \textbf{(\textit{iv}) Attention and Causality:}
We investigated the attention mechanism between queries.
\textit{Full AR \vs No Inter-query Attn:} We compared full causal attention (where queries attend to the image features in addition to the previous queries and their corresponding instances) against a masked approach, where queries attend to the image but not to each other, assessing the trade-off between relational reasoning and computational efficiency. Full AR enables in-context learning among the queries, where the model can learn to skip predicting the positions occupied by instances from previous queries. In contrast, the inter-query masking attention forces the model to focus on the image content to predict the instances for a particular query. Although the ablation in Tab.~\ref{tab:querymask} shows a marginal advantage for the query masking variant, since both the strategies rely on complimentary information for prediction, we employ them both in different stages of our model training, as detailed in Sec.~\ref{sec:recipe}. 
\begin{table}[t!]
\begin{minipage}{0.32\textwidth}
    \centering
    \setlength{\tabcolsep}{4pt}
    \captionof{table}{AdamW \vs Muon\vspace{-0.2cm}}
    \label{tab:optim}
    \adjustbox{width=0.95\columnwidth}{
    \begin{tabular}{c|cc}
        \toprule
         \textbf{Optimizer} & \textbf{PBench-\textit{Det}} & \textbf{SaCo-\textit{Det}} \\
         \midrule
         AdamW & 56.4 & 49.0 \\
         Muon & 57.7 & 53.8 \\
        \bottomrule
        
    \end{tabular}
    }

\end{minipage}
\begin{minipage}{0.32\textwidth}
    \centering
    \setlength{\tabcolsep}{4pt}
    \captionof{table}{Impact of Gram Loss\vspace{-0.2cm}}
    \label{tab:gram}
    \adjustbox{width=0.95\columnwidth}{
    \begin{tabular}{c|cc}
        \toprule
         \textbf{Feature Reg.} & \textbf{PBench-\textit{Seg}} & \textbf{SaCo-\textit{Seg}} \\
         \midrule
         No Gram & 52.7 & 51.1 \\
         With Gram & 53.8 & 52.6 \\
        \bottomrule
    \end{tabular}
    }
\end{minipage}
\begin{minipage}{0.33\textwidth}
    \centering
    \setlength{\tabcolsep}{4pt}
    \captionof{table}{Inter-query Masking\vspace{-0.2cm}\label{tab:querymask}}
    \adjustbox{width=0.95\columnwidth}{
    \begin{tabular}{c|cc}
        \toprule
         \textbf{Ordering} & \textbf{PBench-\textit{Det}} & \textbf{SaCo-\textit{Det}} \\
         \midrule
         Full AR & 53.2 & 53.1 \\
         Query Masking & 54.2 & 53.3 \\
        \bottomrule
    \end{tabular}
    }
    
\end{minipage}

\end{table}

\begin{figure}[h]
\centering

\begin{minipage}{0.6\textwidth}
    \centering
    \includegraphics[width=\linewidth]{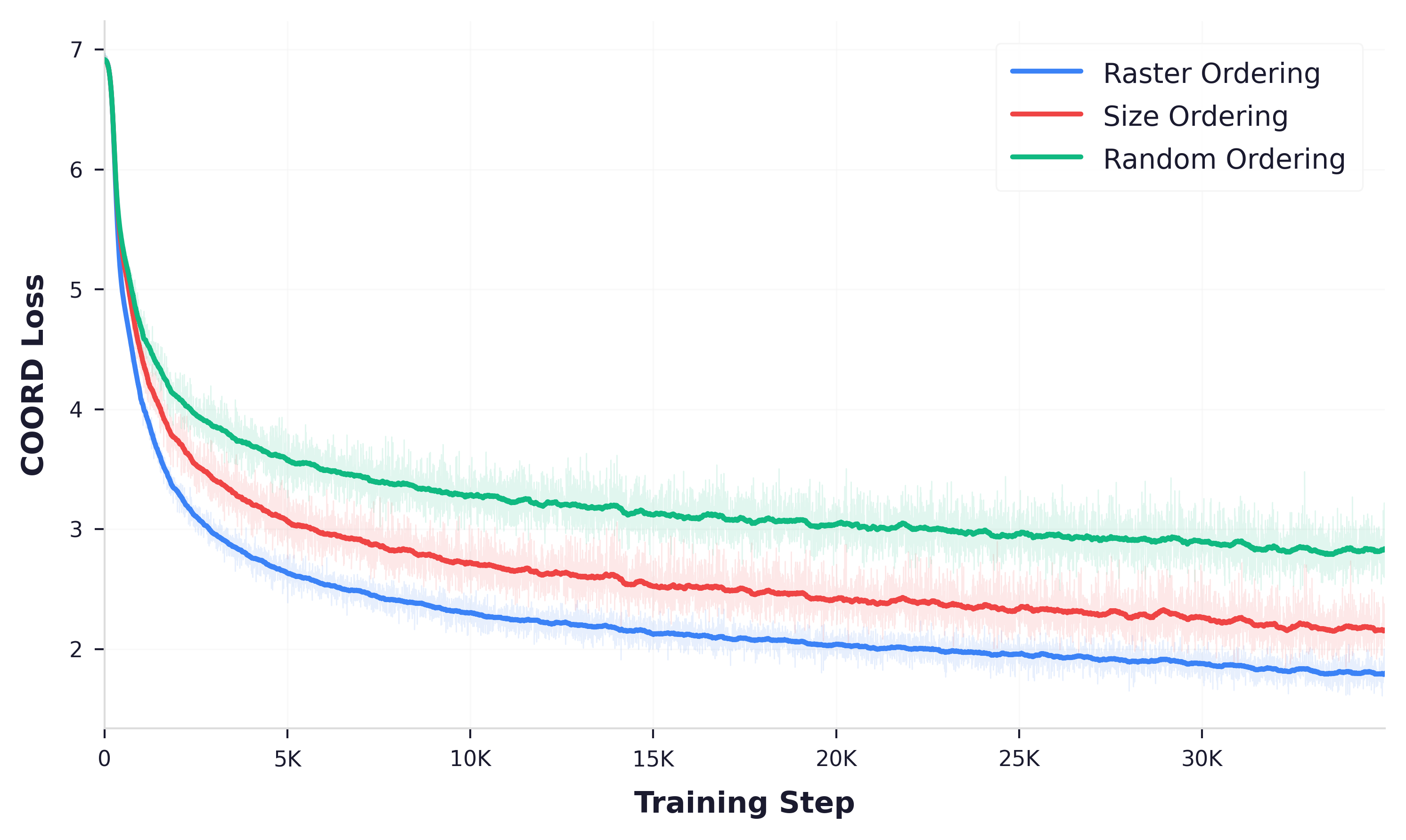}
\end{minipage}
\hfill
\begin{minipage}{0.39\textwidth}
    \centering
    \setlength{\tabcolsep}{5pt}
    \adjustbox{width=0.9\columnwidth}{
    \begin{tabular}{c|cc}
        \toprule
         \textbf{Ordering} & \textbf{PBench-\textit{Det}} & \textbf{SaCo-\textit{Det}} \\
         \midrule
         Random & 52.2 & 46.3 \\
         Size & 57.7 & 53.8 \\
         Raster & 59.3 & 56.2 \\
        \bottomrule
        
    \end{tabular}
    }
   
\end{minipage}

\caption{Raster ordering of instances results in lower training loss and better performance on both benchmarks, compared to random and size orderings.\label{fig:ordering}}

\end{figure}

\noindent \textbf{(\textit{v}) Mask Ordering:}
When multiple objects are present, the order of target sequences matters. We compared between \textit{random}, \textit{size} (largest instance to smallest) and \textit{raster} ordering. We observe that raster ordering converges faster to a lower loss for the coord head among the three variants (Fig.~\ref{fig:ordering}). We hypothesize that raster is better since it enables our model to search for objects in a clear order (which is not possible for the other two variants) and learn faster with lesser ambiguity. And since the coordinate token is predicted first for an instance, lower coord loss (\ie, a better learned coord head) directly translates to an improved detection performance in terms of macroF$_1$ on both PBench and SaCo benchmarks.

\noindent \textbf{(\textit{vi}) Sequence format:} 
Since segmentation masks are highly expressive and bounding boxes can be trivially extracted from them, it is natural to ask why we utilize the \texttt{<coord><size><seg>} sequence rather than simply predicting \texttt{<seg>}. We justify our design choice based on two primary factors: (i) When trained solely on \texttt{<seg>}, the model defaults to predicting a merged semantic mask for all instances. Resolving this would require integrating an expensive segmentation mask encoder. Instead, we use a lightweight \texttt{<coord>} encoder; conditioning the \texttt{<seg>} prediction on these bounding boxes natively enables distinct instance masks. (ii) Generating \texttt{<seg>} requires invoking the upsampling module. Our sequence formulation allows the model to operate as a pure object detector by simply not decoding the \texttt{<seg>} token, bypassing the upsampling module entirely for faster inference.

\subsection{Objective functions}
\label{sec:loss}

\noindent We train Falcon Perception with a weighted sum of language modeling, structured geometry, dense mask, and feature-alignment losses. Each object is $O_k=(c_k,s_k,m_k)$ with center $c_k\in[0,1]^2$, size $s_k\in[0,1]^2$, and binary mask $m_k\in\{0,1\}^{H\times W}$. Coordinate/size/mask supervision is applied \emph{only} for the positive (present) queries, while absent (negative) queries contribute only through the language modeling loss.

\noindent \textbf{Language modeling loss:}
Let $X$ be the unified sequence and let $p_\theta(x_i\mid x_{<i},I)$ be the backbone next-token distribution. We compute cross-entropy on non-ignored tokens and normalize by the \emph{global} average number of valid tokens across data-parallel ranks to avoid bias from variable numbers of ignored tokens (\eg, due to variable image token counts):
\begin{equation}
\mathcal{L}_{\text{lm}} \;=\; \frac{\sum_{i:\,y_i\neq \texttt{IGNORE}} -\log p_\theta(y_i \mid X_{<i})}{\max(1,\;\overline{N}_{\text{text}})}\,,
\end{equation}
where $\overline{N}_{\text{text}}$ denotes the average number of valid text/task labels per loss-rank.

\noindent \textbf{Coordinate loss:}
The coordinate head predicts discrete bins with $B=1024$ bins per axis. We discretize ground-truth coordinates by
\begin{equation}
q(c) \;=\; \mathrm{clip}\!\Big(\mathrm{round}\big(c\cdot(B-1)\big),\;0,\;B-1\Big)\,,
\end{equation}
applied elementwise to $c_k=(c^x_k,c^y_k)$. Let $\ell^c_k\in\mathbb{R}^{2\times B}$ be the coordinate logits from the coordinate head. The loss is
\begin{equation}
\mathcal{L}_{\text{coord}} \;=\; \frac{1}{\max(1,\overline{N}_{\text{coord}})}\sum_{k\in\mathcal{K}^+}\sum_{a\in\{x,y\}}
\mathrm{CE}\!\left(\ell^{c,a}_k,\; q(c^a_k)\right),
\end{equation}
where $\mathcal{K}^+$ is the set of positive objects and $\overline{N}_{\text{coord}}$ is the global-average count of supervised coordinate targets.

\noindent \textbf{Size loss (log-scale binning):}
Size is supervised on a log scale to allocate more resolution to small objects. With $B=1024$ bins and $s_k=(s^w_k,s^h_k)$, we clamp and map to a normalized log space:
\begin{equation}
\tilde{s} \;=\; \frac{\log_2(\max(s,\;1/B)) - \log_2(1/B)}{0-\log_2(1/B)} \in [0,1]\,,
\qquad
q_s(s) \;=\; \mathrm{round}\big(\tilde{s}\cdot(B-1)\big)\,.
\end{equation}
Let $\ell^s_k\in\mathbb{R}^{2\times B}$ be size logits. We use
\begin{equation}
\mathcal{L}_{\text{size}} \;=\; \frac{1}{\max(1,\overline{N}_{\text{size}})}\sum_{k\in\mathcal{K}^+}\sum_{a\in\{w,h\}}
\mathrm{CE}\!\left(\ell^{s,a}_k,\; \mathrm{clip}(q_s(s^a_k),0,B-1)\right).
\end{equation}

\noindent \textbf{Mask losses (focal + dice):}
Given predicted mask logits $\hat{m}_k\in\mathbb{R}^{H\times W}$ (Equation~(4) in Section~\ref{sec:heads}) and ground-truth mask $m_k$, we use a weighted focal loss and a dice loss:
\begin{equation}
\mathcal{L}_{\text{seg}} \;=\; \beta\,\mathcal{L}_{\text{focal}} \;+\; \alpha\,\mathcal{L}_{\text{dice}},
\qquad \alpha=10,\;\beta=200,
\end{equation}
with normalization by the global-average number of supervised masks $\overline{N}_{\text{seg}}$:
\begin{equation}
\mathcal{L}_{\text{focal}} \;=\; \frac{1}{\max(1,\overline{N}_{\text{seg}})}\sum_{k\in\mathcal{K}^+}\mathrm{Focal}(\hat{m}_k,m_k),
\qquad
\mathcal{L}_{\text{dice}} \;=\; \frac{1}{\max(1,\overline{N}_{\text{seg}})}\sum_{k\in\mathcal{K}^+}\mathrm{Dice}(\hat{m}_k,m_k).
\end{equation}

\noindent \textbf{Gram feature alignment:}
To prevent loosing the image features from the distillation phase, we add a Gram loss between student and teacher patch features (when available). Let $F_s,F_t\in\mathbb{R}^{N\times d}$ denote (normalized) student/teacher visual features for the $N$ valid patches, and let $M\in\{0,1\}^{N}$ be the valid-patch mask. Define Gram matrices $G_s=F_sF_s^\top$ and $G_t=F_tF_t^\top$, and mask out invalid patch pairs. The loss is
\begin{equation}
\mathcal{L}_{\text{gram}} \;=\; \frac{\| (G_s-G_t)\odot (MM^\top)\|_F^2}{\max(1,\langle MM^\top, \mathbf{1}\rangle)}.
\end{equation}

\noindent \textbf{Final objective:}
Putting everything together, the final training objective is given by
\begin{equation}
\mathcal{L} \;=\; \mathcal{L}_{\text{lm}} \;+\; \mathcal{L}_{\text{coord}} \;+\; \mathcal{L}_{\text{size}} \;+\; \mathcal{L}_{\text{seg}} \;+\; \lambda_{\text{gram}}\mathcal{L}_{\text{gram}}\,,
\qquad
\lambda_{\text{gram}}=0.1.
\end{equation}

\subsection{Training Recipe\label{sec:recipe}}

Following the multi-teacher distillation, we train the model on the perception task for a total of approximately 685 Gigatokens (GT). We adopt  three stages  designed to first build a robust, globally-aware scene understanding, and subsequently specialize the model for the independent query task required at inference time. Throughout all stages, we strictly maintain a 1:1 ratio of positive to negative samples to mitigate hallucination.

\noindent \textbf{Stage 1: In-Context Listing (450 GT).} We train the model to predict the \textit{entire} sequence, including the text expressions and presence tokens. The objective here is full autoregressive modeling: the model effectively learns to "list" the inventory of the scene. This phase is important for building global context. By predicting objects in a sequence (\eg, a fork, then a knife, then a plate), the model learns object co-occurrence statistics and scene composition. We use the inverse learning rate decay from $4e^{-4}$ to $1e^{-4}$, and the maximum number of masks per expression is capped at 100 for training stability.

\noindent \textbf{Stage 2: Task alignment (225 GT).} While stage 1 builds context, it introduces a dependency that is undesirable at inference time: the model might learn to rely on the presence of previous objects to detect the current one. To resolve this, we transition to a decay stage that aligns the model with the final independent-query task. We introduce two critical masking strategies. First, \textit{Query masking (Attention Masking)}: we modify the attention mask so that tokens in query block $i$ cannot attend to tokens in query block $j$ (for $i \neq j$). This effectively simulates a batch of independent queries within a single sequence, forcing the model to ground each query based solely on the image. Second, \textit{Prompt Masking (Loss Masking)}: we mask the loss on the text expression tokens. Since the model has already learned language in Stage 1, we now want to focus its entire capacity on the spatial outputs. By removing the penalty for text prediction, the gradient signal becomes dominated by the presence classification and localization, sharpening the model's precision. We continue training with an inverse learning rate decay from $1e^{-4}$ to $4e^{-6}$, and increase the mask limit to 150 per expression.

\noindent \textbf{Stage 3: Long-Context Finetuning (10 GT).} The final stage is a short adaptation phase designed for extreme density, such as the "Crowded" split of our benchmark. We drastically increase the mask limit to 600 per experession. To prevent forgetting of the features learned in the previous 675 GT, the learning rate is dropped to a constant, minimal value of $1e^{-6}$. This stage primarily serves to adapt the position embeddings and causal buffers to the extended sequence lengths required for dense scenes.

%% file: sections/results.tex
\section{Results}
\label{sec:experiments}

\begin{figure}[t]
  \centering
  \includegraphics[width=1.0\textwidth, trim=4.5cm 4.5cm 4.5cm 4.5cm, clip]{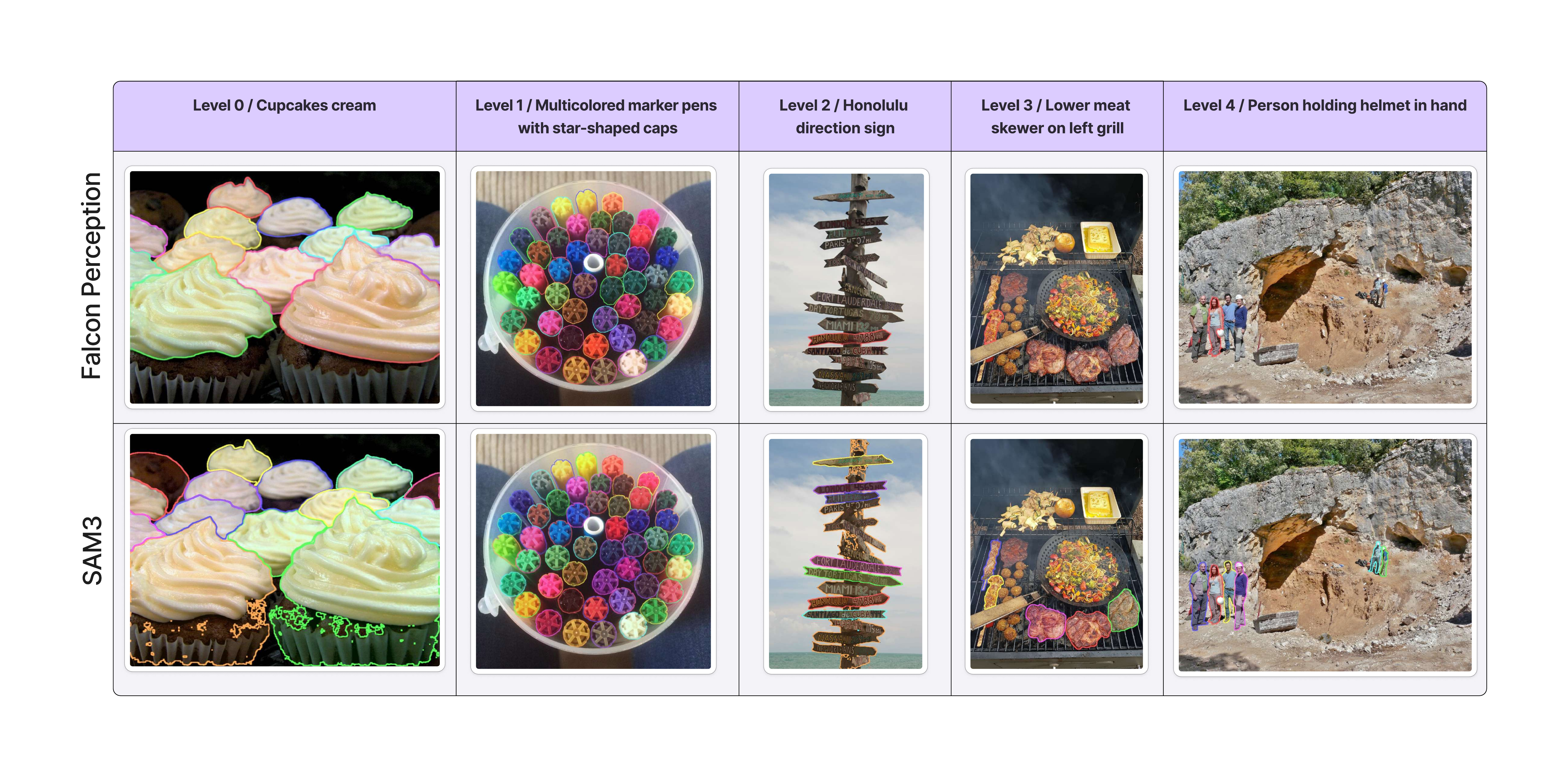}

\caption{\textbf{Qualitative comparison across PBench complexity levels.} Each column shows a different image and a prompt written to target one dominant capability (Levels~0--4; Table~\ref{tab:complexity_levels})). The top row shows Falcon Perception predictions and the bottom row shows SAM~3. Falcon Perception produces cleaner, more instance-specific masks as prompts become more compositional (OCR, spatial, relational), while SAM~3 more often fragments the target or returns extra/incorrect masks on higher-level prompts.}

\label{fig:main_result}
\end{figure}

\begin{table}[t]
    \centering
    \caption{\textbf{SA-Co Benchmark Results.} Comparison of Falcon Perception against state-of-the-art open-vocabulary segmentation models. We report Macro F$_1$, Positive Micro F$_1$ (pmF$_1$), and Image-Level MCC across all splits.\vspace{-0.2cm}}
    \label{tab:saco_results_booktabs}
    \resizebox{\textwidth}{!}{
    \begin{tabular}{l | ccc | ccc | ccc | ccc | ccc | ccc | ccc | ccc}
    \toprule
    \multirow{2}{*}{\textbf{Model}} & \multicolumn{3}{c}{\textbf{Average}} & \multicolumn{3}{c}{\textbf{Metaclip}} & \multicolumn{3}{c}{\textbf{SA-1B}} & \multicolumn{3}{c}{\textbf{Crowded}} & \multicolumn{3}{c}{\textbf{Food\&Drink}} & \multicolumn{3}{c}{\textbf{Sports Equip.}} & \multicolumn{3}{c}{\textbf{Attributes}} & \multicolumn{3}{c}{\textbf{Wiki-Common}} \\
    \cmidrule(lr){2-4} \cmidrule(lr){5-7} \cmidrule(lr){8-10} \cmidrule(lr){11-13} \cmidrule(lr){14-16} \cmidrule(lr){17-19} \cmidrule(lr){20-22} \cmidrule(lr){23-25}
    & F$_1$ & pmF$_1$ & MCC & F$_1$ & pmF$_1$ & MCC & F$_1$ & pmF$_1$ & MCC & F$_1$ & pmF$_1$ & MCC & F$_1$ & pmF$_1$ & MCC & F$_1$ & pmF$_1$ & MCC & F$_1$ & pmF$_1$ & MCC & F$_1$ & pmF$_1$ & MCC \\
    
    \midrule
    \multicolumn{25}{l}{\cellcolor{gray!10}\textbf{Baselines}} \\
    \quad gDino-T~\cite{Liu2023GroundingDM} & - & 16.2 & 0.15 & - & 13.9 & 0.21 & - & 15.4 & 0.20 & - & 3.4 & 0.08 & - & 9.8 & 0.10 & - & 11.2 & 0.10 & - & 47.3 & 0.29 & - & 12.1 & 0.06 \\
    \quad OWLv2*~\cite{minderer2024scalingopenvocabularyobjectdetection} & - & 42.0 & 0.57 & - & 34.3 & 0.52 & - & 26.8 & 0.50 & - & 30.7 & 0.51 & - & 49.4 & 0.65 & - & 56.2 & 0.64 & - & 56.2 & 0.63 & - & 40.3 & 0.54 \\
    \quad OWLv2~\cite{minderer2024scalingopenvocabularyobjectdetection} & - & 36.8 & 0.46 & - & 31.3 & 0.39 & - & 21.7 & 0.45 & - & 24.8 & 0.36 & - & 47.9 & 0.51 & - & 47.0 & 0.52 & - & 48.2 & 0.54 & - & 36.6 & 0.49 \\
    \quad LLMDet-L~\cite{fu2025llmdet} & - & 27.3 & 0.21 & - & 19.4 & 0.23 & - & 22.8 & 0.23 & - & 13.7 & 0.18 & - & 29.1 & 0.19 & - & 25.3 & 0.17 & - & 57.1 & 0.39 & - & 23.3 & 0.16 \\
    \quad APE-D~\cite{shen2024aligning} & - & 36.9 & 0.40 & - & 30.1 & 0.42 & - & 10.0 & 0.22 & - & 20.3 & 0.35 & - & 45.0 & 0.51 & - & 56.5 & 0.56 & - & 57.3 & 0.47 & - & 39.5 & 0.41 \\
    \quad DINO-X~\cite{ren2024dino} & - & 55.2 & 0.38 & - & 49.2 & 0.35 & - & 40.9 & 0.48 & - & 37.5 & 0.34 & - & 61.7 & 0.49 & - & 69.4 & 0.41 & - & \textbf{74.0} & 0.42 & - & 53.5 & 0.35 \\
    
    \midrule
    \multicolumn{25}{l}{\cellcolor{gray!10}\textbf{Vision-Language Models}} \\
    \quad Gemini 2.5~\cite{comanici2025gemini} & - & 46.1 & 0.29 & - & 33.8 & 0.29 & - & 32.1 & 0.41 & - & 30.3 & 0.27 & - & 59.5 & 0.33 & - & 53.5 & 0.28 & - & 63.1 & 0.30 & - & 50.3 & 0.25 \\
    \quad Qwen3-VL-2B~\cite{qwen3} & 50.3 & 36.8 & 0.65 & 45.4 & 31.9 & 0.63 & 46.4 & 31.5 & 0.71 & 26.4 & 17.1 & 0.62 & 53.4 & 36.7 & 0.70 & 60.5 & 51.1 & 0.75 & 68.4 & 54.5 & 0.61 & 51.7 & 35.1 & 0.52 \\
    \quad Qwen3-VL-4B~\cite{qwen3} & 59.0 & 46.1 & 0.69 & 54.9 & 41.3 & 0.65 & 55.9 & 41.3 & 0.77 & 37.5 & 27.2 & 0.67 & 58.7 & 40.7 & 0.77 & 68.1 & 61.1 & 0.77 & 75.5 & 64.1 & 0.68 & 62.4 & 46.9 & 0.54 \\
    \quad Qwen3-VL-8B~\cite{qwen3} & 56.4 & 41.1 & 0.79 & 51.4 & 36.2 & 0.76 & 53.3 & 34.8 & 0.82 & 32.2 & 19.9 & 0.77 & 56.3 & 37.4 & \textbf{0.85} & 67.4 & 54.9 & 0.87 & 74.4 & 62.3 & \textbf{0.77} & 59.7 & 42.2 & 0.77 \\
    \quad Qwen3-VL-30B~\cite{qwen3} & 62.8 & 49.9 & 0.64 & 59.0 & 45.5 & 0.59 & 57.8 & 44.9 & 0.73 & 40.7 & 31.1 & 0.57 & 63.1 & 41.4 & 0.70 & 73.7 & 66.3 & 0.75 & 78.6 & 69.3 & 0.63 & \textbf{66.8} & 50.7 & 0.49 \\
    \quad Moondream2~\cite{moondream2} & 62.5 & 53.5 & 0.43 & 56.5 & 45.7 & 0.42 & 54.8 & 41.5 & 0.58 & 43.8 & 40.5 & 0.45 & 67.3 & 55.0 & 0.43 & 72.9 & 68.9 & 0.49 & 76.7 & 69.6 & 0.38 & 65.2 & 53.3 & 0.43 \\
    \quad Moondream3~\cite{moondream3}  & 58.0 & 47.8 & 0.45 & 53.2 & 40.2 & 0.43 & 51.9 & 36.8 & 0.61 & 35.6 & 30.6 & 0.44 & 59.5 & 46.2 & 0.45 & 69.2 & 63.9 & 0.51 & 75.4 & 67.6 & 0.42 & 61.0 & 49.5 & 0.44 \\
    \quad SAM 3~\cite{carion2025sam3} & 62.3 & \textbf{66.1} & \textbf{0.82} & 56.0 & \textbf{58.6} & \textbf{0.81} & \textbf{68.3} & \textbf{62.6} & \textbf{0.86} & \textbf{62.7} & \textbf{67.7} & \textbf{0.90} & 58.1 & \textbf{67.3} & 0.79 & 71.2 & \textbf{73.8} & \textbf{0.89} & 71.1 & 72.0 & 0.76 & 49.0 & 60.9 & 0.66 \\
    
    \midrule
    \textbf{Ours: Falcon Perception} 
    & \textbf{68.0} & 62.1 & 0.64 
    & \textbf{63.8} & 53.2 & 0.67 
    & 64.7 & 49.7 & 0.76 
    & 59.1 & 57.9 & 0.68 
    & \textbf{70.3} & 66.2 & 0.61 
    & \textbf{75.2} & 71.3 & 0.75 
    & \textbf{79.3} & 72.2 & 0.60 
    & 63.9 & \textbf{64.3} & 0.38 \\

    \bottomrule
    \end{tabular}
    }
    \vspace{-0.3cm}
\end{table}

\noindent \textbf{Metrics:} We follow SAM 3~\citep{carion2025sam3} and report both Positive Micro F$_1$ (pmF$_1$) and Image-Level MCC (IL\_MCC) as the primary metrics for open-vocabulary perception-centric segmentation. We also consider Macro F$_1$ per sample (F$_1$). Additional details on the metrics are given in Appendix~\ref{appendix:metrics}.

\subsection{Main Results}
We evaluate Falcon Perception on: (1) The SA-Co benchmark for open-vocabulary segmentation, (2) Our new Perception Benchmark (PBench) for fine-grained capability analysis, and (3) The standard RefCOCO suite. To enable a fair comparison on segmentation metrics, we equip detection-only models (Qwen3-VL~\citep{qwen3}, Moondream~\citep{moondream2,moondream3}) with SAM~\citep{kirillov2023segment}, using their output boxes as prompts to generate pixel masks.

\noindent \textbf{SA-Co (open-vocabulary segmentation):} Table \ref{tab:saco_results_booktabs} presents our performance on the SA-Co benchmark, showing that Falcon Perception is competitive on SA-Co and is particularly strong on \emph{mask quality} as measured by Macro-F$_1$. On average, Falcon Perception reaches \textbf{68.0 F$_1$} compared to \textbf{62.3} for SAM3, and it improves F$_1$ on most splits (\eg, \textit{Food\&Drink:} \textbf{70.3 \vs 58.1}, \textit{Sports:} \textbf{75.2 \vs 71.2}, \textit{Attributes:} \textbf{79.3 \vs 71.1}). Falcon Perception lags behind SAM~3 in \emph{presence calibration}, and this gap is the main driver of lower overall robustness on splits with many hard negatives (\eg, \textit{Average MCC:} \textbf{0.64 \vs 0.82}, \textit{Wiki-Common MCC:} \textbf{0.38 \vs 0.66}). Falcon Perception lags behind SAM3 in presence calibration is primarily architectural. SAM3 utilizes fixed number object queries with bipartite Hungarian matching, meaning most queries natively learn to predict the empty class during training. This allows SAM 3 to inherently handle negatives (achieving a $cgF_1$of 34\% even without explicit negative samples). In contrast, our autoregressive formulation possesses no inherent empty-class mechanism; without explicit negative modeling, our $cgF_1$ of 34\% collapses to zero. Therefore, achieving an MCC of 0.64 represents a substantial relative gain, demonstrating the high efficacy of our targeted negative sampling strategy given the generative nature of the architecture. Indeed, preliminary results applying GRPO to our method using  $cgF_1$ as a reward have already yielded an 8-point improvement in MCC. In future releases, we plan to further close this presence calibration gap through reinforcement learning.

\begin{wraptable}{r}{0.6\textwidth}
    \centering
    \small
    \caption{\textbf{PBench \& RefCOCO Results.} \textbf{Top:} PBench performance (macro F$_1$) across complexity levels and crowdedness stress test. \textbf{Bottom:} RefCOCOm performance (macro F$_1$) on validation set.}
    \label{tab:pbench_refcoco}
    \setlength{\tabcolsep}{4pt}

    \adjustbox{max width=0.6\textwidth}{
    \begin{tabular}{p{4.5cm}cccccccc}
    \toprule
    \textbf{Benchmark} & \rotatebox{70}{\textbf{Qwen3-VL-2B}} & \rotatebox{70}{\textbf{Qwen3-VL-4B}} & \rotatebox{70}{\textbf{Qwen3-VL-8B}} & \rotatebox{70}{\textbf{Qwen3-VL-30B}} & \rotatebox{70}{\textbf{Moondream2-2B}} & \rotatebox{70}{\textbf{Moondream3-9B}} & \rotatebox{70}{\textbf{SAM 3-0.9B}} & \rotatebox{70}{\textbf{Falcon Perception}} \\
    \midrule
    \multicolumn{9}{l}{\cellcolor{gray!10}\textbf{PBench}} \\
    \quad L0: Simple objects & 54.1 & 67.3 & 65.6 & \textbf{69.2} & 68.0 & 63.2 & 64.3 & 65.1 \\
    \quad L1: Attribute & 50.1 & 63.9 & 65.6 & \textbf{68.8} & 58.4 & 59.6 & 54.4 & 63.6 \\
    \quad L2: OCR guided & 41.2 & 58.2 & 58.2 & \textbf{61.2} & 46.6 & 41.8 & 24.6 & 38.0 \\
    \quad L3: Spatial understanding & 35.1 & 49.0 & 49.1 & 52.9 & 43.8 & 45.4 & 31.6 & \textbf{53.5} \\
    \quad L4: Relation binding & 36.7 & 48.3 & 50.9 & \textbf{55.2} & 36.9 & 42.3 & 33.3 & 49.1 \\
    \quad Dense & 4.6 & 8.4 & 4.7 & 8.9 & 14.2 & 12.9 & 58.4 & \textbf{72.6} \\
    \midrule
    \quad Average & 37.0 & 49.2 & 49.0 & 52.7 & 44.7 & 50.5 & 44.4 & \textbf{57.0} \\
    \midrule
    \multicolumn{9}{l}{\cellcolor{gray!10}\textbf{RefCOCO}} \\
    \quad RefCOCOm & 54.0 & 64.4 & 64.7 & 72.1 & 62.0 & \textbf{87.1} & 36.4 & 77.3 \\
    \bottomrule
    \end{tabular}
    }
\end{wraptable}

\noindent \textbf{PBench (capability profiling):}
While SA-Co studies performance across seven domains, PBench isolates \emph{what capability is required} by construction (Levels~0--4 \& Dense). Table~\ref{tab:pbench_refcoco} shows that Falcon Perception stays robust as prompts become more compositional. Compared to SAM~3, gains are modest on Level~0 (simple classes), but become substantial on higher levels where semantic understanding is critical: \textbf{+9.2} on Level~1 (attributes), \textbf{+13.4} on Level~2 (OCR-guided),  
and a remarkable \textbf{+21.9} on Level~3 (spatial understanding). This confirms that while SAM~3 excels at promptable segmentation, it is not designed to resolve complex semantic ambiguities (Levels 2--4) without an external reasoning module.
\noindent Crucially, Falcon Perception (0.6B) is highly competitive with far larger VLMs. It outperforms Moondream2-2B and Moondream3-9B on almost all complex reasoning tiers (L1--L4) and the Dense split, despite being $3\times$--$15\times$ smaller. Against the Qwen3-VL family, Falcon Perception matches or exceeds the 8B model on spatial (L3) and relational (L4) tasks, and even surpasses the massive Qwen3-VL-30B on the Dense split (\textbf{72.6} \vs 8.9). This highlights the efficiency of our native dense architecture: generalist VLMs struggle to maintain precise localization in crowded scenes (Dense split), often hallucinating or merging instances, whereas Falcon Perception's Chain-of-Perception interface remains stable as the number of targets increases.

\begin{figure}[t]
    \centering
    \includegraphics[width=\linewidth]{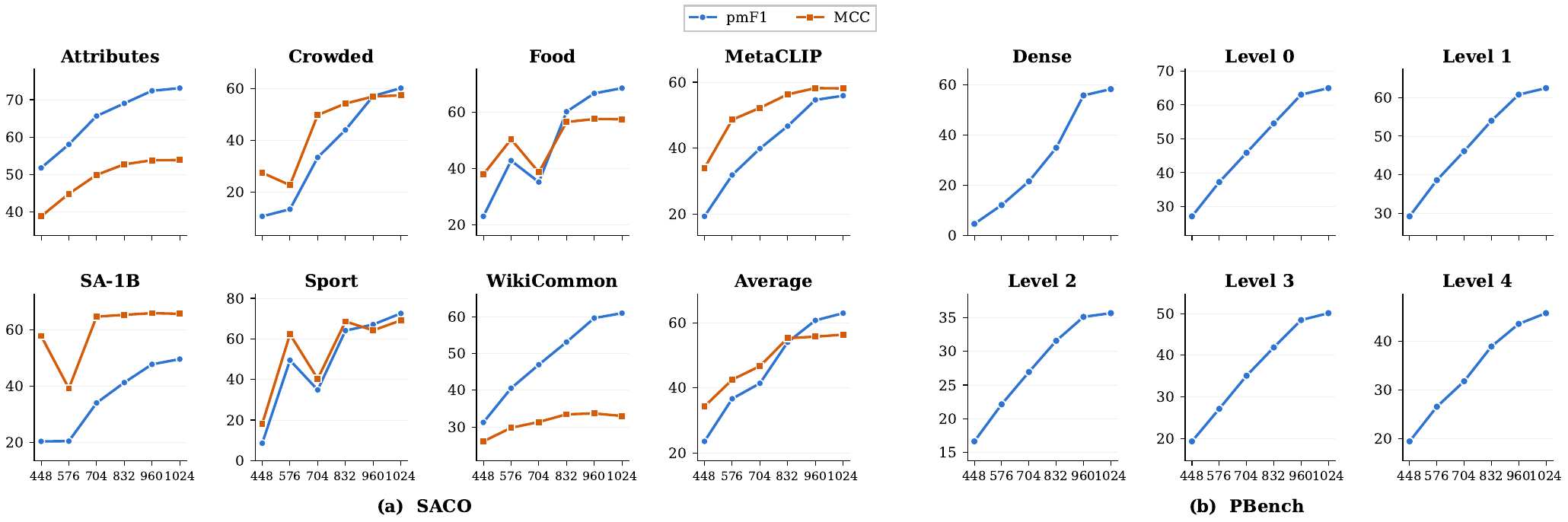}
    \caption{\textbf{Resolution Scaling.} Performance as a function of image resolution. While semantic tasks (\eg, Attributes) are relatively robust to lower resolutions, dense and small-object tasks (\eg, Crowded, Sports, Dense) exhibit a phase transition, requiring high resolution to resolve individual instances.}
    \label{fig:resolution}
\end{figure}

\noindent \textbf{Impact of resolution:} We analyze the effect of input resolution on dense perception performance by evaluating Falcon Perception across a range of resolutions from $448^2$ to $1024^2$. As shown in Figure~\ref{fig:resolution}, increasing resolution yields consistent performance gains, but the impact is non-uniform across metrics and tasks.

\noindent  we observe a  "Dense" phase transition on \textbf{PBench Dense stress test}, where the model is effectively blind at $448^2$ (3.9\% micro-F1) but reaches 61.0\% at $1024^2$. This $\mathbf{15\times}$ improvement confirms that for crowded scenes, while the transformer backbone can understand semantics at low resolution, the "bottleneck" for dense perception is strictly spatial: resolving high details in the input image is a prerequisite for the specialized heads to operate effectively. Furthermore, we notice a decoupling between recognition and grounding. At $448^2$, the model achieves a respectable IL\_MCC of 0.34 (detecting presence) but a poor pmF1 of 23.6 (drawing masks). As resolution increases to $1024^2$, localization quality (pmF1) improves by $\mathbf{2.7\times}$, while classification (MCC) grows more modestly by $\mathbf{1.6\times}$. This suggests that low-resolution features are sufficient for semantic recognition ("is there a cat?"), but high-resolution input is  necessary for precise spatial grounding ("where exactly is the cat?").

\noindent \textbf{Inference Budget (Fixed \vs Adaptive Resizing):} We investigate the trade-off between inference cost and performance by comparing two resizing strategies: (1) \textit{Adaptive ("Smart") Resizing}, where images smaller than $1024^2$ are processed at their native resolution to save compute, and (2) \textit{Fixed Resizing}, where all images are upscaled to $1024^2$.

\begin{figure}[t]
    \centering
    \includegraphics[width=\linewidth]{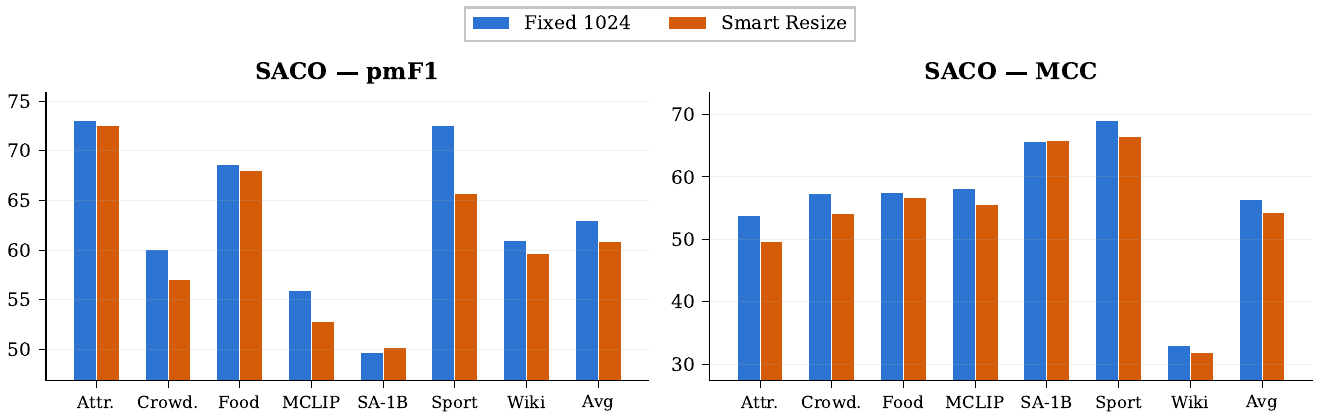}
    \caption{\textbf{Resizing Strategy.} Comparison of Adaptive \vs Fixed ($1024^2$) resizing on SA-Co. Upscaling all images to a fixed high resolution consistently improves performance, particularly for crowded scenes.}
    \label{fig:resize_strategy}
\end{figure}

\noindent We find that Fixed Resizing consistently outperforms the adaptive approach, even for images natively smaller than the target resolution. On the SA-Co benchmark, Fixed Resizing yields an average improvement of \textbf{+1.6 pmF1} (63.9 \vs 62.3) and \textbf{+1.6 MCC}. The gain is most pronounced on the \textit{Crowded} split (\textbf{+3.7 pmF1}), suggesting that upscaling small images onto a finer grid is critical for resolving dense clusters and small objects.

\subsection{Effect of Sampling\label{sec:sampling}}

\begin{figure*}[t]
    \centering
    \begin{subfigure}[b]{0.77\textwidth}
        \centering
        \includegraphics[width=\textwidth, trim=3cm 9cm 3cm 0.1cm, clip]{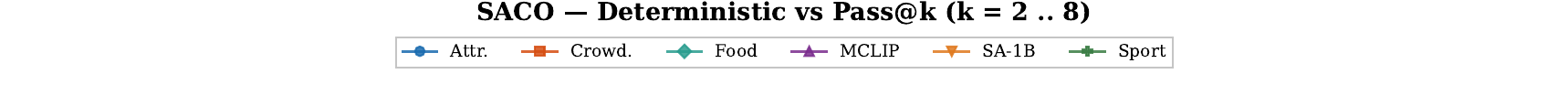}\\
        \includegraphics[width=\textwidth, trim=0 0 0 4cm, clip]{figures/sampling/sampling_comparison.pdf}
        \caption{SACO}
        \label{fig:saco_plots}
    \end{subfigure}
    \hfill
    \begin{subfigure}[b]{0.22\textwidth}
        \centering
        \includegraphics[width=\textwidth]{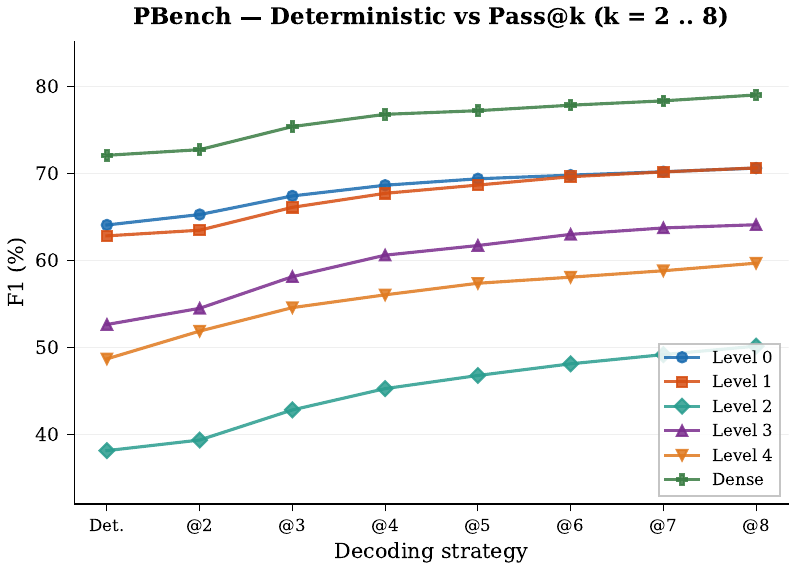}
        \caption{PBench}
        \label{fig:pbench_plots}
    \end{subfigure}
    
    \caption{\textbf{Pass@k Performance scaling.} Evolution of detection metrics across SACO (left) and PBench (right) as the number of samples $k$ increases. Performance consistently scales with $k$, particularly on the most challenging subsets.}
    \label{fig:multipass_plots}
\end{figure*}

To further investigate the latent capabilities of Falcon Perception, we evaluate its performance under a sampling-based inference regime, and report Pass@$k$ (best score out of $k$ predictions). Doing so, we hypothesize that the maximum likelihood objective used during training allows the model to capture a rich distribution of potential outputs, some of which may be more accurate than the most certain prediction in complex scenarios.

\noindent \textbf{Sampling Mechanism:} 
We enable stochastic decoding by sampling from the output heads of the model across three primary dimensions:
\begin{itemize}[nosep]
    \item \textit{Language Head:} We sample from the vocabulary logits, allowing for diversity in the generated tokens. This mainly influences the end of generation, \textit{i.e.}, whether the model decides if there is an object corresponding to the query in the image or not, and how many objects there are to detect and segment.
    \item \textit{Center Head:} We sample from the $x$ and $y$ bin logits, which governs the localization of object centers. Thus, this influences the exact localization of the object and which object to detect at the current timestep. 
    \item \textit{Size Head:} We sample from the $h$ and $w$ bin logits, influencing the predicted dimensions of the bounding boxes.
\end{itemize}

\noindent \textbf{Experimental Protocol:}
We evaluate the model on both the SA-Co and PBench benchmarks. For each test instance, we generate $k$ independent samples and select the "best" prediction (Pass@$k$) based on the task-specific metric. We compare the baseline deterministic (argmax) performance against $k \in \{2, 4, 6, 8\}$.

\begin{wraptable}{r}{0.5\textwidth}
    \centering
    \caption{\textbf{Multi-Pass PBench Benchmark Results.} Comparison of Falcon Perception across different baseline detection passes on PBench levels and dense scenarios against SAM3.}
    \label{tab:pbench_multipass}
    \setlength{\tabcolsep}{6pt}

    \adjustbox{max width=0.5\textwidth}{
    \begin{tabular}{l ccc ccc}
    \toprule
    \textbf{Model} & \textbf{Level 0} & \textbf{Level 1} & \textbf{Level 2} & \textbf{Level 3} & \textbf{Level 4} & \textbf{Dense}\\
    \midrule
    \multicolumn{7}{l}{\cellcolor{gray!10}\textbf{State-of-the-Art}} \\
    SAM 3 & 64.3 & 54.4 & 24.6 & 31.6 & 33.3 & 58.4 \\
    \midrule
    \multicolumn{7}{l}{\cellcolor{gray!10}\textbf{Falcon Perception}} \\
    Det. (baseline) & 64.0 & 62.8 & 38.1 & 52.6 & 48.7 & 72.1\\
    Pass@2        & 65.2 & 63.4 & 39.3 & 54.5 & 51.8 & 72.7 \\
    Pass@4        & 68.6 & 67.7 & 45.2 & 60.6 & 56.0 & 76.8 \\
    Pass@6        & 69.8 & 69.6 & 48.1 & 63.0 & 58.0 & 77.8 \\
    Pass@8        & \textbf{70.6} & \textbf{70.6} & \textbf{50.1} & \textbf{64.1} & \textbf{59.7} & \textbf{79.0} \\
    \bottomrule
    \end{tabular}
    }
\end{wraptable}

\noindent \textbf{Results:} As shown in Table~\ref{tab:saco_multipass} \& ~\ref{tab:pbench_multipass} and Figure~\ref{fig:multipass_plots}, performance across all metrics and subsets consistently improves as $k$ increases. Especially, we note that we outperform the deterministic baseline from pass@$2$.
On the SA-Co benchmark (Table~\ref{tab:saco_multipass}), the average $cgF_1$ score jumps from $34.7$ (Baseline) to $54.3$ (Pass@8), on-par with SAM3, \textit{i.e.}, a \textbf{+19.6} point absolute increase. The most significant delta is observed in the \textit{Wiki-Common} subset, where $cgF_1$ improves from $19.3$ to $45.0$, outperforming SAM3 from pass@$4$. 
On PBench (Table~\ref{tab:pbench_multipass}), we observe a similar trend across all difficulty levels. The "hard" scenarios show the most significant recovery: for \textit{Level 2}, \textit{Level 3}, and \textit{Level 4}, $F_1$ scores increase by \textbf{+12.0}, \textbf{+11.5}, and \textbf{+11.0} points respectively when moving from deterministic to Pass@8.
Furthermore, we find that sampling is particularly effective in "hard" scenes with extreme occlusion, dense object clusters, or expressions requiring complex reasoning. In these instances, the model's first-best prediction might fail, but the underlying probability distribution often contains the correct solution within the top few samples (see Section~\ref{sec:appendix_sampling} for the qualitative analysis).

\begin{table}[t]
    \centering
    \caption{\textbf{Multi-Pass SA-Co Benchmark Results.} Comparison of Falcon Perception across different baseline detection passes against SAM 3.\vspace{-0.2cm}}
    \label{tab:saco_multipass}
    \resizebox{\textwidth}{!}{
    \begin{tabular}{l ccc ccc ccc ccc ccc ccc ccc ccc}
    \toprule
    \multirow{2}{*}{\textbf{Model}} & \multicolumn{3}{c}{\textbf{Average}} & \multicolumn{3}{c}{\textbf{Metaclip}} & \multicolumn{3}{c}{\textbf{SA-1B}} & \multicolumn{3}{c}{\textbf{Crowded}} & \multicolumn{3}{c}{\textbf{Food\&Drink}} & \multicolumn{3}{c}{\textbf{Sports Equip.}} & \multicolumn{3}{c}{\textbf{Attributes}} & \multicolumn{3}{c}{\textbf{Wiki-Common}} \\
    \cmidrule(lr){2-4} \cmidrule(lr){5-7} \cmidrule(lr){8-10} \cmidrule(lr){11-13} \cmidrule(lr){14-16} \cmidrule(lr){17-19} \cmidrule(lr){20-22} \cmidrule(lr){23-25}
    & cgF$_1$ & pmF$_1$ & MCC & cgF$_1$ & pmF$_1$ & MCC & cgF$_1$ & pmF$_1$ & MCC & cgF$_1$ & pmF$_1$ & MCC & cgF$_1$ & pmF$_1$ & MCC & cgF$_1$ & pmF$_1$ & MCC & cgF$_1$ & pmF$_1$ & MCC & cgF$_1$ & pmF$_1$ & MCC \\
    
    \midrule
    \multicolumn{25}{l}{\cellcolor{gray!10}\textbf{State-of-the-Art}} \\
    \quad SAM 3 & 54.2 & 66.1 & \textbf{0.82} & 47.5 & 58.6 & \textbf{0.81} & \textbf{53.8} & \textbf{62.6} & \textbf{0.86} & \textbf{60.9} & \textbf{67.7} & \textbf{0.90} & 53.2 & 67.3 & 0.79 & 65.7 & 73.8 & \textbf{0.89} & 54.7 & 72.0 & 0.76 & 40.2 & 60.9 & \textbf{0.66} \\
    
    \midrule
    \multicolumn{25}{l}{\cellcolor{gray!10}\textbf{Falcon Perception}} \\
    \quad Det. (Baseline) & 34.7 & 62.9 & 0.55 & 31.0 & 55.1 & 0.56 & 32.2 & 49.8 & 0.65 & 34.3 & 60.5 & 0.57 & 38.8 & 68.6 & 0.57 & 48.3 & 72.2 & 0.67 & 39.4 & 73.1 & 0.54 & 19.3 & 60.9 & 0.32 \\
    \quad Pass@2 & 40.5 & 64.6 & 0.63 & 37.0 & 57.3 & 0.65 & 36.5 & 51.6 & 0.71 & 39.4 & 61.1 & 0.65 & 44.2 & 69.9 & 0.63 & 54.4 & 73.0 & 0.74 & 45.9 & 74.5 & 0.62 & 25.8 & 64.9 & 0.40 \\
    \quad Pass@4 & 48.1 & 67.8 & 0.71 & 43.7 & 60.4 & 0.72 & 41.5 & 54.4 & 0.76 & 45.0 & 63.1 & 0.71 & 54.3 & 73.7 & 0.74 & 62.3 & 76.1 & 0.82 & 54.1 & 77.0 & 0.70 & 35.7 & 69.6 & 0.51 \\
    \quad Pass@6 & 51.8 & 68.9 & 0.75 & 47.2 & 62.0 & 0.76 & 43.9 & 55.5 & 0.79 & 47.9 & 63.9 & 0.75 & 59.1 & 75.5 & 0.78 & 65.4 & 77.2 & 0.85 & 58.2 & 78.2 & 0.74 & 41.0 & 70.4 & 0.58 \\
    \quad Pass@8 & \textbf{54.3} & \textbf{69.8} & 0.78 & \textbf{49.5} & \textbf{63.0} & 0.79 & 45.4 & 56.3 & 0.81 & 49.8 & 64.3 & 0.77 & \textbf{61.9} & \textbf{76.3} & \textbf{0.81} & \textbf{67.8} & \textbf{77.9} & 0.87 & \textbf{60.9} & \textbf{79.1} & \textbf{0.77} & \textbf{45.0} & \textbf{71.4} & 0.63 \\
    \bottomrule
    \end{tabular}
    \vspace{-0.2cm}
    }
\end{table}

\subsection{Initialization: Random \texorpdfstring{\vs}{vs} Distillation\label{sec:init_compare}}
Here, we analyze the impact of initializing the Falcon-Perception weights using the multi-teacher distillation, detailed earlier in Sec.~\ref{sec:mtd}. To this end, we pretrain and decay (stages 1 and 2 of our recipe) a smaller 300M-sized model on the detection task (bounding boxes) 
\begin{wrapfigure}{r}{0.5\textwidth}
\centering
\includegraphics[width=0.8\linewidth]{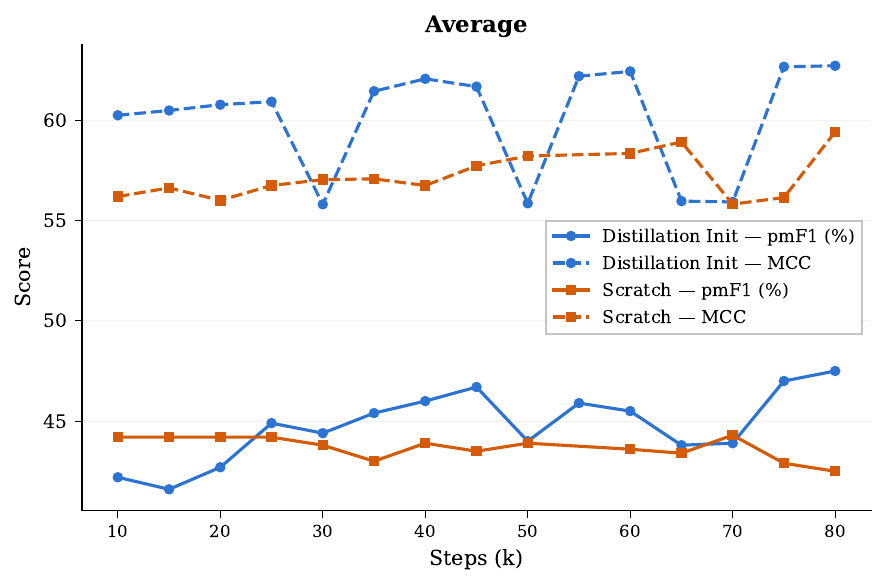}
\caption{\textbf{Impact of initialization:} Random weight initialization results in lower performance, compared to distillation initialization, even after being trained long enough for 230K steps over pretraining and decay (stages 1 and 2) combined.\label{fig:distil}}
\end{wrapfigure}
with different weights initialization. From Figure~\ref{fig:distil}, we observe that even after 230K steps of training across both stages, the average performance of randomly initialized model lags the distillation initialized model on the Sa-Co benchmark on both pmF1 and classification (MCC) metrics by a significant margin. Additionally, we note that training with scratch initialization for the segmentation task diverged in the early stages of pretraining itself. This is likely attributed to the fact that accurately predicting the instance masks requires enhanced image features that encode the structure of the scene, which is not possible in early stages of the training with random initialization. Consequently, it results in high gradients for the image tokens leading to divergence of the model weights. Both these ablations clearly indicate the necessity and effectiveness of the multi-teacher distillation stage in our training pipeline.

%% file: sections/ocr.tex
\section{OCR Extension}
\label{sec:ocr}

Modern OCR has evolved far beyond simple text extraction from scanned documents. Today's systems must handle diverse layouts, mathematical formulas, tables, charts, and multilingual content, all within a single unified framework. The recent emergence of Vision-Language Models (VLMs) for OCR~\citep{li2025monkeyocr, cui2025paddleocrvl, li2025dotsocr} has demonstrated that large-scale end-to-end models can outperform traditional pipeline approaches. However, many of these systems remain large (1B--3B+ parameters) and are either limited to a single task (\eg, document parsing only) or require costly multi-stage pipelines involving separate layout analysis modules.

We extend the Falcon Perception architecture to text-heavy document understanding with \textbf{FalconOCR}, a compact 300M-parameter model that repurposes our early-fusion dense transformer backbone for OCR tasks. Our design philosophy is to maximize the utility of our unified architecture: the same model that performs open-vocabulary segmentation can, with task-specific training data, learn to recognize and structure text. Rather than building a monolithic end-to-end document parser, we adopt a modular two-stage pipeline that decouples layout analysis from text recognition. This separation offers practical advantages: each component can be developed, debugged, and upgraded independently, and the OCR model can focus exclusively on element-level recognition without needing to also solve the layout problem.

\subsection{Pipeline}

Our inference pipeline operates in two stages, following the approach adopted by several recent OCR systems~\citep{li2025monkeyocr, niu2025mineru25decoupledvisionlanguagemodel, cui2025paddleocrvl} that combine a dedicated layout detector with a VLM-based recognizer.

\noindent \textbf{Stage 1 -- Layout Detection:} Given a full-page document image, we first apply PP-DocLayoutV3~\citep{cui2025paddleocr30technicalreport}, a lightweight and publicly available layout analysis model, to detect document elements and produce axis-aligned bounding boxes. PP-DocLayoutV3 categorizes each detected region into one of several element types: text blocks, tables, mathematical formulas, figures, headers, footers, and captions, providing both spatial coordinates and semantic labels. This off-the-shelf detector has been trained on large-scale document layout datasets and provides robust element localization across diverse document styles, including academic papers, invoices, scanned books, and web-captured screenshots, without requiring any additional fine-tuning from our side.

\noindent \textbf{Stage 2 -- Element-level OCR:} Each detected bounding box is cropped from the original high-resolution image and fed independently into our 300M FalconOCR model. The model performs end-to-end recognition and structured output generation conditioned on the element type. Specifically, text blocks are transcribed into plain-text, mathematical formulas are converted to \LaTeX{}, and tables are rendered as HTML. The element-type label from Stage~1 is encoded as a task-specific prompt prefix, allowing the model to adapt its output format accordingly. This design enables a single compact model to handle heterogeneous document elements within a unified autoregressive framework.

\noindent \textbf{Post-processing and Assembly:} After all cropped elements have been processed, their outputs are reassembled into a single structured document following the spatial reading order determined by the layout detector. This yields a complete Markdown representation of the page, with formulas in \LaTeX{} and tables in HTML, ready for downstream consumption by RAG systems or LLM training pipelines.

\subsection{Model Architecture}

FalconOCR uses the same unified dense transformer architecture described in Section~\ref{sec:architecture}: a single autoregressive stack that processes image patches and text tokens in a shared parameter space with hybrid attention masking (bidirectional for image tokens, causal for text tokens). The 300M-parameter (22-layer) variant is used. Crucially, unlike the Falcon Perception segmentation model, FalconOCR is initialized \textit{from scratch}, \ie, without the multi-teacher distillation stage described in Section~\ref{sec:mtd}. This choice was motivated by the observation that the visual features required for OCR (fine-grained glyph recognition, stroke-level discrimination) differ substantially from the object-level features learned during distillation from DINOv3 and SigLIP2. Training from a randomly initialized model allows the backbone to develop text-optimized representations from the ground-up.
The model processes cropped element images at their native aspect ratio (up to the maximum token budget, by resizing the maximum width to 1024 pixels) using the same scatter-and-pack strategy as the perception model. Since each input is a cropped region rather than a full page, the typical sequence length is shorter, allowing efficient batched processing.

\subsection{Training Recipe}

\noindent \textbf{Data:} We focus only on English language. We train FalconOCR on a curated mixture of OCR-specific datasets spanning three core tasks: (i)~general document text parsing across multiple document types - digital PDFs, old scans, typewritten documents, \etc. (ii)~mathematical and scientific formula recognition, and (iii)~table structure recognition. Additionally, we include handwriting, real world scenes with text, and other miscellaneous text-heavy datasets. The training data includes both publicly available datasets and synthetic samples generated from rendered \LaTeX{} and HTML sources, ensuring broad coverage of document styles, fonts, languages, and degradation levels.

\noindent \textbf{Schedule:} Training proceeds in two phases. The first phase consists of \num{250}k iterations of standard pre-training at a constant learning rate, during which the model learns the core OCR capabilities across all element types. This is followed by a cosine learning rate decay over \num{20}k steps, during which the learning rate is annealed to near zero. We monitor validation performance throughout the decay phase and select the best checkpoint based on a held-out set of document parsing samples. This two-phase approach, \ie, long stable training followed by a short annealing, has been shown to yield robust final checkpoints in language model pre-training~\citep{hu2024minicpm}, and we find it equally effective for the OCR task.

\noindent \textbf{Objective:} The training objective follows the same formulation described in Section~\ref{sec:loss}, \ie, standard cross-entropy on the autoregressive text tokens with the image tokens excluded from the loss. The sequence packing infrastructure, global loss normalization across data-parallel ranks, and native resolution handling are shared with the perception training pipeline. No additional OCR-specific losses (\eg, CTC, attention-based sequence losses) are used and the model learns purely through next-token prediction on structured text outputs.

\subsection{Deployment}

Efficient serving is critical for OCR workloads, where a single document may contain dozens of elements that must be processed in rapid succession. FalconOCR is integrated with vLLM~\citep{kwon2023efficient}, which provides continuous batching, PagedAttention, and optimized CUDA kernels for autoregressive generation. Thanks to the compact 300M-parameter footprint, the model achieves a throughput of approximately \num{3000} output tokens per second on a single GPU, making it practical for large-scale document processing pipelines where millions of pages must be digitized. The two-stage design also lends itself naturally to parallelism: multiple cropped elements from the same page (or different pages) can be batched together, maximizing GPU utilization.

Additionally, for latency critical applications, one can skip the layout detection step and perform OCR on the whole image of the document. While this leads to slightly degraded performance on dense documents due to our maximum resolution being capped at 1024, it is still quite useful for sparse documents and applications where latency is critical.

\subsection{Evaluation Benchmarks}

We evaluate FalconOCR on two complementary benchmarks: olmOCR~\citep{poznanski2025olmocr} and OmniDocBench~\citep{ouyang2025omnidocbench} that test document parsing from two complementary angles. Together, they provide a comprehensive picture of OCR robustness across input diversity, layout complexity, and evaluation granularity.

\noindent \textbf{olmOCR}~\citep{poznanski2025olmocr} emphasizes \textit{input diversity}. Its eight category splits span a wide range of real-world conditions: old scans with degraded ink and bleed-through, tiny text requiring high-resolution discrimination, handwritten content, ArXiv papers with dense mathematical notation, multi-column layouts, and standard base documents. The evaluation protocol is \textit{binary}: for each test case, the benchmark checks discrete properties of the predicted output, \ie, whether a specific text phrase is present or absent, whether one text span correctly follows another in the reading order, and whether cells in predicted HTML tables appear under the correct column headers. A prediction either passes or fails each check, and the final accuracy is the fraction of checks passed. This binary paradigm makes olmOCR a strict test of factual correctness, but it does not penalize minor formatting differences or measure the quality of the full parsed output.
We drop non-English documents from our evaluation.

\noindent \textbf{OmniDocBench}~\citep{ouyang2025omnidocbench} takes a complementary approach, emphasizing \textit{layout complexity} with \textit{continuous} evaluation metrics. Its document categories focus on digitally typeset content, \ie, slides, academic articles, and newspapers, where the primary challenge is not input degradation but structural complexity. Newspapers, for instance, present some of the hardest parsing scenarios: reading order can switch mid-page due to section breaks and headlines, with multiple vertical columns that interleave in non-trivial ways. The evaluation operates on the \textit{entire parsed page}: the full predicted text is aligned against the ground truth at the element level, and three continuous metrics are computed. For plain text, the \textit{Edit Distance} (lower is better) measures character-level accuracy between matched text blocks. For mathematical formulas, the \textit{CDM} (Character Detection Metric; higher is better) evaluates \LaTeX{} output fidelity. For tables, \textit{TEDS} (Tree-Edit-Distance-based Similarity; higher is better) compares the structural and content accuracy of predicted HTML against the ground-truth table tree. The \textit{Overall} score aggregates these three sub-metrics. Crucially, all three metrics depend on a matching algorithm that pairs predicted elements with ground-truth elements before scoring, making the evaluation sensitive to the quality of this alignment step.
We use the version 1.5 of OmniDocBench which adds a few complex documents over the original version, and focus our evaluation on the \textit{English-only} subset of the documents, discarding Chinese language documents.

\subsection{Results on olmOCR}

Table~\ref{tab:olmocr_results_booktabs} compares FalconOCR against state-of-the-art OCR systems spanning three categories: commercial APIs (Mistral OCR 3~\citep{mistral2025ocr}, Gemini 3 Pro/Flash~\citep{gemini_deepmind}, GPT 5.2~\citep{openai_gpt52}), specialized OCR models (Chandra~\citep{chandra2025}, DeepSeek OCR v2~\citep{wei2026deepseek}), and modular VLM pipelines (PaddleOCR VL \citep{cui2025paddleocrvl} / VL 1.5~\citep{cui2026paddleocrvl15}). All comparison models have significantly more parameters and/or rely on proprietary infrastructure.

\begin{table}[t]
    \centering
    \caption{\textbf{olmOCR Benchmark Results.} Category-wise performance comparison of FalconOCR against state-of-the-art OCR models. We report accuracy (\%) across all category splits.}
    \label{tab:olmocr_results_booktabs}
    \resizebox{\textwidth}{!}{
    \begin{tabular}{l c cccccccc}
    \toprule
    \textbf{Model} & \textbf{Average} & \textbf{ArXiv Math} & \textbf{Base} & \textbf{Hdr/Ftr} & \textbf{TinyTxt} & \textbf{MultCol} & \textbf{OldScan} & \textbf{OldMath} & \textbf{Tables} \\
    \midrule
    Mistral OCR 3~\citep{mistral2025ocr}       & 81.7          & 85.4          & 99.9 & 93.8          & 88.9          & 82.1          & 48.8          & 68.3          & 86.1          \\
    Chandra~\citep{chandra2025}             & 82.0          & 81.4          & 99.8          & 88.8          & 91.9 & 82.9          & 49.2          & 73.6          & 88.2          \\
    Gemini 3 Pro~\citep{gemini_deepmind}        & 80.2          & 70.6          & 99.8          & 84.0          & 90.3          & 79.2          & 47.5          & 84.9          & 84.9          \\
    PaddleOCR VL 1.5~\citep{cui2026paddleocrvl15}    & 79.3          & 85.4          & 98.8          & 96.9 & 80.8          & 82.6          & 39.2          & 66.4          & 84.1          \\
    PaddleOCR VL~\citep{cui2025paddleocrvl}        & 79.2          & 85.4          & 98.6          & 96.9 & 80.8          & 82.5          & 38.8          & 66.4          & 83.9          \\
    DeepSeek OCR v2~\citep{wei2026deepseek}     & 78.8          & 81.9          & 99.8          & 95.6          & 88.7          & 83.6          & 33.7          & 68.8          & 78.1          \\
    Gemini 3 Flash~\citep{gemini_deepmind}      & 77.5          & 66.5          & 99.8          & 83.8          & 88.2          & 73.7          & 46.0          & 85.8 & 75.9          \\
    GPT 5.2~\citep{openai_gpt52}          & 69.8          & 61.0          & 99.8          & 75.6          & 62.2          & 70.2          & 34.6          & 75.8          & 79.0          \\
    \midrule
    \textbf{FalconOCR} & 80.3 & 80.5 & 99.5 & 94.0 & 78.5 & 87.1 & 43.5 & 69.2 & 90.3 \\
        \bottomrule
    \end{tabular}
    }
\end{table}

\noindent \textbf{Overall Performance.} Our model achieves an average accuracy of {80.3\%}, placing it competitively among systems that are often orders of magnitude larger. With only 300M parameters, it matches or exceeds the performance of Gemini 3 Pro (80.2\%), PaddleOCR VL 1.5 (79.3\%), DeepSeek OCR v2 (78.8\%), and GPT 5.2 (69.8\%). The gap to the top-performing systems, \ie, Chandra (82.0\%) and Mistral OCR 3 (81.7\%) is modest at 1.7 and 1.4 points respectively, despite these models leveraging significantly larger backbones and proprietary training data.

\noindent \textbf{Category-level Analysis.} FalconOCR demonstrates particularly strong performance on categories that require understanding of spatial document structure. On \textbf{Multi-Column} layouts, it achieves {87.1\%}, the highest score across all models, suggesting that the early-fusion backbone, which processes image patches with bidirectional attention, is well-suited for capturing cross-column reading order. Similarly, on \textbf{Tables} (90.3\%), FalconOCR ranks at the top, reasonably surpassing Chandra (88.2\%), confirming that the model effectively learns HTML table serialization from the training data. Performance on \textbf{Headers/Footers} ({94.0\%}) and \textbf{Base} documents ({99.5\%}) is near-saturated and on par with the best systems, indicating robust handling of standard document content.

\noindent \textbf{Weaknesses and Limitations.} Two categories reveal clear limitations. On \textbf{OldScan} (43.5\%), FalconOCR lags behind Chandra (49.2\%), Mistral OCR 3 (48.8\%), and Gemini 3 Pro (47.5\%). This gap likely stems from insufficient representation of heavily degraded historical documents that have faded ink, bleed-through, and non-standard typefaces in our training data. On \textbf{TinyText} (78.5\%), the model underperforms Chandra (91.9\%) and Gemini 3 Pro (90.3\%) by a substantial margin. Since FalconOCR operates on cropped regions from the layout detector, very small text elements may arrive at the model with limited pixel resolution, making fine-grained glyph discrimination difficult for a 300M model. Both gaps point to concrete directions for improvement: targeted data augmentation for degraded scans and higher-resolution processing for small text regions.

\subsection{Results on OmniDocBench}

\begin{wraptable}{r}{0.5\textwidth}
    \centering
    \caption{\textbf{OmniDocBench Results:} Performance comparison on full-page document parsing. Overall$\uparrow$ aggregates the three sub-metrics. Edit$\downarrow$ measures text edit distance (lower is better). CDM$\uparrow$ evaluates formula recognition accuracy. TEDS$\uparrow$ measures table structure similarity.\vspace{-0.2cm}}
    \label{tab:omnidocbench_results}
    \setlength{\tabcolsep}{6pt}
    \adjustbox{width=0.49\textwidth}{
    \begin{tabular}{l cccc}
    \toprule
    \textbf{Model} & \textbf{Overall}$\uparrow$ & \textbf{Edit}$\downarrow$ & \textbf{CDM}$\uparrow$ & \textbf{TEDS}$\uparrow$ \\
    \midrule
    PaddleOCR VL 1.5~\citep{cui2026paddleocrvl15}    & 94.37 & 0.075 & 94.4 & 91.1 \\
    PaddleOCR VL~\citep{cui2025paddleocrvl}        & 91.76 & 0.024 & 91.7 & 85.9 \\
    Chandra~\citep{chandra2025}             & 88.97 & 0.046 & 88.1 & 89.5 \\
    DeepSeek OCR v2~\citep{wei2026deepseek}     & 87.66 & 0.037 & 89.2 & 77.5 \\
    GPT 5.2~\citep{openai_gpt52}             & 86.56 & 0.061 & 88.0 & 77.7 \\
    Mistral OCR 3~\citep{mistral_ocr3_2025}       & 85.20 & 0.053 & 84.3 & 76.1 \\
    \midrule
    \textbf{FalconOCR}  & 88.64 & 0.055 & 86.8 & 84.6 \\
    \bottomrule
    \end{tabular}
    }
\end{wraptable}
Table~\ref{tab:omnidocbench_results} presents results on OmniDocBench~\citep{ouyang2025omnidocbench}, where the evaluation shifts from binary correctness checks to continuous metrics over full-page parses. The tables in OmniDocBench are significantly more complex than those in olmOCR, featuring nested structures, multi-level headers, and spanning cells that stress the structural prediction capabilities of each system. 

\noindent \textbf{Overall Performance:} FalconOCR achieves an Overall score of \textbf{88.64}, placing it ahead of DeepSeek OCR~v2 (87.66), GPT~5.2 (86.56), and Mistral OCR~3 (85.20), while trailing the recent top modular pipelines PaddleOCR VL~1.5 (94.37) and PaddleOCR VL (91.76). Table parsing is quite competitive: our TEDS of \textbf{84.6} is close to PaddleOCR VL (85.9) and substantially above DeepSeek OCR~v2 (77.5), GPT~5.2 (77.7), and Mistral OCR~3 (76.1). Text edit distance (\textbf{0.055}) is in the mid-range, reflecting that full-page element matching on complex layouts is more challenging for our two-stage pipeline than for systems that have been specifically tuned for OmniDocBench's evaluation protocol.

\noindent \textbf{Table Parsing and Matching Challenges:} Although our model's TEDS score of \textbf{84.6} trails PaddleOCR VL~1.5 (91.1) and Chandra (89.5), we attribute part of this gap to inherent ambiguities in the TEDS evaluation protocol. HTML table representations are not unique: visually identical tables can be encoded in structurally different ways. For instance, a horizontal row separator can be represented either by a \texttt{rowspan} attribute or by inserting explicit empty cells, both produce the same visual rendering but yield different tree structures under TEDS. We apply some post-processing to normalize our HTML output, but do not attempt exhaustive structural canonicalization, which would require heuristic rules that risk introducing other errors.

\noindent \textbf{Formula Matching Sensitivity:} While our model's CDM of \textbf{86.8} is below the top systems (PaddleOCR VL~1.5 at 94.4, PaddleOCR VL at 91.7), we observe that the element-matching step in OmniDocBench's evaluation pipeline penalizes certain valid predictions. Specifically, when our model outputs a formula together with its equation number as a single block (a common typographical convention), the matching algorithm may fail to align it with the ground truth, which separates the formula from its number. This ``clubbing'' penalty artificially deflates CDM on affected samples. Similarly, character encoding discrepancies, such as predicting the Unicode symbol $\beta$ \vs the \LaTeX{} command \texttt{\textbackslash beta} incur edit penalties even when both representations are semantically correct, as the ground-truth sometimes favors one convention over the other. These matching artifacts affect our model significantly because we perform only minimal output normalization, prioritizing generality over benchmark-specific tuning.

\noindent \textbf{Efficiency--Accuracy Trade-off:} Across both benchmarks, the key takeaway is that FalconOCR achieves competitive accuracy at a fraction of the cost of larger systems. At 300M parameters, with ${\sim}\num{6000}$ tokens/second and an average $2.8$ images/s throughput using an efficient vLLM implementation, FalconOCR occupies a favorable point on the efficiency--accuracy frontier: it is the smallest model, yet remains competitive on olmOCR (80.3\%, top-3) and OmniDocBench (88.64, top-4), making it a practical choice for large-scale document digitization.

%% file: sections/discussion.tex
\section{Discussion}
\label{sec:discussion}

Falcon Perception is built around a simple idea, \ie, a single early-fusion Transformer with the right interface is enough for dense perception tasks. We keep one scalable backbone and move complexity to training signal and sequence interface. The results in Section~\ref{sec:experiments} suggest this direction is viable, especially when prompts require OCR, spatial constraints, relations, or long-context dense outputs.

\noindent\textbf{A ``bitter lesson'' view of open-vocabulary segmentation:}
In this field, it is very tempting to solve every failure mode with a new module: a stronger vision encoder, an extra fusion block, a new matching trick, or more post-processing. This can work, but it also makes systems harder to scale and harder to reason about. Falcon Perception is ``bitter'' on purpose: one backbone, one training objective family, and small heads only where outputs are continuous and dense. The hope is that most improvements should come from more data, more compute, and better training signals, not from making the pipeline more complicated.

\noindent\textbf{A scalable interface:}
Dense perception is not a fixed-size prediction problem. The number of instances can be small or extremely large, and the prompt can ask for simple objects or complicated prompts requiring reasoning. Autoregressive generation gives a clean interface for this: the model can emit as many instances as needed, and the sequence length becomes the knob for ``how much perception'' we want. This matters for crowded scenes. If we want to push to $K\!\sim\!1000$ masks, we do not need a new model family; we need to make long-context generation stable and efficient. In our design, we keep generation cheap by emitting only a few task tokens per instance, while producing masks in parallel via the segmentation head.

\noindent\textbf{Early fusion is enough:}
Existing open-vocabulary systems  rely on a separate vision encoder and a late fusion stage. Our results support a different view: if image tokens and text tokens share the same backbone from the first layer, the model can learn the interaction just fine. This is especially useful for OCR-guided and spatial prompts, where the prompt should influence feature formation, not only the final decoder. In Falcon Perception, we still use specialized heads, but only for decoding coordinates, size, and masks efficiently. The backbone doing all the main localization work stays one dense model.

\noindent\textbf{Scaling path: patchify and mix data without redesign:}
Because the model is a Transformer over tokens, the scaling is straightforward. We can improve visual grounding by adding more images and harder prompts, and we can improve language knowledge by mixing text-only data, captioning data, or interleaved vision-language data. Nothing in the architecture blocks this: it is still one sequence model. 

\noindent\textbf{RL and sampling: the model already contains the solution:}
Sampling is a strong feature of an autoregressive design. Our Pass@$k$ results show that the model often has correct localizations in its distribution, but greedy decoding does not pick them. This looks similar to LLMs before RL post-training: maximum likelihood learns a rich distribution, but the probability mass is not shaped for the decision we want. This behavior is reminiscent of DeepSeek-R1~\citep{guo2025deepseek}, where the ``raw'' model already contains high-quality solutions. RL-style post-training can then reshape the distribution by rewarding the desired outcome, effectively pulling up correct but lower-probability predictions. This is aligned with risk-based training views such as~\citep{pinto2023tuning}. We intentionally evaluate up to Pass@$8$ since group-based RL methods (e.g., GRPO-style) commonly use similar group sizes. The fact that Pass@$k$ helps across levels suggests we can form candidate groups with real discriminative signal using Falcon Perception.

\noindent\textbf{Limitations and what we expect next:}
A single-stack autoregressive model is not free: training is more expensive and decoding can be slower than a fully parallel DETR like models. But the upside is a clean scaling interface. Going forward, we expect most gains to come from: (i) better data mixtures (including text-only, interleaved and captioning when needed), (ii) longer-context training for dense scenes, and (iii) post-training (RL) to improve selection of the right prediction from the model's own distribution.

%% file: sections/appendix.tex
\appendix

\setcounter{table}{0}
\setcounter{figure}{0}
\renewcommand{\thetable}{A.\arabic{table}}
\renewcommand{\thefigure}{A.\arabic{figure}}

\section{Metrics\label{appendix:metrics}}
\paragraph{Localization -- positive micro F$_1$ (pmF$_1$):}
For a datapoint with $N$ predicted masks $\{m_i\}_{i=1}^N$ and $M$ ground-truth masks $\{\hat m_j\}_{j=1}^M$, we compute the IoU matrix
\begin{equation}
\mathrm{IoU}_{ij} = \mathrm{IoU}(m_i, \hat m_j).
\end{equation}
We then compute an optimal bipartite assignment (Hungarian matching) $\sigma$ maximizing total IoU. For an IoU threshold $\tau$, a prediction $i$ is a true positive if it is matched and $\mathrm{IoU}_{i,\sigma(i)} \ge \tau$; otherwise it is a false positive. Unmatched ground-truth masks are false negatives. Let $TP_\tau$, $FP_\tau$, and $FN_\tau$ denote the resulting counts. We compute a micro-F$_1$ at threshold $\tau$ by accumulating these counts across all \emph{positive} datapoints (those with $M>0$),
\begin{equation}
\mathrm{pmF_1}_\tau =
\frac{2\,TP_{\tau,\mathrm{total}}}{2\,TP_{\tau,\mathrm{total}} + FP_{\tau,\mathrm{total}} + FN_{\tau,\mathrm{total}}}.
\end{equation}
Following SAM3, we evaluate at 10 thresholds $\tau \in \{0.50, 0.55, \dots, 0.95\}$ and average:
\begin{equation}
\mathrm{pmF_1} = \frac{1}{10}\sum_{\tau}\mathrm{pmF_1}_{@\tau}.
\end{equation}
We also report the per-threshold values (\eg, $\mathrm{pmF_1}_{@0.50}$ and $\mathrm{pmF_1}_{@0.75}$) when relevant.

\paragraph{Classification -- image-level MCC (IL\_MCC):} We adopt the same image-level presence classification metric as SAM3: we treat each datapoint as a binary decision (object present vs.\ not present) based on whether the model predicts any mask, and summarize the resulting confusion matrix with the Matthews Correlation Coefficient (IL\_MCC). We refer readers to SAM3 for the exact definition and implementation details.

\paragraph{Combined -- classification-gated F$_1$ (cgF$_1$):}
Combining localization and classification:
\begin{equation}
\mathrm{cgF_1} = \mathrm{pmF_1}\cdot IL\_\mathrm{MCC}.
\end{equation}

\paragraph{Macro-F$_1$ (per-sample):}
In addition to micro aggregation, we report a macro variant computed as the mean of per-sample scores. For positive samples, the per-sample score is the mean local F$_1$ across thresholds; for true negatives (no ground truth and no prediction) the per-sample score is set to $1$, while false positives and false negatives receive $0$. This metric highlights per-image stability rather than being dominated by high-mask-count samples.

\section{Architecture Study}
\label{sec:arch}

\begin{figure}[t]
\centering

\begin{minipage}{0.72\linewidth}
\centering
\includegraphics[width=\linewidth, trim=7cm 13.6cm 7cm 0, clip]{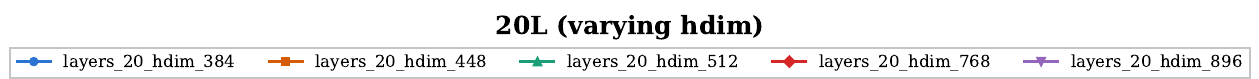} \\
\includegraphics[width=\linewidth, trim=0 0 0 2.5cm, clip]{figures/sweeps/det_20l_hdim_macro_F1}
\end{minipage}
\hfill
\begin{minipage}{0.25\linewidth}
\centering
{\footnotesize
\captionof{table}{Metrics averaged over the seven SaCo splits at 60k steps.\vspace{-0.3cm}\label{tab:appendix_width}}}
\setlength{\tabcolsep}{6pt}
\adjustbox{width=\textwidth}{
\begin{tabular}{cc|cc}
\toprule
$L$ & $d_{\text{model}}$ & MCC & Macro F$_1$ \\
\midrule
20 & 384  & 46.4 & 55.7 \\
20 & 448  & 48.2 & 57.5 \\
20 & 512  & 49.8 & 58.5 \\
20 & 768  & \textbf{53.2} & \textbf{61.5} \\
20 & 896  & 52.4 & 60.9 \\
\bottomrule
\end{tabular}
}
\end{minipage}
\caption{\textbf{Impact of architectural choices:} Width scaling at fixed depth ($L{=}20$). \label{fig:appendix_arch_width}}
\end{figure}

\begin{figure}[t]

\begin{minipage}{0.72\textwidth}
\centering
\includegraphics[width=\linewidth, trim=7cm 13.6cm 7cm 0, clip]{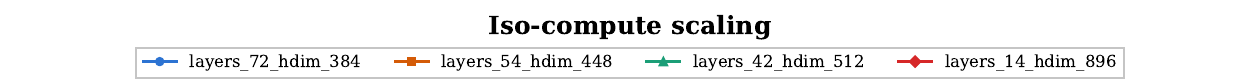} \\
\includegraphics[width=\linewidth, trim=0 0 0 2.5cm, clip]{figures/sweeps/det_iso_compute_macro_F1.pdf}
\end{minipage}
\hfill
\begin{minipage}{0.25\textwidth}
\centering
{\footnotesize
\captionof{table}{Metrics averaged over the seven SaCo splits at 60k steps.\vspace{-0.3cm}\label{tab:appendix_isocompute}}}
\adjustbox{width=\textwidth}{
\begin{tabular}{cc|cc}
\toprule
$L$ & $d_{\text{model}}$ & MCC & Macro F$_1$ \\
\midrule
72 & 384   & 52.4 & 60.5 \\
54 & 448   & 50.0 & 58.8 \\
42 & 512   & \textbf{54.0} & \textbf{61.9} \\
20 & 768   & 53.2 & 61.5 \\
14 & 896   & 50.8 & 59.4 \\
\bottomrule
\end{tabular}
}
\end{minipage}

\caption{\textbf{Impact of architectural choices:} Iso-compute depth--width trade-off.}
\label{fig:appendix_arch_isocompute}

\end{figure}

A central design question for unified dense-prediction architectures is how to
distribute model capacity between depth (number of layers~$L$) and width
(hidden dimension~$d_{\text{model}}$).  Because visual features and
autoregressive task tokens share the same transformer stack, there is a risk
of \emph{modality crowding}: the channel bandwidth may become a bottleneck
when both modalities compete for representational capacity.  We investigate
this through two controlled studies on the SaCo benchmark, evaluating
bounding-box detection at an IoU threshold of~$0.75$ and reporting the average
MCC and Macro~F$_1$ across all seven splits after 60k training steps.

\paragraph{Hyperparameter transfer:}
To ensure that performance differences reflect architectural capacity rather
than optimization artifacts, we adopt the Maximal Update Parameterization
($\mu$P)~\citep{yang2019wide} for learning-rate transfer across widths.  We
first tune the optimal learning rate on a reference configuration
(20L\,/\,$d_{\text{model}}{=}768$) and transfer it to every other width via
\begin{equation}
  \mathrm{lr}_{\text{target}}
  = \mathrm{lr}_{\text{ref}}
    \cdot \sqrt{\frac{d_{\text{target}}}{d_{\text{ref}}}}\,,
  \label{eq:mup}
\end{equation}
so that the effective magnitude of weight updates remains consistent
regardless of hidden dimension.  All other hyperparameters are kept identical
across configurations.

\subsection{Width Scaling at Fixed Depth}
\label{sec:width}

We fix the network depth at $L{=}20$ layers and vary the hidden dimension
across $d_{\text{model}} \in \{384,\, 448,\, 512,\, 768,\, 896\}$.
Results are summarized in Table~\ref{tab:appendix_width} and Figure~\ref{fig:appendix_arch_width}.
Performance improves monotonically from $d_{\text{model}}{=}384$
(MCC\,$=$\,46.4, F$_1$\,$=$\,55.7) to $d_{\text{model}}{=}768$
(MCC\,$=$\,53.2, F$_1$\,$=$\,61.5), a gain of \textbf{+6.8}~MCC and
\textbf{+5.8}~Macro~F$_1$.  Increasing width further to~896 yields a slight
regression (MCC\,$=$\,52.4), indicating that at 20~layers the architecture
saturates around $d_{\text{model}} \approx 768$.  The sharp degradation at
small widths is consistent with a capacity-bottleneck effect: when visual and
language features must share a narrow hidden state, neither modality is
adequately represented.

\subsection{Depth--Width Trade-off at Fixed Compute}
\label{sec:isocompute}

To disentangle width from overall scale, we fix an approximate compute budget
of 300M parameters and sweep the aspect
ratio across five configurations spanning a~$5{\times}$ range in depth.
Results are shown in Table~\ref{tab:appendix_isocompute} and
Figure~\ref{fig:appendix_arch_isocompute}.
The two balanced configurations: 42L\,/\,512 (MCC\,$=$\,54.0,
F$_1$\,$=$\,61.9) and 20L\,/\,768 (MCC\,$=$\,53.2,
F$_1$\,$=$\,61.5) achieve the highest scores and lie within one~MCC point
of each other.  Both substantially outperform the extremes: the
deepest-narrowest variant (72L\,/\,384: MCC\,$=$\,52.4) and the
shallowest-widest variant (14L\,/\,896: MCC\,$=$\,50.8) each trail the best
configuration by 2--4~points.  Notably, the 54L\,/\,448 model
(MCC\,$=$\,50.0) under\-performs 72L\,/\,384 despite having a wider hidden
state, suggesting that this particular depth--width combination falls into an
unfavorable operating regime.

\subsection{Discussion}

Both studies converge on two findings:
\begin{enumerate}[nosep,leftmargin=*]
  \item \textbf{Width is necessary but not sufficient:}
        Study~A shows that widening the hidden dimension from 384 to 768
        yields a +6.8~MCC improvement at fixed depth, confirming that
        sufficient channel capacity is critical for dense prediction in a
        unified stack.  However, gains saturate beyond
        $d_{\text{model}} \approx 768$.

  \item \textbf{Balanced aspect ratios are optimal under iso-compute:}
        Study~B shows that, for a fixed compute budget, moderate
        configurations (42L\,/\,512 and 20L\,/\,768) outperform both
        deep-narrow and shallow-wide extremes by 2--4~MCC.  This indicates
        that representational diversity from depth and channel capacity from
        width are independently necessary and neither alone compensates for a
        deficit in the other.
\end{enumerate}
Taken together, these results suggest that the unified single-stack
architecture does not inherently suffer from modality crowding.  The optimal operating point in our
explored range is 42L\,/\,512, though 20L\,/\,768 offers a competitive
alternative at a shallower depth that may be preferable when inference latency
is a consideration.

\section{Qualitative Analysis}

\subsection{Image Feature Quality}
\begin{figure}[t]
  \centering

\begin{minipage}{0.5\textwidth}
\centering
\includegraphics[width=\textwidth, trim=0 120cm 0 0, clip]{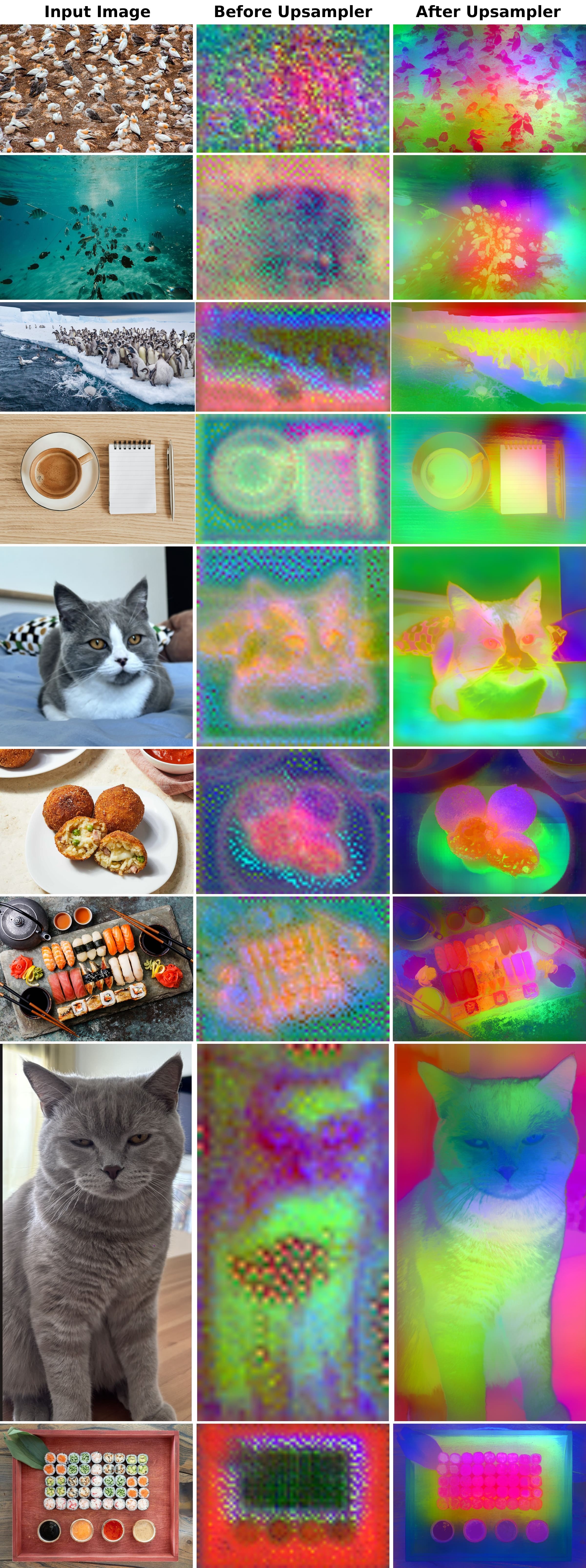}
\end{minipage}
\hfill
\begin{minipage}{0.49\textwidth}
\centering
\includegraphics[width=1\textwidth, trim=0 223.5cm 0 0, clip]{figures/pca_comparisons.jpg}\\
\includegraphics[width=0.95\textwidth, trim=0 0 0 110cm, clip, height=0.392\textheight]{figures/pca_comparisons.jpg}
\end{minipage}
\vspace{-0.1cm}
\caption{\textbf{PCA maps before and after upsampling:} Image features before upsampling already distinguish multiple objects, \eg, the \textit{coffee cup, notebook}, and \textit{pen} in the fourth row on the left. Similarly, \textit{chopsticks} along with various food items in second row on the right and the four \textit{cups} in the bottom right row can be distinguished. The feature upsampler enhances these representations, making instances more visible, suggesting that the pre-upsampling features capture sufficient scene information for instance segmentation and demonstrating the effectiveness of the early-fusion design.\vspace{-0.2cm}}
\label{fig:appendix_pca}
\end{figure}

Here, we assess the quality of the image features output by our model, which employs early fusion of the image and text tokens. Figure~\ref{fig:appendix_pca} shows the PCA maps of the image features at the Falcon Perception output layer (before the AnyUp feature upsampler) and after being upsampled by the upsampler. We observe that our model learns to distinguish between objects in the image to a reasonable extent, as evident in the image features before upsampling (\eg, the \textit{coffee cup, notebook} and \textit{pen} in the fourth row on the left are clearly distinguishable in the image features before upsampling). Furthermore, at the feature upsampler output, the image features are enhanced with high-fidelity and the instances are better visible, indicating that the features output by our model before upsampling encompass sufficient scene information that is required for promptable instance segmentation. These results show the effectiveness of early fusion design in our Falcon Perception.

\subsection{Varying the upsampling factor\label{appendix:upsampling_factor}}
\begin{figure}[t]
  \centering
  \includegraphics[width=1.0\textwidth]{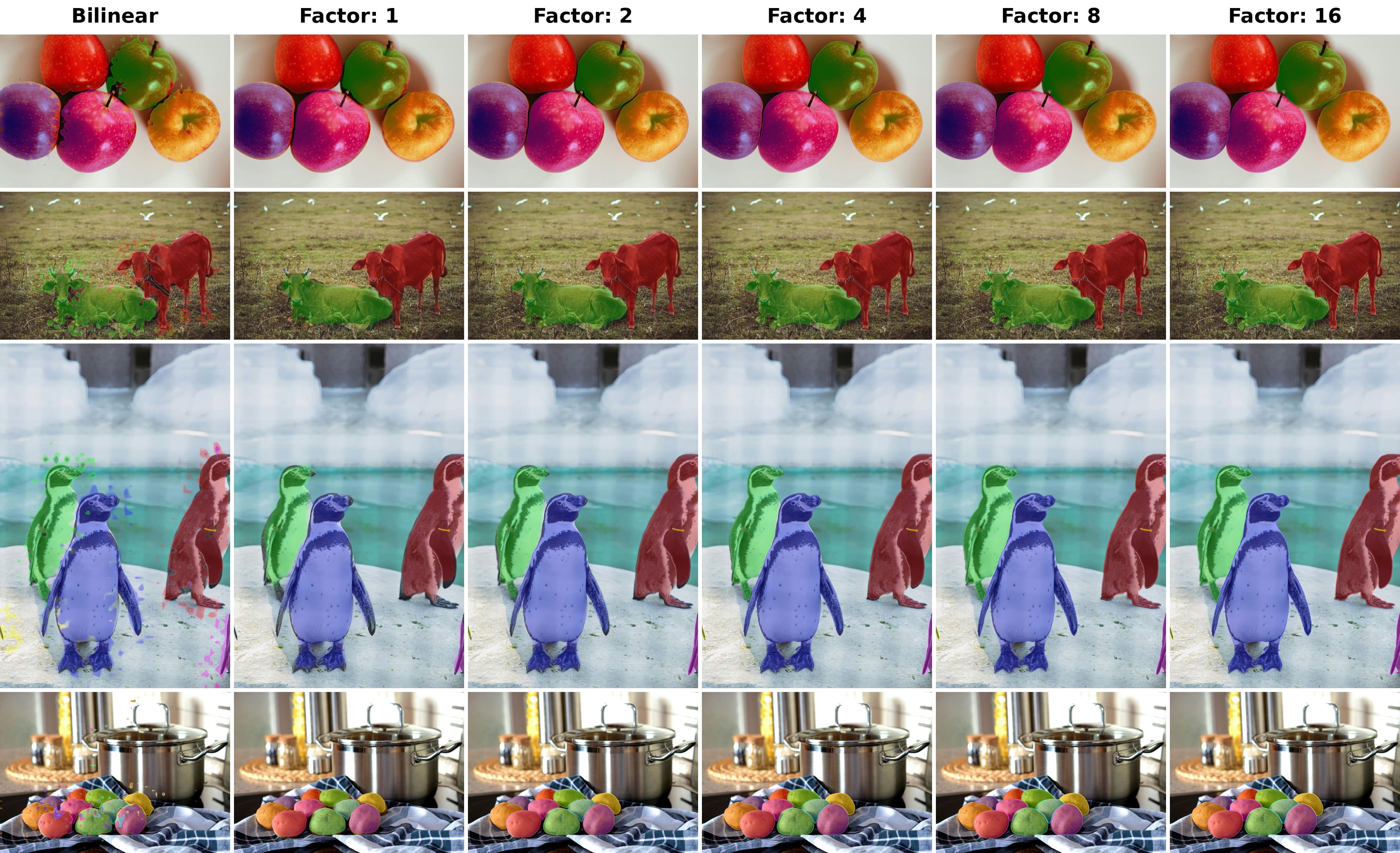}\vspace{-0.2cm}
\caption{\textbf{Impact of the upsampling factor during inference:} Bilinearly upsampling the image features at the perception model output (16x downsampled) and computing the instance masks results in degraded quality (leftmost column) with holes inside the instances and mask regions outside the instances. Processing the model output via the Anyup upsampler with different upsampling factors, followed by mask computation and bilinear upsampling of mask to image size significantly improves the mask quality (columns 2 to 6). Particularly, as the Anyup upsampling factor is increased, the instance boundaries are better refined. \Eg, in the top row, the boundaries of the fourth apple instance from left improves as the factor increases to 8 and 16. Similarly, in the third row, the beak of left penguin is incomplete for factors 1 to 4, while it is accurate for higher factors.\vspace{-0.2cm}}
\label{fig:appendix_impact_factor}
\end{figure}

Here, we qualitatively analyze the effect of the upsampling factor on the mask quality. We compare the output instance masks generated with different impact factors in Figure~\ref{fig:appendix_impact_factor}. Our Falcon Perception uses a patch size of 16 for the image patchification, resulting in 16$\times$ downscaled image features at the model output before mask computation. In order to increase the resolution of the image features, we use the AnyUp upsampler, whose upscaling factor can be varied to compute masks at different image resolutions. Note that the final masks are bilinearly interpolated to image size, for upsampling factors less than 16. As a baseline, we use bilinear upsampling (16$\times$) technique (left column in Figure~\ref{fig:appendix_impact_factor}), which results in masks that have holes inside objects and artifacts outside. The effectiveness of using the feature upsampler is clearly seen even when the upscaling factor is set to 1 (second column from left), where the mask quality significantly improves over the baseline. Furthermore, as the factor is increased, we observe mask refinement with improvements at the instance boundaries. This is noticeable in the first row of the figure, where the boundaries of the fourth apple instance from left improves for upsampling factor $\ge$8. Similarly, in the third row of the figure, the mask of left penguin is incomplete with its beak missing partially for factors 1 to 4, while it is accurate for higher factors. 

\subsection{Qualitative Comparison with SAM 3}
\begin{figure}[!t]
  \centering
  \includegraphics[width=\textwidth, trim=9cm 9cm 9cm 9cm, clip]{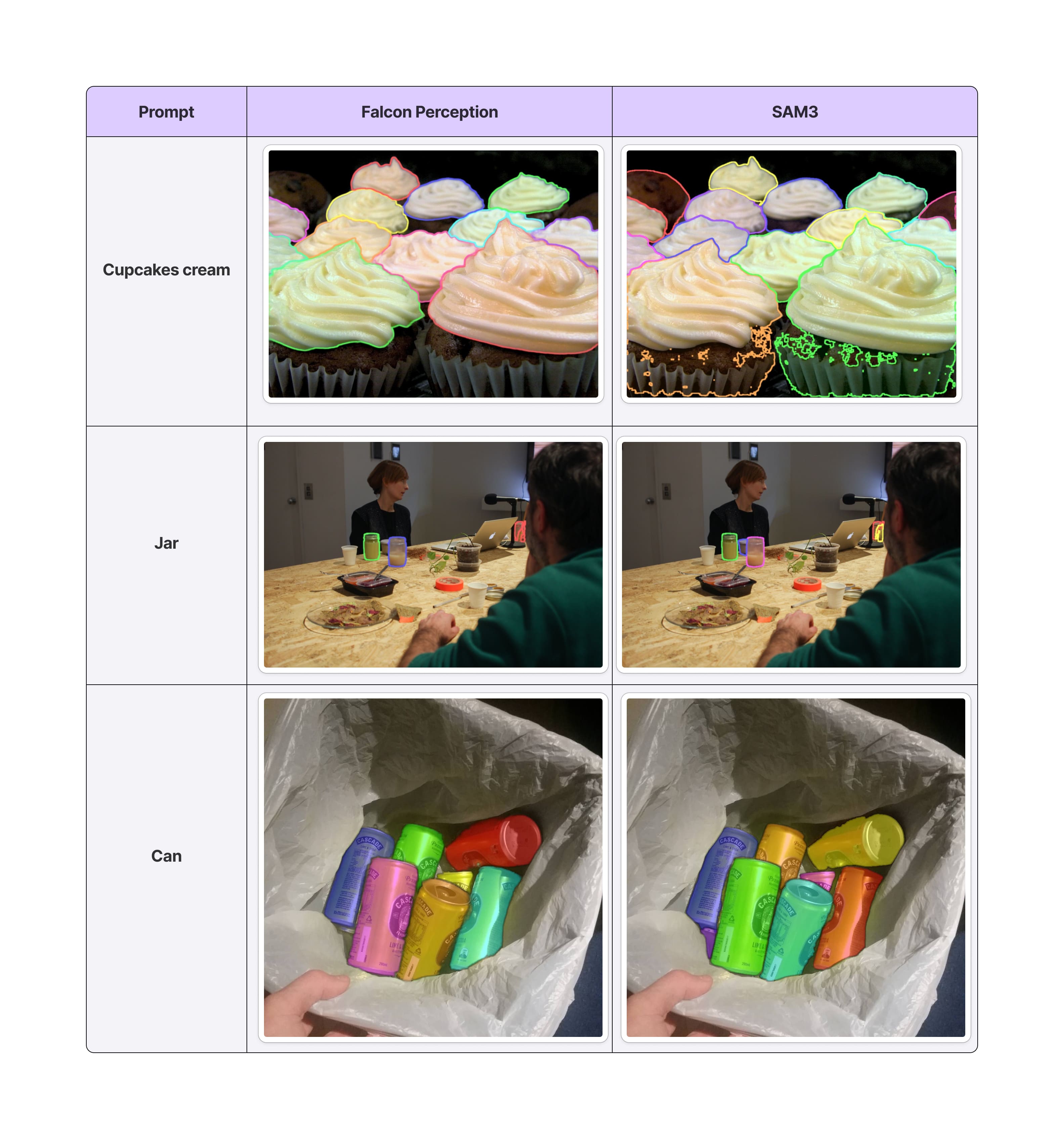}
  \caption{\textbf{Level 0:} While our Falcon Perception predicts correct masks, SAM 3 incorrectly predicts cupcakes in the masks for the \textit{cupcakes cream} example on top row and an additional \textit{jar} in the middle (middle row).\vspace{-0.5cm}}
  \label{fig:level_0}
\end{figure}

\begin{figure}[t]
  \centering
  \includegraphics[width=1.0\textwidth, trim=9cm 9cm 9cm 9cm, clip]{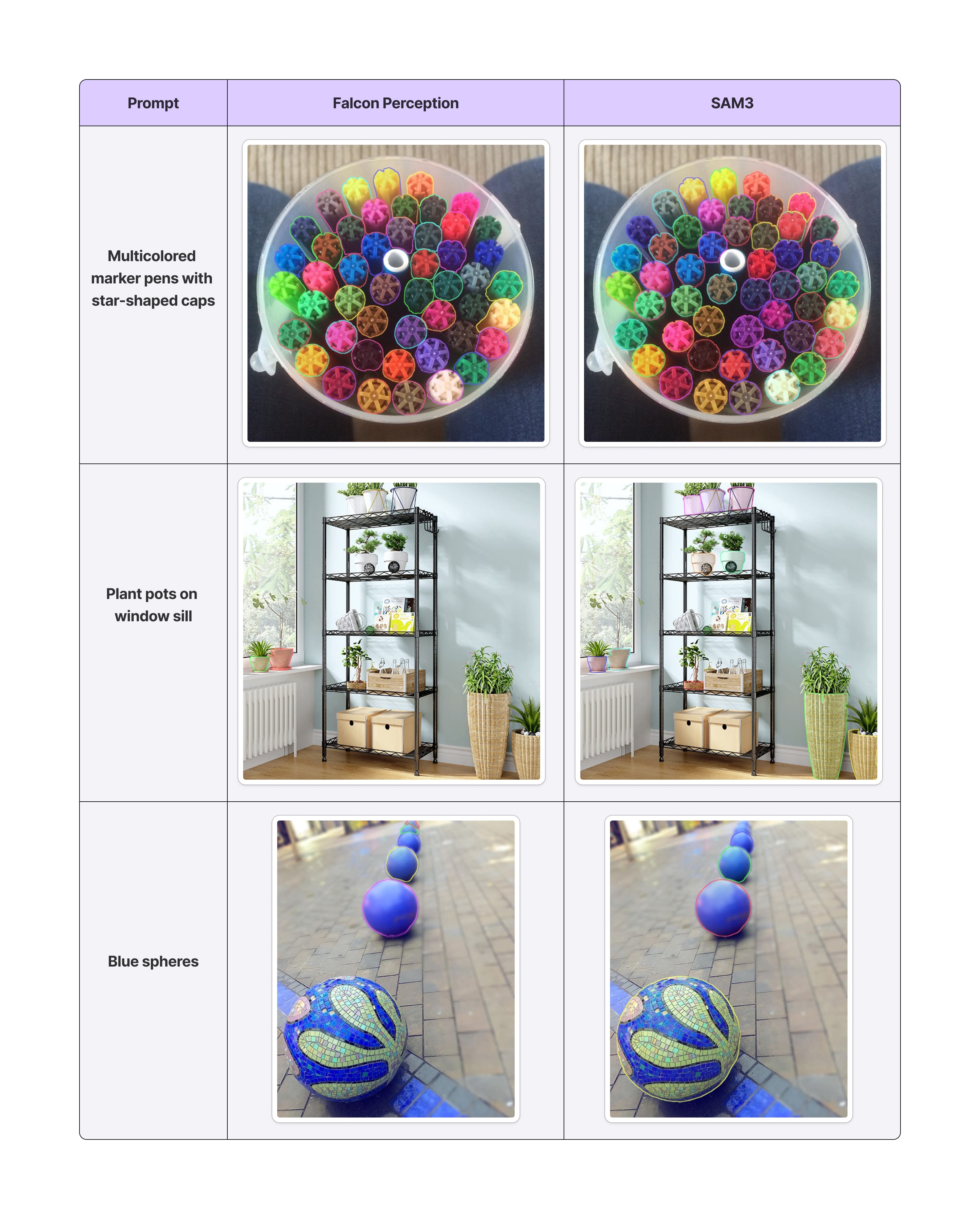}
  \caption{\textbf{Level 1:} Our Falcon perception misses few marker pens in the top row, clearly distinguishes the plant pots on the window sill from the other plant pots in the middle row. In contrast, SAM 3 identifies all the marker pens correctly but fails to distinguish between plant pots in the middle row by predicting false positives for pots not placed on window sill.}
  \label{fig:level_1}
\end{figure}

\begin{figure}[t]
  \centering
  \includegraphics[width=1.0\textwidth, trim=9cm 9cm 9cm 9cm, clip]{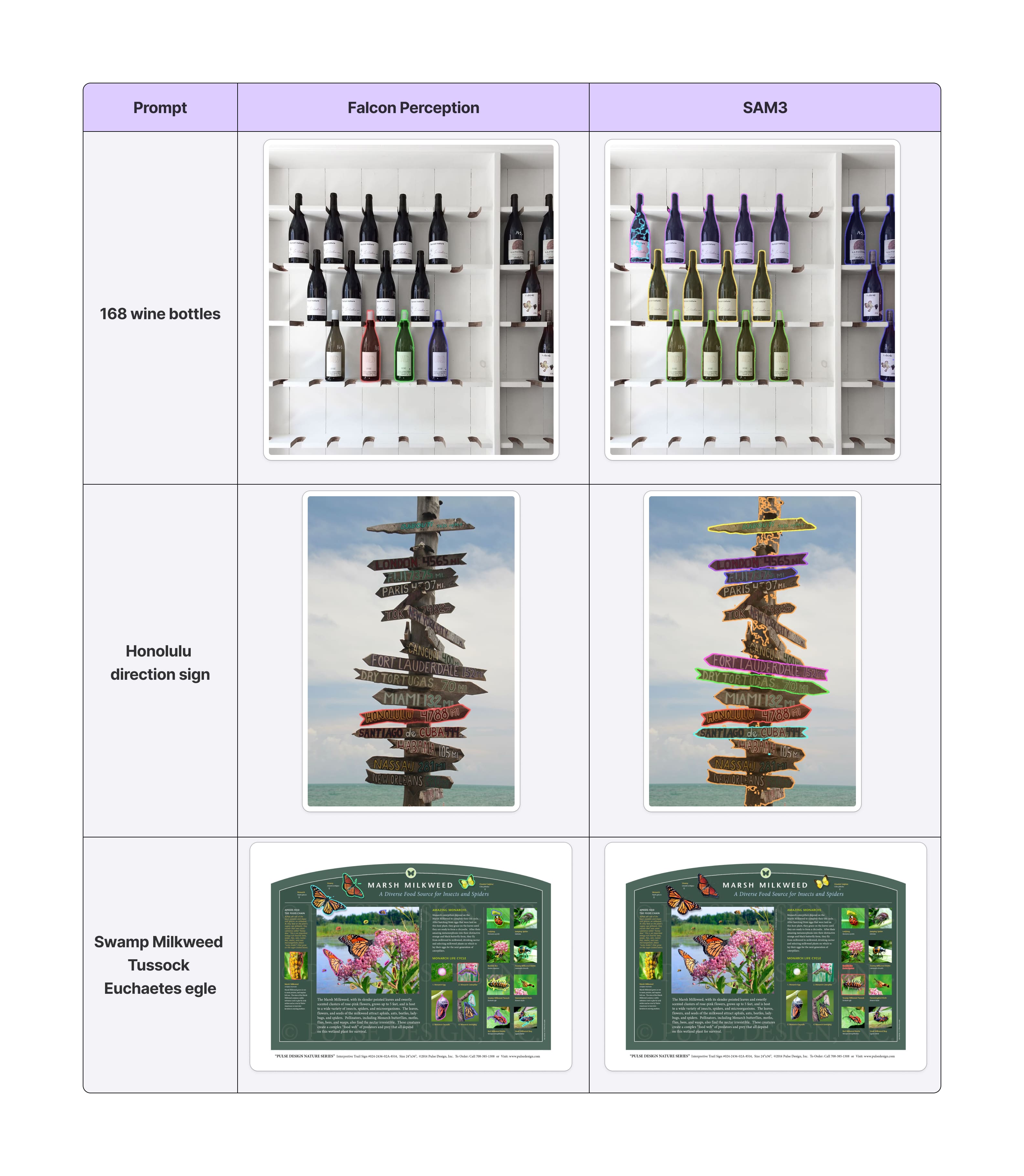}
  \caption{\textbf{Level 2:} Different to SAM 3 that incorrectly predicts masks for each bottle (top row) and each sign (middle row), our Falcon Perception model correctly identifies the \textit{168 wine bottles} (marked with 168) and the \textit{Honolulu direction sign}, indicating that our model can better relate the query prompt to the OCR text in the image. }
  \label{fig:level_2}
\end{figure}

\begin{figure}[t]
  \centering
  \includegraphics[width=1.0\textwidth, trim=9cm 9cm 9cm 9cm, clip]{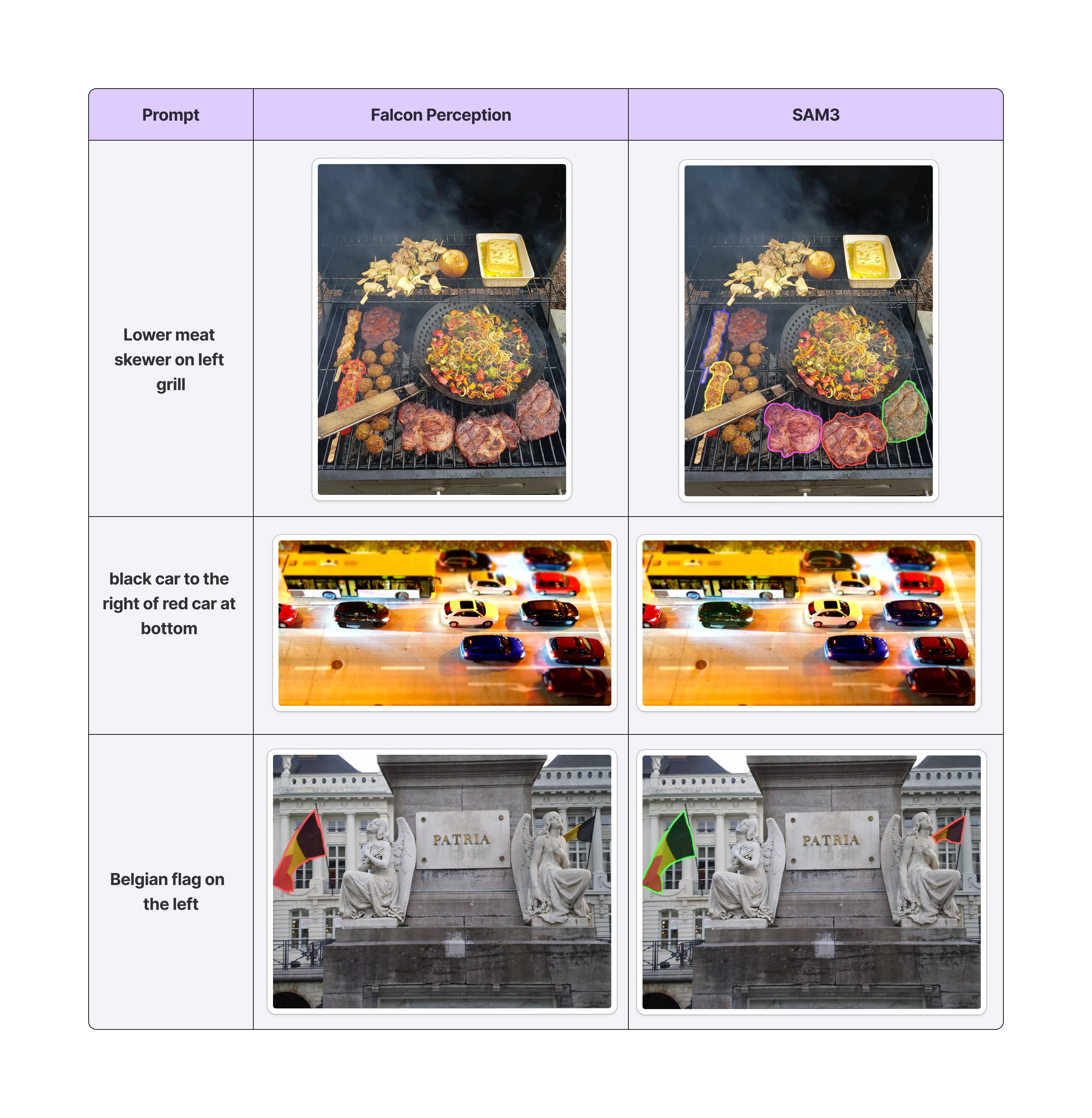}
  \caption{\textbf{Level 3:} In contrast to SAM 3 predicting false positives for meat (top row), black car (middle row) and Belgian flag (bottom row), our model correctly identifies the \textit{lower meat skewer on left grill}, {black car to the right of red car at bottom} and \textit{Belgian flag on the left} in the three examples. This shows that our Falcon Perception has better spatial understanding of the scene.}
  \label{fig:level_3}
\end{figure}

\begin{figure}[t]
  \centering
  \includegraphics[width=1.0\textwidth, trim=9cm 9cm 9cm 9cm, clip]{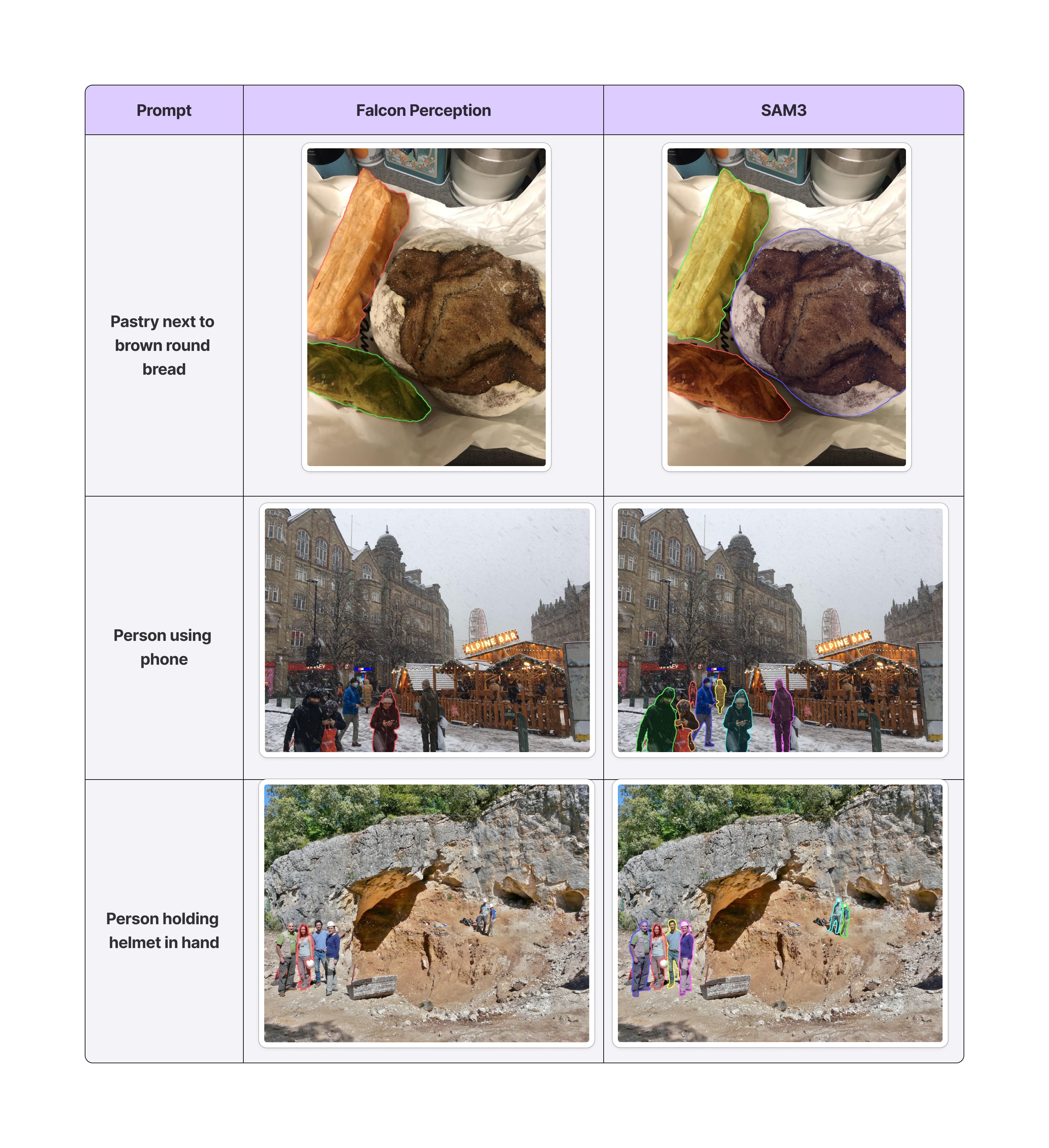}
  \caption{\textbf{Level 4:} SAM 3 wrongly outputs masks for all bread (top row), persons (middle and bottom rows). However, our Falcon Perception model correctly identifies the object instances for \textit{pastry next to brown round bread}, \textit{person holding phone} and \textit{person holding helmet in hand}, thereby showcasing the ability to understand the object interaction in the scene.}
  \label{fig:level_4}
\end{figure}

\begin{figure}[t]
  \centering
  \includegraphics[width=1.0\textwidth, trim=9cm 9cm 9cm 9cm, clip]{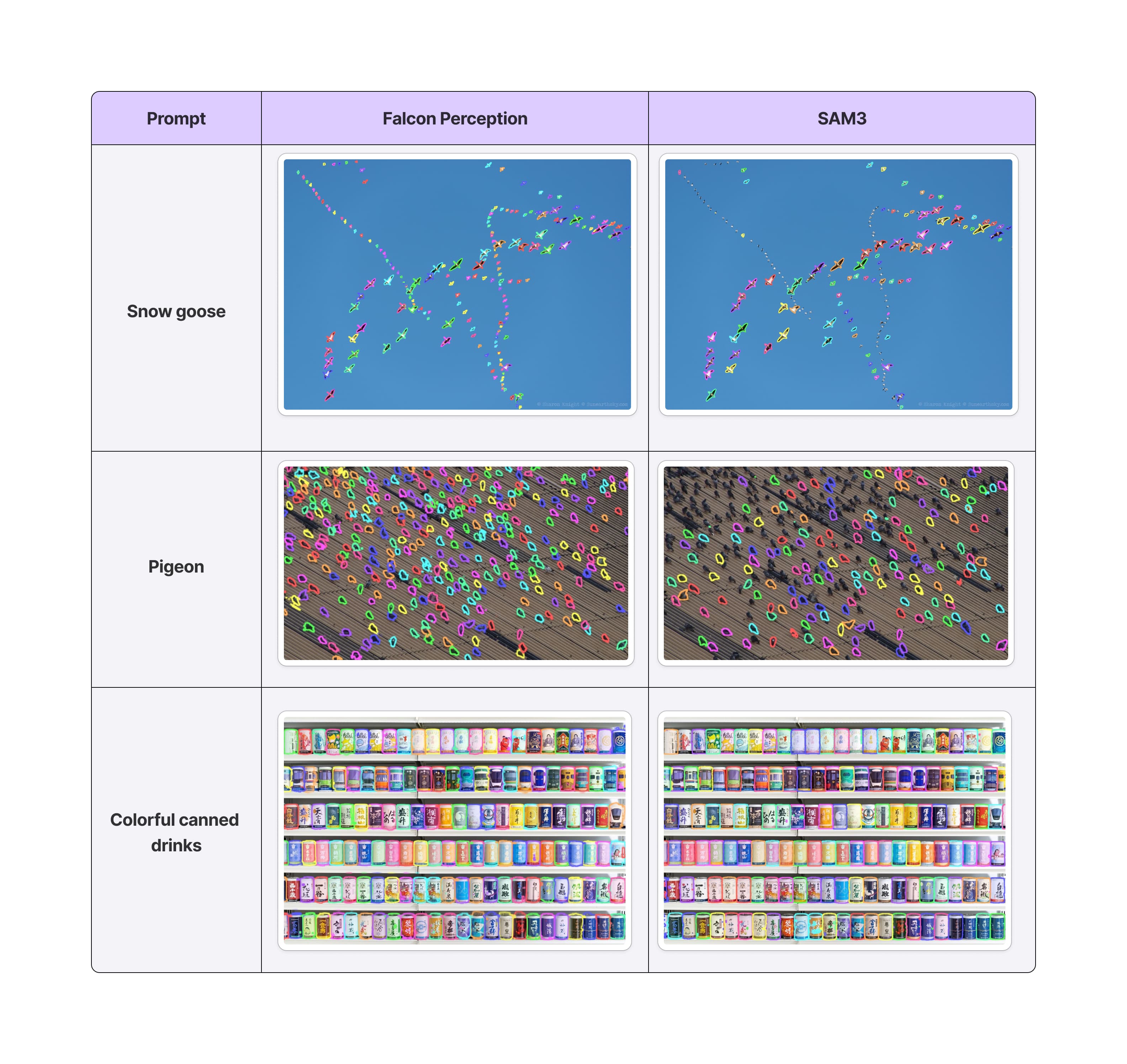}
  \caption{\textbf{Dense Split:} While SAM 3 is capable of identifying many instances of an object, it fails to predict masks when number of instances exceeds 200 due to the fixed number of query tokens in the decoder. Our Falcon Perception, following the autoregressive approach for prediction, is not limited by this and can easily scale higher, as shown for \textit{snow goose}, \textit{pigeon} and \textit{colorful canned drinks} in the three rows.}
  \label{fig:dense}
\end{figure}

Figures~\ref{fig:level_0}-\ref{fig:dense} show the qualitative comparison of the predicted masks between our Falcon Perception and SAM~3\footnote{While we acknowledge that SAM~3 cannot reliably predict masks for levels 2 to 4 since it is not trained in that manner, we compare with it to showcase the added capability of our Falcon Perception in this regard.} on various image-prompt pairs with  prompts varying from Level 0 to 4.  Overall, SAM 3 incorrectly predicts false positives when query prompts requiring visual text recognition (Figure~\ref{fig:level_2}), complex expression prompts are used (Figures~\ref{fig:level_3} and ~\ref{fig:level_4}) and is limited by the number of query tokens in its DETR decoder for dense predictions (Figure~\ref{fig:dense}). In contrast, our Falcon Perception can read text reasonably well, decipher complex expressions and predict the correct instances among other distractors in the scene. Furthermore, since our model's predictions are autoregressive in nature, once trained with such long-contexts, it can easily scale the detections to many instances (up to 600). These results show the added capabilities for our dense transformer based early-fusion perception model. 

\subsection{Effect of Sampling\label{sec:appendix_sampling}}
\begin{figure}[t]
  \centering
  \includegraphics[height=0.88\textheight, trim=4.5cm 4.5cm 4.5cm 4.5cm, clip]{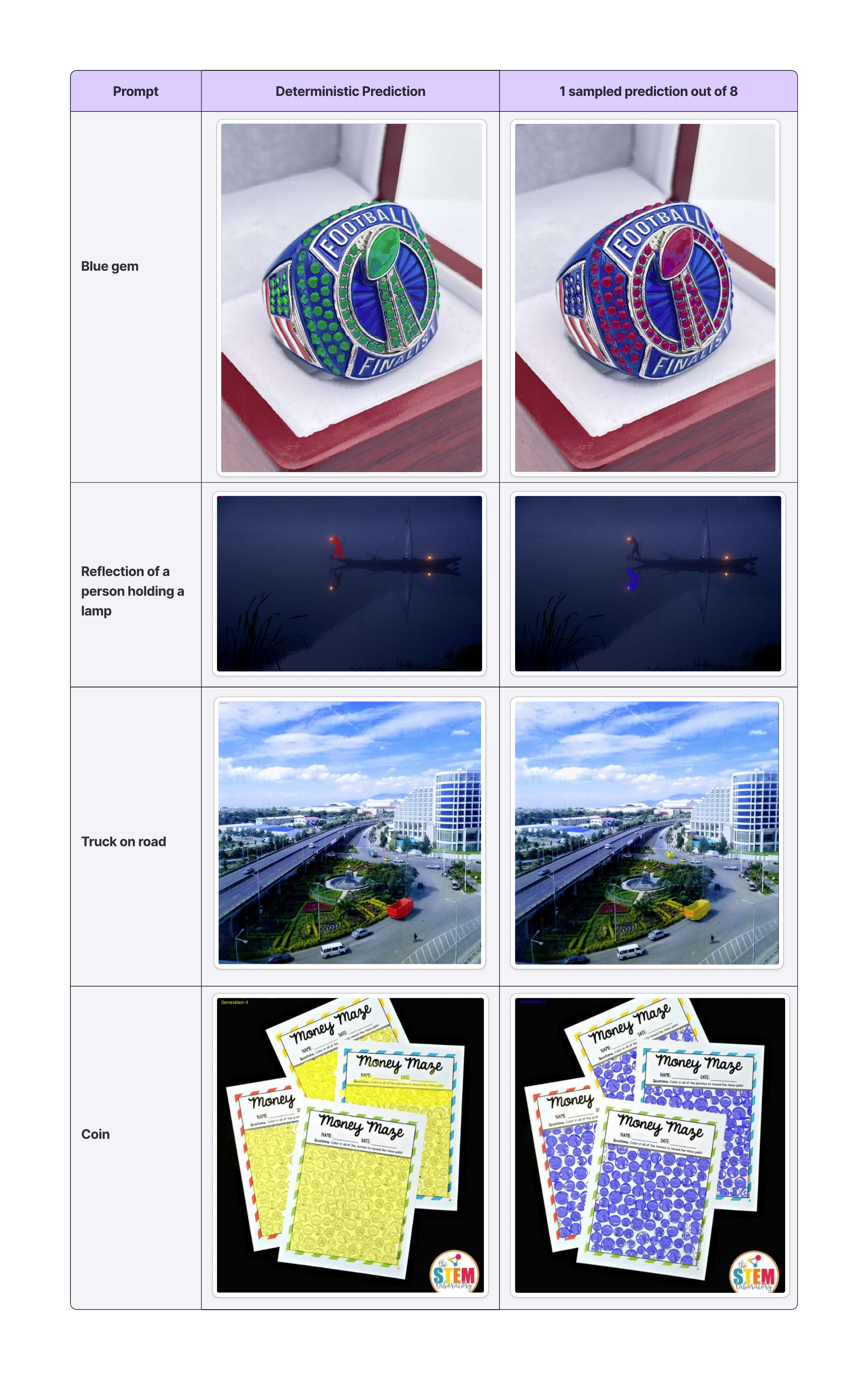}
  \caption{\textbf{Effect of sampling on the quality of predictions:} In the left column, we show the deterministic predictions, selecting the most likely next token, center coordinate, or size value. On the right, we show one of the predictions we obtain by sampling tokens, center, and size with a temperature of $0.7$. Oftentimes, we can achieve better predictions for instances that require reasoning over the query, fine-grained perception, and small or distant objects.}
  \label{appendix:sampling}
\end{figure}

Figure~\ref{appendix:sampling} shows the effect of sampling tokens, center, and size with higher temperature, compared to greedy sampling. We observe that the quality of predictions improves with non-deterministic sampling (\eg, \textit{reflection of person} in the 2$^{nd}$ row) and even smaller scale instances are identified (\eg, \textit{truck on road} in the 3$^{rd}$ row). This, along with the results in Section~\ref{sec:sampling} shows that our model has the required reasoning to predict the instances correctly, and with RL-style post-training, it can perform better.

\clearpage

\section{Qualitative Analysis - OCR}
\begin{figure}[t!]
  \centering
  \includegraphics[width=1.0\textwidth, trim=4.5cm 4.5cm 4.5cm 4.5cm, clip]{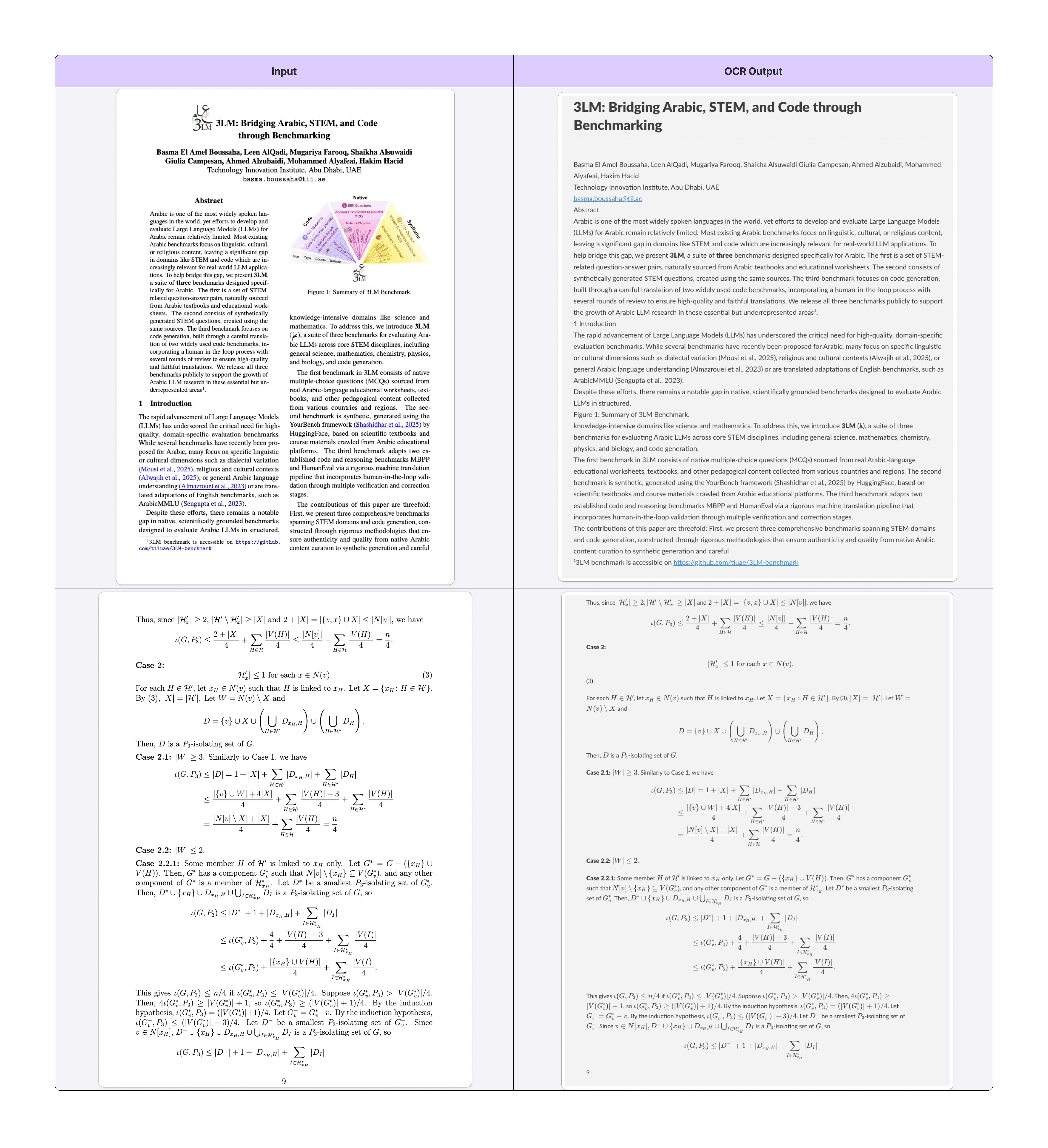}
  \caption{FalconOCR outputs for samples with arxiv papers, scientific formulae. Best viewed zoomed-in.}
  \label{fig:paper2_ocr}
\end{figure}

\begin{figure}[thb]
  \centering
  \includegraphics[width=1.0\textwidth, trim=4.5cm 4.5cm 4.5cm 4.5cm, clip]{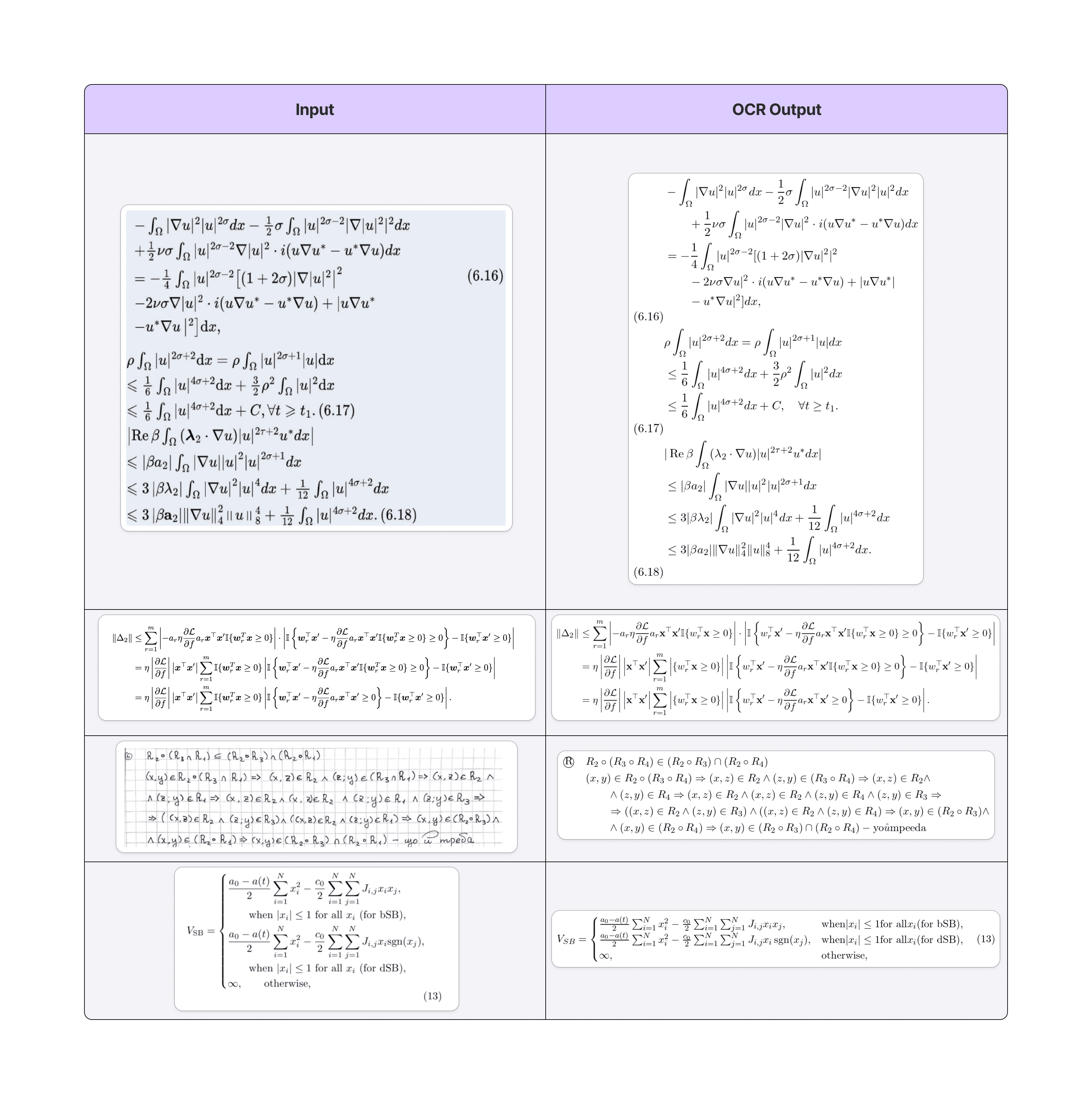}
  \caption{Example outputs for inputs with variations in structure of equations. Best viewed zoomed-in.}
  \label{fig:formulas}
\end{figure}

\begin{figure}[thb]
  \centering
  \includegraphics[width=1.0\textwidth, trim=4.5cm 4.5cm 4.5cm 4.5cm, clip]{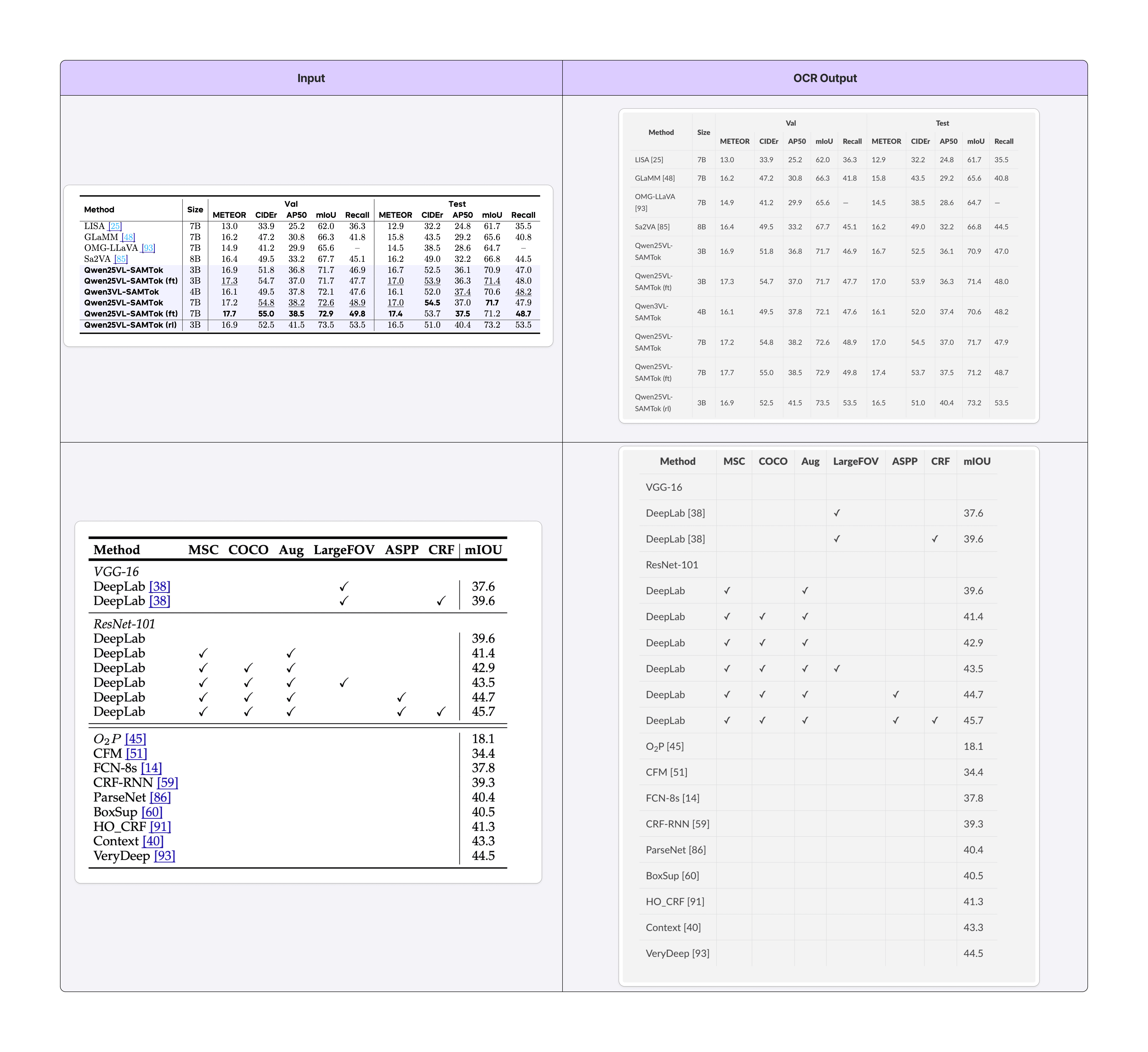}
  \caption{Example outputs for inputs with variations in structure of tables - sparse and dense with multi-column headers. Best viewed zoomed-in.}
  \label{fig:tables}
\end{figure}

In this section, we provide some qualitative results, showcasing the capabilities and some failure cases of our FalconOCR model. As seen from Figures~\ref{fig:paper2_ocr} to \ref{fig:tables}, our model ingests images taken in challenging real world conditions with challenging lighting and the text possessing differing semantics (formulas, tables, handwritten, etc) and output the proper structured text.
\begin{figure}[t!]
  \centering
  \includegraphics[width=1.0\textwidth, trim=4.5cm 4.5cm 4.5cm 4.5cm, clip]{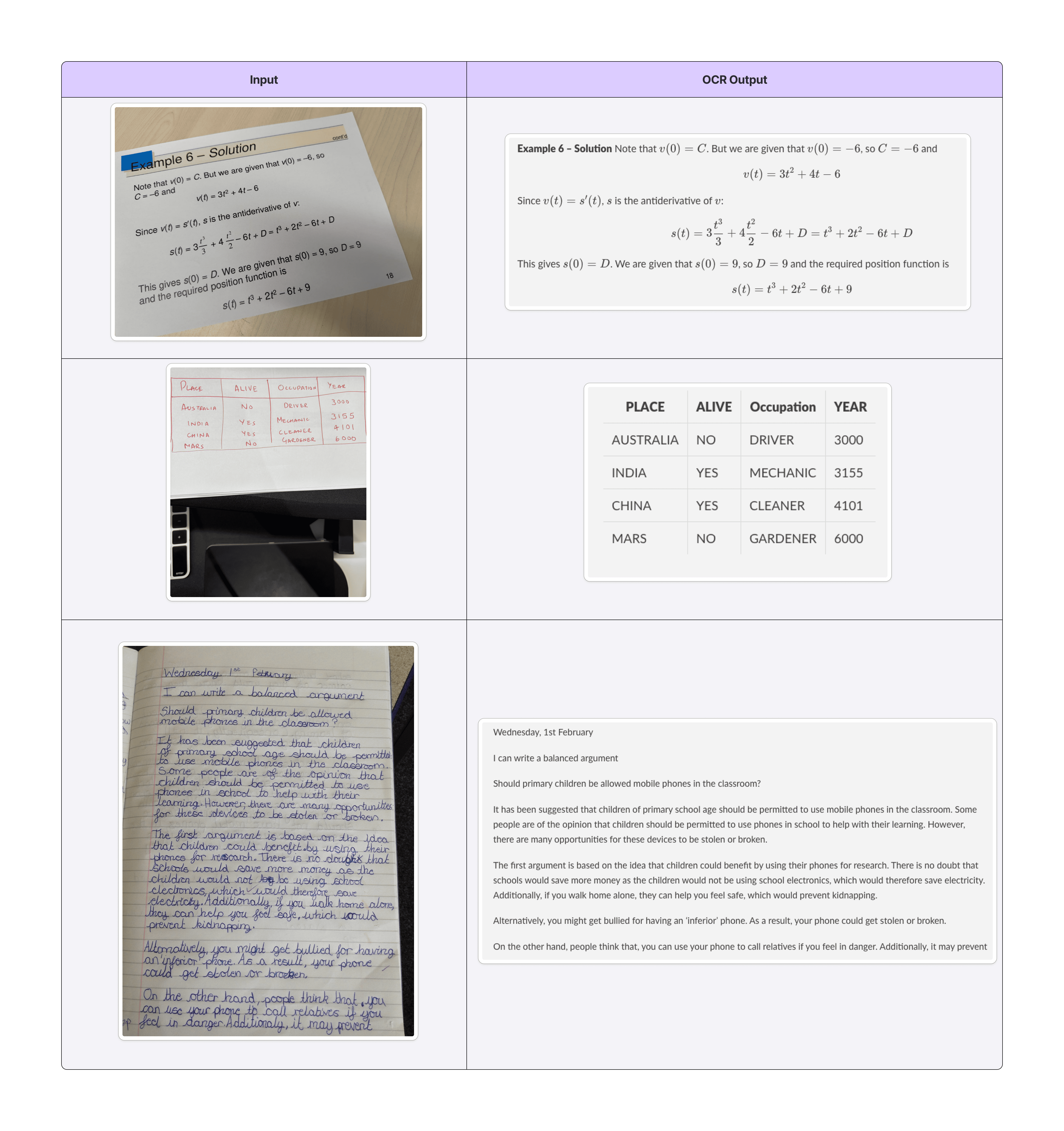}
  \caption{Example OCR output for inputs comprising formulae, tables and handwriting (from top to bottom). Best viewed zoomed-in.}
  \label{fig:realword_ocr}
\end{figure}

